%% file: main.tex
\RequirePackage{fix-cm}
\documentclass{article} %
\usepackage{iclr2027_conference,times}

\input{math_commands.tex}

\usepackage{hyperref}
\usepackage{url}
\hypersetup{hidelinks}
\usepackage{booktabs}
\usepackage{multirow}   %
\usepackage{booktabs}
\usepackage{multirow}
\usepackage[table]{xcolor}
\usepackage{graphicx}

\usepackage{tikz}

\definecolor{good}{HTML}{1B7837}   %
\definecolor{bad}{HTML}{B2182B}    %

\usepackage[utf8]{inputenc} %
\usepackage[T1]{fontenc}    %
\usepackage{url}            %
\usepackage{microtype}      %

\usepackage{amsmath}
\usepackage{amsfonts}       %
\usepackage{amssymb}
\usepackage{bm}
\usepackage{nicefrac}       %

\usepackage{makecell}
\usepackage{wrapfig}
\usepackage{caption}
\usepackage{subcaption}
\usepackage{textcomp}
\usepackage{verbatim}
\usepackage{xspace}

\usepackage{cleveref}       %

\newcommand{\baby}{ROAR\xspace}

\usepackage{pgfplots}
\usepackage{amsthm}

\usepackage{amsthm}

\usepackage{graphicx}
\usepackage{tikz}
\usepackage{textcomp}
\usepackage{amssymb}

\usepackage{booktabs}
\usepackage{booktabs,multirow,graphicx,amsmath}
\usepackage[table]{xcolor}
\usepackage{wrapfig}   %
\usepackage{booktabs}  %
\usepackage{multirow}  %
\usepackage{graphicx}  %
\usepackage{amsmath}
\usepackage{pgfplots}
\pgfplotsset{compat=1.18}
\usetikzlibrary{calc}
\usepackage{pgfplots}
\usepgfplotslibrary{groupplots}

\title{ROAR: Retrieval Opportunity-Aware Refinement for Zero-Shot Time Series Forecasting}

\author{%
\begin{tabular}{c}
\textbf{Jinjin Chi$^{1,2,3}$ \qquad Lei Feng$^{1}$ \qquad Yiming Wang$^{1}$ \qquad Lulu Zhang$^{1}$} \\
\textbf{Yongcheng Jing$^{2}$ \qquad Chenyi Li$^{1}$ \qquad Ximing Li$^{1}$} \\[4pt]
\normalfont\small $^1$College of Computer Science and Technology, Jilin University \\
\normalfont\small $^2$College of Computing and Data Science, Nanyang Technological University \\
\normalfont\small $^3$Key Laboratory of Symbolic Computation and Knowledge Engineering, \\
\normalfont\small MoE, Jilin University
\end{tabular}
}
\iclrfinalcopy
\begin{document}

\maketitle

\begin{abstract}
Retrieval augmentation provides time series forecasters with
historical continuations, yet even candidates that outperform
the base forecast may fail to improve the final prediction.
We propose ROAR, a Retrieval Opportunity-Aware Refinement
framework for zero-shot time series forecasting.
To better exploit these improvement opportunities, its
training objective allocates additional emphasis across
queries based on base-forecast difficulty and the relative
improvement offered by retrieved candidates.
Using this objective, ROAR first learns to aggregate aligned
historical candidates and uses a learned gate to control
their correction strength against a fixed base forecaster.
It then jointly calibrates the forecasting module and gate
to coordinate their contributions, while anchoring the combined
prediction to the first-stage refined output.
We derive exact decompositions of refinement gains and the
opportunity-weighted training loss, and characterize additional
gains from joint calibration under a local linearization.
Experiments on seven benchmarks show that ROAR achieves
the lowest average MSE among the compared methods.
Further evaluations demonstrate average forecasting improvements
across multiple backbone families and retrieval-augmented forecasters.
\end{abstract}

\input{introduction}
\input{related_work}
\input{our_method}
\input{experiment}

\section{Conclusion}
\label{sec:conclusion}
We presented ROAR, a framework that learns to refine time series
forecasts using retrieved historical continuations.
ROAR allocates additional training emphasis to difficult source
queries with beneficial candidates, then uses reference-guided joint
calibration to coordinate the base forecast and learned correction.
Our analyses characterize realized refinement gains, the training allocation induced by opportunity weighting, and conditions for additional gains from joint calibration under a local linearization.
Experiments on seven benchmarks demonstrate improved average forecasting accuracy across three backbone families and multiple retrieval-augmented hosts without target-specific parameter updates.
Stage ablations further show that refinement and joint calibration contribute complementary MSE improvements.

\newpage

\input{main.bbl}
\newpage
\appendix
\input{appendix}
\end{document}

%% file: math_commands.tex
\usepackage{amsmath,amsfonts,bm}

\def\eqref#1{equation~\ref{#1}}

\def\1{\bm{1}}

\DeclareMathAlphabet{\mathsfit}{\encodingdefault}{\sfdefault}{m}{sl}
\SetMathAlphabet{\mathsfit}{bold}{\encodingdefault}{\sfdefault}{bx}{n}

%% file: introduction.tex
\section{Introduction}

Time series forecasting predicts future observations from historical
data and supports applications including weather prediction
\citep{sun2022accurate,sapankevych2009time}, energy consumption
modeling \citep{tzelepi2023deep}, healthcare
\citep{jin2018treatment,tomar2025at4ts}, and financial analysis
\citep{qiu2025easytime,qiudbloss}.
Recently, Time Series Foundation Models (TSFMs) have demonstrated
promising zero-shot forecasting capabilities by learning temporal
patterns from large and diverse pretraining corpora
\citep{das2024decoder,ansari2024chronos}.
Nevertheless, their performance can deteriorate when the patterns
encountered at inference differ from those represented during
training \citep{kim2025comprehensive,zhou2023one,wang2024deep}.

Retrieval augmentation complements these learned patterns with
historical examples relevant to the current input
\citep{lewis2020retrieval,zhao2026retrieval,wu2024retrieval}.
Each historical example contains an observed context and its
subsequent trajectory, providing a candidate continuation for
the current query.
Previous studies have explored adaptive combinations of retrieved
forecasts and model outputs, as well as corrections based on
retrieved historical residuals
\citep{du2026predicting,he2026ratl}.
These developments motivate a key question:
\emph{how can useful historical candidates be translated into
improvements over an existing forecast?}

To investigate this question, we conduct an empirical analysis of
candidate usefulness and realized retrieval gains. We examine TS-RAG
\citep{ningts} and Cross-RAG \citep{lee2026cross} using Chronos-Bolt
as a common frozen backbone, and evaluate each context-aligned
historical continuation as an individual candidate forecast.
Fig.~\ref{fig:rag_analysis} shows that at least one candidate has
lower MSE than Chronos-Bolt for 46\% and 35\% of queries, respectively.
Yet for 17\% and 9\% of all queries, respectively, the
retrieval-augmented forecast has higher MSE than Chronos-Bolt despite
the availability of a better candidate.
The trajectory examples in Fig.~\ref{fig:rag_analysis}(c,d)
illustrate successful correction and a missed opportunity to improve
the forecast. These findings reveal a gap between the availability of
useful historical evidence and its effective integration into the
final forecast. Further details appear in Appendix~\ref{app:preliminary}.

{\parskip=0pt\par}
\begin{wrapfigure}{r}{0.65\columnwidth}
\centering
\vspace{-0.2em}
\input{plot/preliminary/pre_exp_latex_4_7data}
\caption{\textbf{Useful candidates do not always improve forecasts.}
(a, b) MSE changes of TS-RAG and Cross-RAG relative to Chronos-Bolt,
split by whether a better candidate exists.
(c, d) TS-RAG examples of a successful correction and a missed
opportunity.}
\label{fig:rag_analysis}
\vspace{2em}
\end{wrapfigure}
Motivated by this gap, we propose a
\textbf{R}etrieval \textbf{O}pportunity-\textbf{A}ware
\textbf{R}efinement framework (\textbf{ROAR}) that learns
retrieval-based forecast corrections on source data for
zero-shot transfer.
Its opportunity-aware objective allocates additional training
emphasis across queries according to base-forecast difficulty
and the relative improvement offered by retrieved candidates.
ROAR uses this objective to learn candidate aggregation and
correction strength while keeping the base forecaster fixed.
Reference-guided joint calibration then updates the trainable
forecasting module and gate, anchoring their combined prediction
to the initially refined forecast.

Our contributions are summarized as follows:
\WFclear
\begin{itemize}
    \item We provide an empirical diagnosis of retrieval-augmented
    forecasting, revealing that useful historical candidates can
    coexist with degraded forecasts.

    \item We introduce ROAR, which allocates training emphasis
    across queries according to candidate improvement opportunities
    and base-forecast difficulty, and coordinates the base prediction
    with learned corrections through reference-guided joint calibration.

 \item We derive exact decompositions of refinement gains and
the opportunity-weighted forecasting loss.
Under local linearization, we characterize the MSE advantage
of joint over single-component calibration at their
reference-regularized optima.

\item We evaluate ROAR on seven benchmarks in zero-shot
forecasting, demonstrating the lowest average MSE among
the compared methods and improvements across multiple
backbone families and retrieval-augmented forecasters.
\end{itemize}

%% file: plot/preliminary/pre_exp_latex_4_7data.tex
\definecolor{pxGT}{HTML}{C8323C}%
\definecolor{pxBolt}{HTML}{4E79A7}%
\definecolor{pxRag}{HTML}{E07B39}%
\definecolor{pxBest}{HTML}{3F7A49}%
\definecolor{pxNbr}{HTML}{C9C7C1}%
\definecolor{pxHist}{HTML}{9A9892}%
\definecolor{pxShade}{HTML}{F4F3F0}%
\definecolor{pxInk}{HTML}{6B6A66}%
\definecolor{pxGrid}{HTML}{EFEEEA}%
\definecolor{pxRingGP}{HTML}{4A4845}%
\definecolor{pxRingGz}{HTML}{CFCDC6}%
\definecolor{pxRed}{HTML}{CC7272}%
\definecolor{pxGreen}{HTML}{6B9E70}%
\definecolor{pxRedL}{HTML}{EFC8C7}%
\definecolor{pxGreenL}{HTML}{C9DEC8}%
\definecolor{pxDark}{HTML}{1A1A19}%
\providecommand{\textminus}{\ensuremath{-}}%
\providecommand{\prexpDelta}{{\usefont{U}{psy}{m}{n}D}}%
\providecommand{\prexpdelta}{{\usefont{U}{psy}{m}{n}d}}%
\providecommand{\prexpTick}{\pgfmathparse{\tick<0}\ifnum\pgfmathresult=1 \textminus\fi\pgfmathparse{abs(\tick)}\pgfmathprintnumber[assume math mode=true]{\pgfmathresult}}%
\begingroup%
\noindent\resizebox{\linewidth}{!}{\begin{minipage}{\textwidth}%
\setbox0=\hbox{\begin{tikzpicture}[x=5.2cm, y=5.2cm]
\filldraw[fill=pxGreen, draw=white, line width=2pt] ([shift={(0.0,0)}]19.980:0.76) arc[start angle=19.980, end angle=83.340, radius=0.76] -- ([shift={(0.0,0)}]83.340:1.2) arc[start angle=83.340, end angle=19.980, radius=1.2] -- cycle;
\node[text=white, inner sep=0pt, font=\fontsize{33}{36}\selectfont\bfseries] at (0.6079,0.7687) {17\%};
\node[anchor=west, align=left, inner sep=2pt, text=pxDark, font=\fontsize{33}{36}\selectfont\bfseries] at (1.0262,0.9412) {\prexpDelta: \textminus{}7.4\%\\\prexpdelta: \textminus{}4.9\%};
\filldraw[fill=pxRed, draw=white, line width=2pt] ([shift={(0.0,0)}]-83.340:0.76) arc[start angle=-83.340, end angle=19.980, radius=0.76] -- ([shift={(0.0,0)}]19.980:1.2) arc[start angle=19.980, end angle=-83.340, radius=1.2] -- cycle;
\node[text=white, inner sep=0pt, font=\fontsize{33}{36}\selectfont\bfseries] at (0.8340,-0.5147) {29\%};
\node[anchor=west, align=left, inner sep=2pt, text=pxDark, font=\fontsize{33}{36}\selectfont\bfseries] at (1.1859,-0.6302) {\prexpDelta: +8.0\%\\\prexpdelta: +6.0\%};
\filldraw[fill=pxGreenL, draw=white, line width=2pt] ([shift={(0.0,0)}]83.340:0.76) arc[start angle=83.340, end angle=190.421, radius=0.76] -- ([shift={(0.0,0)}]190.421:1.2) arc[start angle=190.421, end angle=83.340, radius=1.2] -- cycle;
\node[text=pxDark, inner sep=0pt, font=\fontsize{33}{36}\selectfont\bfseries] at (-0.7153,0.6699) {30\%};
\node[anchor=east, align=right, inner sep=2pt, text=pxDark, font=\fontsize{33}{36}\selectfont\bfseries] at (-1.1022,0.8202) {\prexpDelta: \textminus{}13.0\%\\\prexpdelta: \textminus{}8.5\%};
\filldraw[fill=pxRedL, draw=white, line width=2pt] ([shift={(0.0,0)}]190.421:0.76) arc[start angle=190.421, end angle=276.660, radius=0.76] -- ([shift={(0.0,0)}]276.660:1.2) arc[start angle=276.660, end angle=190.421, radius=1.2] -- cycle;
\node[text=pxDark, inner sep=0pt, font=\fontsize{33}{36}\selectfont\bfseries] at (-0.5824,-0.7882) {24\%};
\node[anchor=east, align=right, inner sep=2pt, text=pxDark, font=\fontsize{33}{36}\selectfont\bfseries] at (-1.0086,-0.9651) {\prexpDelta: +6.1\%\\\prexpdelta: +4.2\%};
\draw[white, line width=5pt] (0.0881,0.7549) -- (0.1392,1.1919);
\draw[white, line width=5pt] (0.0881,-0.7549) -- (0.1392,-1.1919);
\filldraw[fill=pxRingGP, draw=white, line width=2.5pt] ([shift={(0.0,0)}]-83.340:0.64) arc[start angle=-83.340, end angle=83.340, radius=0.64] -- ([shift={(0.0,0)}]83.340:0.71) arc[start angle=83.340, end angle=-83.340, radius=0.71] -- cycle;
\filldraw[fill=pxRingGz, draw=white, line width=2.5pt] ([shift={(0.0,0)}]83.340:0.64) arc[start angle=83.340, end angle=276.660, radius=0.64] -- ([shift={(0.0,0)}]276.660:0.71) arc[start angle=276.660, end angle=83.340, radius=0.71] -- cycle;
\node[align=center, text=pxRingGP, font=\fontsize{33}{36}\selectfont\bfseries] at (0.290,0) {better\\nbr.\\46\%};
\node[align=center, text=pxInk, font=\fontsize{33}{36}\selectfont\bfseries] at (-0.290,0) {no\\better\\54\%};
\filldraw[fill=pxGreen, draw=white, line width=2pt] ([shift={(5.035,0)}]28.980:0.76) arc[start angle=28.980, end angle=63.540, radius=0.76] -- ([shift={(5.035,0)}]63.540:1.2) arc[start angle=63.540, end angle=28.980, radius=1.2] -- cycle;
\node[text=white, inner sep=0pt, font=\fontsize{33}{36}\selectfont\bfseries] at (5.7126,0.7080) {9\%};
\node[anchor=west, align=left, inner sep=2pt, text=pxDark, font=\fontsize{33}{36}\selectfont\bfseries] at (6.1102,0.8670) {\prexpDelta: \textminus{}10.0\%\\\prexpdelta: \textminus{}7.7\%};
\filldraw[fill=pxRed, draw=white, line width=2pt] ([shift={(5.035,0)}]-63.540:0.76) arc[start angle=-63.540, end angle=28.980, radius=0.76] -- ([shift={(5.035,0)}]28.980:1.2) arc[start angle=28.980, end angle=-63.540, radius=1.2] -- cycle;
\node[text=white, inner sep=0pt, font=\fontsize{33}{36}\selectfont\bfseries] at (5.9708,-0.2911) {26\%};
\node[anchor=west, align=left, inner sep=2pt, text=pxDark, font=\fontsize{33}{36}\selectfont\bfseries] at (6.2797,-0.3564) {\prexpDelta: +15.3\%\\\prexpdelta: +8.9\%};
\filldraw[fill=pxGreenL, draw=white, line width=2pt] ([shift={(5.035,0)}]63.540:0.76) arc[start angle=63.540, end angle=211.860, radius=0.76] -- ([shift={(5.035,0)}]211.860:1.2) arc[start angle=211.860, end angle=63.540, radius=1.2] -- cycle;
\node[text=pxDark, inner sep=0pt, font=\fontsize{33}{36}\selectfont\bfseries] at (4.3102,0.6596) {41\%};
\node[anchor=east, align=right, inner sep=2pt, text=pxDark, font=\fontsize{33}{36}\selectfont\bfseries] at (3.9260,0.8076) {\prexpDelta: \textminus{}24.1\%\\\prexpdelta: \textminus{}16.6\%};
\filldraw[fill=pxRedL, draw=white, line width=2pt] ([shift={(5.035,0)}]211.860:0.76) arc[start angle=211.860, end angle=296.460, radius=0.76] -- ([shift={(5.035,0)}]296.460:1.2) arc[start angle=296.460, end angle=211.860, radius=1.2] -- cycle;
\node[text=pxDark, inner sep=0pt, font=\fontsize{33}{36}\selectfont\bfseries] at (4.7675,-0.9428) {24\%};
\node[anchor=east, align=right, inner sep=2pt, text=pxDark, font=\fontsize{33}{36}\selectfont\bfseries] at (4.0430,-0.9867) {\prexpDelta: +14.0\%\\\prexpdelta: +8.6\%};
\draw[white, line width=5pt] (5.3736,0.6804) -- (5.5697,1.0743);
\draw[white, line width=5pt] (5.3736,-0.6804) -- (5.5697,-1.0743);
\filldraw[fill=pxRingGP, draw=white, line width=2.5pt] ([shift={(5.035,0)}]-63.540:0.64) arc[start angle=-63.540, end angle=63.540, radius=0.64] -- ([shift={(5.035,0)}]63.540:0.71) arc[start angle=63.540, end angle=-63.540, radius=0.71] -- cycle;
\filldraw[fill=pxRingGz, draw=white, line width=2.5pt] ([shift={(5.035,0)}]63.540:0.64) arc[start angle=63.540, end angle=296.460, radius=0.64] -- ([shift={(5.035,0)}]296.460:0.71) arc[start angle=296.460, end angle=63.540, radius=0.71] -- cycle;
\node[align=center, text=pxRingGP, font=\fontsize{33}{36}\selectfont\bfseries] at (5.325,0) {better\\nbr.\\35\%};
\node[align=center, text=pxInk, font=\fontsize{33}{36}\selectfont\bfseries] at (4.745,0) {no\\better\\65\%};
\pgfresetboundingbox
\path[use as bounding box] (-2.3925,-1.23) rectangle (7.4275,1.23);
\end{tikzpicture}}%
\setbox2=\hbox{\begin{tikzpicture}
\begin{axis}[name=ax0, width=7.2cm, height=4.3cm, scale only axis, xmin=-103.95, xmax=70.95, ymin=-14.684, ymax=15.185, axis x line*=bottom, axis y line*=left, axis line style={pxInk, -stealth, line width=2.2pt}, tick style={line width=1.4pt}, tick style={pxInk}, tick label style={color=pxInk}, xtick={-80,-40,0,40}, ymajorgrids, xticklabel={\prexpTick}, yticklabel={\prexpTick}, grid style={pxGrid, line width=0.6pt}, tick label style={font=\fontsize{15.8}{19}\selectfont}, label style={font=\fontsize{11.664773617421273}{13.997728340905528}\selectfont\bfseries}, ylabel={Value}]
\path[fill=pxShade] (axis cs:0,-14.684) rectangle (axis cs:63,15.185);
\addplot[pxHist, line width=1.1pt] coordinates {(-96,9.808) (-95,9.808) (-94,9.808) (-93,9.808) (-92,9.808) (-91,9.808) (-90,9.808) (-89,9.808) (-88,9.808) (-87,9.808) (-86,9.808) (-85,9.808) (-84,9.808) (-83,9.808) (-82,9.808) (-81,9.808) (-80,9.808) (-79,9.808) (-78,9.808) (-77,9.808) (-76,9.808) (-75,9.808) (-74,9.808) (-73,9.808) (-72,9.808) (-71,9.808) (-70,9.808) (-69,9.808) (-68,9.808) (-67,9.808) (-66,9.808) (-65,9.808) (-64,9.808) (-63,9.808) (-62,9.808) (-61,9.808) (-60,9.808) (-59,9.808) (-58,9.808) (-57,9.808) (-56,9.808) (-55,9.808) (-54,9.808) (-53,8.742) (-52,11.051) (-51,11.300) (-50,11.584) (-49,10.838) (-48,11.194) (-47,11.087) (-46,10.980) (-45,10.661) (-44,10.696) (-43,11.549) (-42,10.589) (-41,9.630) (-40,11.087) (-39,10.021) (-38,10.305) (-37,9.701) (-36,8.244) (-35,8.955) (-34,8.990) (-33,9.026) (-32,9.630) (-31,9.594) (-30,8.848) (-29,8.884) (-28,8.706) (-27,9.239) (-26,9.772) (-25,9.914) (-24,10.305) (-23,10.376) (-22,9.204) (-21,7.818) (-20,6.396) (-19,5.401) (-18,3.625) (-17,0.249) (-16,-2.878) (-15,-3.696) (-14,-4.548) (-13,-6.538) (-12,-7.320) (-11,-9.132) (-10,-9.630) (-9,-11.122) (-8,-11.194) (-7,-12.473) (-6,-10.909) (-5,-9.808) (-4,-13.326) (-3,-12.686) (-2,-12.970) (-1,-12.082) (0,-12.864)};
\addplot[pxNbr, line width=0.8pt] coordinates {(0,-1.694) (1,-0.831) (2,-0.413) (3,0.371) (4,0.998) (5,1.390) (6,1.312) (7,2.410) (8,2.697) (9,3.350) (10,4.605) (11,5.337) (12,6.330) (13,7.010) (14,7.716) (15,8.578) (16,9.048) (17,8.238) (18,8.290) (19,8.656) (20,8.865) (21,9.179) (22,9.493) (23,9.571) (24,9.676) (25,9.963) (26,10.146) (27,10.172) (28,10.120) (29,9.807) (30,9.885) (31,10.303) (32,10.146) (33,11.139) (34,11.949) (35,12.028) (36,11.192) (37,11.296) (38,10.460) (39,11.949) (40,11.505) (41,11.113) (42,11.270) (43,11.217) (44,10.695) (45,10.616) (46,11.087) (47,11.061) (48,10.591) (49,10.773) (50,8.656) (51,8.709) (52,11.819) (53,11.453) (54,11.139) (55,11.061) (56,9.728) (57,11.793) (58,11.270) (59,11.479) (60,10.800) (61,12.132) (62,10.669) (63,11.008)};
\addplot[pxNbr, line width=0.8pt] coordinates {(0,0.215) (1,-0.399) (2,-0.668) (3,-0.104) (4,4.974) (5,0.313) (6,2.889) (7,3.109) (8,2.962) (9,4.606) (10,5.636) (11,3.943) (12,4.557) (13,6.519) (14,7.059) (15,7.402) (16,7.770) (17,8.015) (18,8.138) (19,8.236) (20,8.432) (21,8.506) (22,8.530) (23,8.579) (24,8.506) (25,8.481) (26,8.383) (27,8.187) (28,8.089) (29,8.211) (30,9.512) (31,9.389) (32,9.413) (33,9.561) (34,9.978) (35,10.738) (36,10.346) (37,10.616) (38,10.984) (39,11.401) (40,11.450) (41,11.548) (42,11.523) (43,10.861) (44,10.935) (45,10.837) (46,11.646) (47,11.572) (48,11.670) (49,10.689) (50,11.842) (51,11.327) (52,11.548) (53,11.499) (54,11.621) (55,12.038) (56,10.591) (57,10.591) (58,11.303) (59,10.223) (60,10.910) (61,11.082) (62,10.542) (63,11.768)};
\addplot[pxNbr, line width=0.8pt] coordinates {(0,-6.871) (1,-5.940) (2,-5.424) (3,-5.424) (4,-4.928) (5,-4.988) (6,-1.024) (7,-4.631) (8,-4.275) (9,-3.184) (10,0.205) (11,0.443) (12,2.406) (13,-0.250) (14,3.436) (15,4.487) (16,5.359) (17,5.656) (18,6.430) (19,6.787) (20,7.064) (21,7.401) (22,7.500) (23,7.540) (24,7.659) (25,7.956) (26,8.115) (27,8.234) (28,7.956) (29,8.134) (30,8.095) (31,8.036) (32,8.016) (33,7.936) (34,7.976) (35,7.797) (36,7.976) (37,7.916) (38,7.818) (39,7.778) (40,8.888) (41,8.571) (42,8.511) (43,8.353) (44,9.324) (45,9.245) (46,9.284) (47,9.621) (48,10.038) (49,9.720) (50,9.304) (51,9.423) (52,9.978) (53,9.284) (54,9.641) (55,9.581) (56,9.146) (57,8.928) (58,8.987) (59,9.780) (60,9.443) (61,9.165) (62,9.324) (63,7.837)};
\addplot[pxNbr, line width=0.8pt] coordinates {(0,-2.073) (1,-2.327) (2,-0.727) (3,-2.428) (4,-1.540) (5,-4.028) (6,-1.591) (7,-1.515) (8,-4.002) (9,0.186) (10,0.491) (11,-1.185) (12,-2.454) (13,-1.337) (14,-1.819) (15,-0.677) (16,-0.067) (17,3.360) (18,0.389) (19,3.487) (20,4.985) (21,6.406) (22,5.746) (23,5.340) (24,6.483) (25,5.949) (26,6.229) (27,6.584) (28,7.625) (29,8.108) (30,8.158) (31,8.488) (32,8.666) (33,8.869) (34,9.275) (35,9.504) (36,9.554) (37,9.580) (38,9.402) (39,9.275) (40,9.529) (41,9.250) (42,8.920) (43,8.945) (44,9.301) (45,9.199) (46,9.402) (47,9.479) (48,9.935) (49,9.935) (50,9.402) (51,9.935) (52,9.884) (53,9.377) (54,10.139) (55,9.504) (56,9.047) (57,9.377) (58,9.631) (59,9.224) (60,9.123) (61,9.174) (62,8.488) (63,9.149)};
\addplot[pxNbr, line width=0.8pt] coordinates {(0,-6.550) (1,-6.311) (2,-6.311) (3,-5.733) (4,-4.031) (5,-2.909) (6,-3.589) (7,-2.875) (8,-2.500) (9,-0.629) (10,3.385) (11,3.317) (12,5.087) (13,5.155) (14,6.652) (15,6.584) (16,6.550) (17,6.447) (18,7.231) (19,6.958) (20,7.163) (21,7.503) (22,7.503) (23,7.571) (24,7.332) (25,7.367) (26,7.231) (27,6.958) (28,7.060) (29,7.503) (30,9.374) (31,9.510) (32,9.204) (33,9.238) (34,10.905) (35,11.756) (36,11.857) (37,12.130) (38,10.293) (39,11.007) (40,12.198) (41,11.892) (42,13.151) (43,12.980) (44,10.599) (45,12.096) (46,12.879) (47,11.313) (48,11.756) (49,13.184) (50,12.096) (51,11.245) (52,10.599) (53,10.190) (54,10.361) (55,9.374) (56,10.225) (57,9.102) (58,8.251) (59,10.326) (60,10.633) (61,9.884) (62,9.408) (63,11.109)};
\addplot[pxNbr, line width=0.8pt] coordinates {(0,-11.834) (1,-7.552) (2,-3.358) (3,-6.848) (4,-8.754) (5,-9.136) (6,-8.490) (7,-7.786) (8,-6.936) (9,-5.910) (10,-5.968) (11,-5.558) (12,-3.769) (13,-1.951) (14,-2.068) (15,-1.423) (16,0.835) (17,3.475) (18,4.090) (19,5.147) (20,5.615) (21,6.701) (22,7.669) (23,7.552) (24,7.463) (25,7.463) (26,7.756) (27,7.845) (28,8.079) (29,7.786) (30,7.552) (31,7.463) (32,8.226) (33,7.288) (34,7.288) (35,9.340) (36,8.636) (37,7.522) (38,7.493) (39,8.314) (40,11.188) (41,11.569) (42,11.804) (43,11.158) (44,10.982) (45,12.126) (46,12.156) (47,12.683) (48,12.772) (49,13.123) (50,12.977) (51,11.892) (52,12.537) (53,11.979) (54,11.569) (55,12.507) (56,12.713) (57,12.683) (58,12.068) (59,13.680) (60,13.534) (61,13.827) (62,12.918) (63,11.950)};
\addplot[pxNbr, line width=0.8pt] coordinates {(0,-3.889) (1,0.622) (2,-4.116) (3,-0.134) (4,-3.385) (5,2.159) (6,-0.311) (7,-2.276) (8,3.217) (9,5.687) (10,6.292) (11,-1.142) (12,0.042) (13,0.420) (14,1.932) (15,2.159) (16,3.092) (17,4.150) (18,5.007) (19,6.267) (20,7.250) (21,7.653) (22,7.728) (23,8.913) (24,8.434) (25,9.568) (26,9.719) (27,9.946) (28,10.097) (29,10.274) (30,10.475) (31,10.702) (32,10.702) (33,10.627) (34,10.526) (35,10.400) (36,10.249) (37,10.148) (38,10.199) (39,10.047) (40,9.518) (41,9.594) (42,8.812) (43,9.216) (44,9.594) (45,10.324) (46,10.173) (47,10.803) (48,10.728) (49,11.055) (50,11.282) (51,10.853) (52,10.627) (53,11.055) (54,9.543) (55,10.173) (56,10.828) (57,10.828) (58,9.946) (59,9.821) (60,9.594) (61,8.989) (62,9.594) (63,9.594)};
\addplot[pxNbr, line width=0.8pt] coordinates {(0,2.252) (1,2.714) (2,4.387) (3,4.676) (4,3.406) (5,3.983) (6,3.983) (7,3.580) (8,3.232) (9,3.232) (10,3.465) (11,2.945) (12,2.771) (13,3.060) (14,2.482) (15,2.367) (16,2.136) (17,1.444) (18,1.617) (19,1.617) (20,1.849) (21,1.617) (22,1.155) (23,-0.748) (24,-0.980) (25,-1.614) (26,-1.788) (27,-1.614) (28,-1.326) (29,0.405) (30,1.617) (31,3.232) (32,5.773) (33,6.697) (34,6.869) (35,6.464) (36,6.464) (37,9.524) (38,10.159) (39,9.755) (40,9.409) (41,8.946) (42,10.620) (43,12.582) (44,12.178) (45,12.006) (46,13.333) (47,13.506) (48,11.832) (49,12.236) (50,11.601) (51,11.717) (52,11.717) (53,10.966) (54,10.331) (55,12.236) (56,8.543) (57,11.312) (58,10.909) (59,9.294) (60,9.409) (61,9.350) (62,9.178) (63,8.889)};
\addplot[pxNbr, line width=0.8pt] coordinates {(0,-2.630) (1,-2.429) (2,-2.026) (3,-1.699) (4,-1.774) (5,-0.589) (6,-0.338) (7,0.116) (8,0.393) (9,0.418) (10,1.200) (11,1.300) (12,1.779) (13,3.543) (14,4.249) (15,5.584) (16,6.138) (17,6.617) (18,7.121) (19,7.323) (20,7.347) (21,7.574) (22,7.675) (23,8.003) (24,8.053) (25,8.053) (26,8.003) (27,7.877) (28,7.751) (29,7.776) (30,7.423) (31,7.398) (32,7.423) (33,7.650) (34,7.499) (35,7.725) (36,7.574) (37,8.129) (38,8.205) (39,7.978) (40,8.834) (41,9.313) (42,10.800) (43,10.875) (44,10.774) (45,9.540) (46,10.976) (47,9.767) (48,10.170) (49,10.951) (50,9.943) (51,9.842) (52,9.389) (53,10.649) (54,10.346) (55,10.472) (56,9.212) (57,10.019) (58,9.666) (59,10.875) (60,10.573) (61,11.001) (62,10.724) (63,9.994)};
\addplot[pxBest, line width=1.5pt, dashed] coordinates {(0,-6.615) (1,-5.931) (2,-6.615) (3,-5.716) (4,-5.033) (5,-4.672) (6,-3.773) (7,-3.306) (8,-2.802) (9,-2.695) (10,-1.651) (11,-1.292) (12,-0.428) (13,0.651) (14,2.125) (15,4.283) (16,4.642) (17,5.434) (18,5.902) (19,6.369) (20,6.800) (21,7.017) (22,7.304) (23,7.916) (24,7.844) (25,7.879) (26,7.952) (27,7.988) (28,7.520) (29,7.628) (30,7.556) (31,8.168) (32,8.059) (33,7.700) (34,7.988) (35,6.729) (36,8.420) (37,8.959) (38,10.578) (39,10.074) (40,10.326) (41,9.714) (42,8.923) (43,10.038) (44,10.146) (45,8.455) (46,9.822) (47,9.966) (48,8.886) (49,9.678) (50,9.750) (51,9.462) (52,10.289) (53,10.398) (54,9.930) (55,10.182) (56,10.110) (57,11.620) (58,9.426) (59,9.462) (60,10.829) (61,9.210) (62,10.254) (63,9.534)};
\addplot[pxGT, line width=1.5pt] coordinates {(0,-12.864) (1,-11.833) (2,-11.513) (3,-9.985) (4,-8.884) (5,-6.645) (6,-6.823) (7,-6.823) (8,-6.076) (9,-0.071) (10,2.736) (11,1.457) (12,3.198) (13,4.193) (14,5.188) (15,5.970) (16,6.432) (17,6.254) (18,6.219) (19,6.396) (20,6.396) (21,6.929) (22,6.716) (23,6.752) (24,6.823) (25,6.503) (26,6.432) (27,6.396) (28,6.396) (29,6.432) (30,6.254) (31,7.604) (32,7.711) (33,7.889) (34,7.782) (35,7.533) (36,8.813) (37,8.635) (38,9.026) (39,9.666) (40,9.097) (41,10.127) (42,9.417) (43,9.132) (44,7.889) (45,10.305) (46,11.655) (47,9.808) (48,10.803) (49,9.168) (50,11.620) (51,11.549) (52,9.950) (53,10.021) (54,10.767) (55,11.655) (56,10.056) (57,11.336) (58,10.199) (59,9.879) (60,9.594) (61,9.594) (62,8.599) (63,9.310)};
\addplot[pxBolt, line width=1.7pt] coordinates {(0,-13.268) (1,-12.702) (2,-11.779) (3,-11.183) (4,-10.954) (5,-10.583) (6,-10.174) (7,-9.417) (8,-8.859) (9,-8.242) (10,-7.414) (11,-6.661) (12,-5.791) (13,-4.934) (14,-4.397) (15,-4.118) (16,-3.850) (17,-3.785) (18,-3.636) (19,-3.232) (20,-2.916) (21,-2.838) (22,-2.639) (23,-2.418) (24,-2.405) (25,-2.414) (26,-2.254) (27,-2.206) (28,-2.236) (29,-2.059) (30,-2.060) (31,-1.665) (32,-1.204) (33,-0.927) (34,-0.621) (35,-0.324) (36,0.356) (37,0.814) (38,1.118) (39,1.506) (40,2.075) (41,2.213) (42,2.223) (43,2.456) (44,2.765) (45,3.097) (46,3.366) (47,3.569) (48,3.533) (49,3.486) (50,3.580) (51,3.552) (52,3.512) (53,3.451) (54,3.279) (55,3.426) (56,3.250) (57,3.277) (58,3.237) (59,3.069) (60,3.269) (61,3.323) (62,3.312) (63,2.867)};
\addplot[pxRag, line width=1.7pt] coordinates {(0,-12.219) (1,-11.408) (2,-10.416) (3,-9.280) (4,-8.829) (5,-8.366) (6,-7.350) (7,-6.220) (8,-5.127) (9,-4.259) (10,-3.322) (11,-2.310) (12,-1.091) (13,-0.060) (14,0.788) (15,1.080) (16,2.045) (17,2.170) (18,2.555) (19,3.405) (20,4.156) (21,4.357) (22,4.353) (23,4.907) (24,5.449) (25,5.441) (26,5.856) (27,6.189) (28,6.393) (29,6.747) (30,6.964) (31,7.454) (32,8.069) (33,8.503) (34,9.019) (35,9.272) (36,9.911) (37,10.540) (38,10.714) (39,11.169) (40,11.907) (41,11.833) (42,11.986) (43,12.169) (44,12.615) (45,12.892) (46,12.881) (47,12.841) (48,12.627) (49,12.290) (50,12.473) (51,12.450) (52,12.367) (53,12.043) (54,12.014) (55,12.056) (56,11.695) (57,11.609) (58,11.540) (59,11.030) (60,11.273) (61,11.236) (62,11.056) (63,10.285)};
\draw[pxInk, -stealth, line width=2.2pt] (rel axis cs:0,0) -- (rel axis cs:1,0);
\end{axis}
\begin{axis}[name=ax1, width=7.2cm, height=4.3cm, scale only axis, xmin=-103.95, xmax=70.95, ymin=8.941, ymax=45.471, axis x line*=bottom, axis y line*=left, axis line style={pxInk, -stealth, line width=2.2pt}, tick style={line width=1.4pt}, tick style={pxInk}, tick label style={color=pxInk}, xtick={-80,-40,0,40}, ymajorgrids, xticklabel={\prexpTick}, yticklabel={\prexpTick}, grid style={pxGrid, line width=0.6pt}, tick label style={font=\fontsize{15.8}{19}\selectfont}, label style={font=\fontsize{11.664773617421273}{13.997728340905528}\selectfont\bfseries}, at={(ax0.south east)}, xshift=1.9cm]
\path[fill=pxShade] (axis cs:0,8.941) rectangle (axis cs:63,45.471);
\addplot[pxHist, line width=1.1pt] coordinates {(-96,37.193) (-95,35.685) (-94,37.779) (-93,37.863) (-92,36.188) (-91,36.942) (-90,36.439) (-89,37.863) (-88,36.607) (-87,37.109) (-86,36.942) (-85,37.109) (-84,37.528) (-83,37.528) (-82,37.779) (-81,38.952) (-80,39.120) (-79,41.046) (-78,40.795) (-77,39.287) (-76,39.120) (-75,41.716) (-74,42.219) (-73,42.219) (-72,40.376) (-71,41.968) (-70,38.785) (-69,43.811) (-68,39.622) (-67,38.952) (-66,41.298) (-65,42.889) (-64,40.041) (-63,39.538) (-62,42.303) (-61,41.800) (-60,39.455) (-59,39.538) (-58,39.120) (-57,39.790) (-56,38.198) (-55,36.942) (-54,35.266) (-53,37.193) (-52,36.272) (-51,36.774) (-50,36.020) (-49,34.261) (-48,34.512) (-47,33.926) (-46,30.994) (-45,31.999) (-44,30.575) (-43,31.162) (-42,31.999) (-41,31.916) (-40,33.507) (-39,32.670) (-38,34.010) (-37,33.423) (-36,29.905) (-35,28.732) (-34,28.816) (-33,26.890) (-32,24.293) (-31,24.376) (-30,28.146) (-29,28.062) (-28,27.560) (-27,27.225) (-26,25.298) (-25,27.560) (-24,29.403) (-23,30.827) (-22,27.308) (-21,30.324) (-20,31.916) (-19,28.816) (-18,27.476) (-17,27.895) (-16,31.581) (-15,34.596) (-14,30.240) (-13,34.094) (-12,35.434) (-11,35.853) (-10,30.659) (-9,27.057) (-8,26.387) (-7,26.890) (-6,27.560) (-5,25.801) (-4,25.130) (-3,22.952) (-2,22.115) (-1,20.858) (0,19.853)};
\addplot[pxNbr, line width=0.8pt] coordinates {(0,28.815) (1,32.625) (2,34.365) (3,35.359) (4,32.625) (5,35.773) (6,32.128) (7,34.696) (8,35.111) (9,32.625) (10,34.200) (11,34.365) (12,34.200) (13,33.702) (14,35.939) (15,35.277) (16,35.773) (17,35.773) (18,39.502) (19,38.839) (20,35.194) (21,37.762) (22,38.507) (23,36.105) (24,40.827) (25,38.590) (26,34.365) (27,38.425) (28,39.004) (29,36.188) (30,39.087) (31,40.164) (32,43.229) (33,41.490) (34,42.650) (35,40.827) (36,40.827) (37,40.744) (38,39.253) (39,40.413) (40,34.779) (41,37.844) (42,38.342) (43,35.939) (44,32.957) (45,34.034) (46,33.785) (47,33.785) (48,30.388) (49,29.809) (50,34.200) (51,32.294) (52,29.228) (53,30.886) (54,32.046) (55,32.708) (56,31.631) (57,28.815) (58,30.140) (59,30.969) (60,31.465) (61,32.957) (62,29.394) (63,31.051)};
\addplot[pxNbr, line width=0.8pt] coordinates {(0,20.649) (1,20.363) (2,20.879) (3,20.649) (4,19.847) (5,20.822) (6,19.904) (7,20.190) (8,20.190) (9,20.076) (10,20.076) (11,20.477) (12,19.674) (13,20.420) (14,20.707) (15,22.714) (16,22.313) (17,23.288) (18,23.345) (19,23.001) (20,23.058) (21,24.091) (22,24.377) (23,24.664) (24,25.123) (25,25.581) (26,27.360) (27,28.507) (28,29.654) (29,29.769) (30,29.883) (31,29.138) (32,30.284) (33,29.826) (34,29.367) (35,29.138) (36,29.080) (37,28.105) (38,28.679) (39,27.360) (40,27.761) (41,27.589) (42,26.843) (43,29.539) (44,28.966) (45,28.335) (46,29.998) (47,27.761) (48,27.303) (49,30.055) (50,27.876) (51,28.450) (52,29.826) (53,27.704) (54,27.303) (55,29.482) (56,29.138) (57,27.589) (58,28.679) (59,29.482) (60,26.729) (61,28.564) (62,29.596) (63,27.073)};
\addplot[pxNbr, line width=0.8pt] coordinates {(0,19.779) (1,20.418) (2,20.128) (3,20.651) (4,20.418) (5,19.604) (6,20.593) (7,19.662) (8,19.953) (9,19.953) (10,19.837) (11,19.837) (12,20.243) (13,19.430) (14,20.185) (15,20.476) (16,22.510) (17,22.104) (18,23.092) (19,23.150) (20,22.801) (21,22.860) (22,23.906) (23,24.196) (24,24.487) (25,24.952) (26,25.417) (27,27.219) (28,28.382) (29,29.544) (30,29.661) (31,29.776) (32,29.021) (33,30.183) (34,29.718) (35,29.253) (36,29.021) (37,28.962) (38,27.974) (39,28.556) (40,27.219) (41,27.626) (42,27.451) (43,26.695) (44,29.427) (45,28.847) (46,28.207) (47,29.893) (48,27.626) (49,27.161) (50,29.951) (51,27.742) (52,28.323) (53,29.718) (54,27.568) (55,27.161) (56,29.370) (57,29.021) (58,27.451) (59,28.556) (60,29.370) (61,26.580) (62,28.439) (63,29.486)};
\addplot[pxNbr, line width=0.8pt] coordinates {(0,28.995) (1,28.384) (2,27.948) (3,28.384) (4,30.393) (5,28.909) (6,26.988) (7,25.940) (8,27.861) (9,27.861) (10,28.821) (11,29.519) (12,29.519) (13,29.345) (14,35.981) (15,34.409) (16,34.846) (17,36.068) (18,35.544) (19,36.591) (20,35.981) (21,34.496) (22,36.679) (23,35.370) (24,33.798) (25,35.631) (26,36.854) (27,36.068) (28,38.949) (29,38.688) (30,39.298) (31,36.679) (32,34.584) (33,36.505) (34,36.068) (35,37.814) (36,39.386) (37,40.259) (38,39.211) (39,40.346) (40,38.688) (41,37.377) (42,37.989) (43,35.719) (44,36.505) (45,36.679) (46,36.679) (47,35.544) (48,35.456) (49,33.886) (50,31.702) (51,34.409) (52,33.361) (53,30.044) (54,30.218) (55,31.091) (56,31.789) (57,29.519) (58,32.226) (59,31.702) (60,31.353) (61,30.654) (62,31.265) (63,30.130)};
\addplot[pxNbr, line width=0.8pt] coordinates {(0,28.893) (1,32.410) (2,32.965) (3,31.762) (4,33.520) (5,33.520) (6,32.039) (7,33.520) (8,33.242) (9,33.520) (10,35.370) (11,39.256) (12,40.182) (13,40.922) (14,37.129) (15,39.072) (16,37.592) (17,39.165) (18,39.905) (19,38.702) (20,38.423) (21,30.745) (22,29.912) (23,31.855) (24,32.039) (25,34.353) (26,33.242) (27,32.872) (28,33.520) (29,32.872) (30,34.075) (31,33.613) (32,31.299) (33,32.872) (34,33.335) (35,31.022) (36,32.039) (37,30.096) (38,31.855) (39,32.595) (40,28.246) (41,31.485) (42,31.115) (43,27.043) (44,28.709) (45,30.929) (46,31.022) (47,30.559) (48,30.466) (49,28.246) (50,26.858) (51,29.172) (52,28.153) (53,29.912) (54,30.466) (55,28.246) (56,27.229) (57,25.655) (58,29.079) (59,30.004) (60,31.022) (61,31.485) (62,31.485) (63,30.374)};
\addplot[pxNbr, line width=0.8pt] coordinates {(0,30.605) (1,33.130) (2,32.457) (3,33.046) (4,29.678) (5,31.699) (6,32.541) (7,29.847) (8,30.183) (9,32.036) (10,32.541) (11,30.689) (12,30.520) (13,31.278) (14,30.183) (15,27.827) (16,35.319) (17,28.668) (18,28.247) (19,30.520) (20,28.921) (21,25.975) (22,29.426) (23,35.235) (24,37.676) (25,36.245) (26,34.056) (27,33.551) (28,31.951) (29,30.605) (30,32.709) (31,34.140) (32,35.235) (33,33.130) (34,35.067) (35,33.719) (36,30.857) (37,32.793) (38,32.036) (39,29.089) (40,28.752) (41,31.447) (42,31.530) (43,28.331) (44,31.278) (45,30.016) (46,28.247) (47,28.583) (48,29.341) (49,30.183) (50,27.321) (51,29.510) (52,28.921) (53,29.762) (54,27.573) (55,27.827) (56,28.752) (57,28.921) (58,28.163) (59,27.237) (60,26.395) (61,29.258) (62,28.247) (63,27.573)};
\addplot[pxNbr, line width=0.8pt] coordinates {(0,24.208) (1,23.673) (2,24.656) (3,24.923) (4,25.997) (5,25.908) (6,24.298) (7,24.030) (8,23.583) (9,22.868) (10,23.315) (11,23.851) (12,24.030) (13,23.225) (14,23.493) (15,24.119) (16,23.046) (17,23.583) (18,22.956) (19,25.192) (20,25.818) (21,25.192) (22,25.728) (23,24.477) (24,25.550) (25,26.623) (26,28.410) (27,27.874) (28,27.427) (29,27.874) (30,27.874) (31,28.410) (32,28.232) (33,28.590) (34,27.070) (35,27.338) (36,26.265) (37,25.997) (38,26.801) (39,26.533) (40,26.623) (41,26.443) (42,25.460) (43,27.785) (44,29.573) (45,29.573) (46,29.037) (47,27.606) (48,26.443) (49,27.695) (50,28.679) (51,28.768) (52,25.371) (53,28.500) (54,27.606) (55,25.371) (56,28.053) (57,26.890) (58,25.192) (59,26.801) (60,27.963) (61,24.566) (62,24.656) (63,25.908)};
\addplot[pxNbr, line width=0.8pt] coordinates {(0,30.096) (1,27.546) (2,28.638) (3,31.189) (4,31.189) (5,32.428) (6,32.354) (7,29.659) (8,28.274) (9,28.055) (10,28.202) (11,28.420) (12,30.096) (13,28.274) (14,26.744) (15,29.950) (16,30.168) (17,30.241) (18,32.501) (19,30.460) (20,28.711) (21,29.732) (22,29.294) (23,30.533) (24,32.282) (25,32.282) (26,28.420) (27,30.388) (28,29.513) (29,27.400) (30,26.234) (31,28.274) (32,28.711) (33,27.910) (34,32.063) (35,31.480) (36,32.574) (37,34.177) (38,33.594) (39,32.865) (40,31.189) (41,31.262) (42,27.473) (43,25.578) (44,25.068) (45,23.392) (46,25.360) (47,25.360) (48,24.485) (49,24.048) (50,23.392) (51,20.914) (52,21.351) (53,24.266) (54,23.975) (55,22.080) (56,22.371) (57,24.631) (58,22.080) (59,23.027) (60,24.558) (61,22.663) (62,22.591) (63,24.121)};
\addplot[pxNbr, line width=0.8pt] coordinates {(0,21.143) (1,20.670) (2,19.772) (3,20.339) (4,20.576) (5,20.718) (6,21.285) (7,21.285) (8,21.711) (9,22.420) (10,22.373) (11,22.657) (12,22.988) (13,23.177) (14,23.366) (15,23.272) (16,23.414) (17,23.319) (18,23.508) (19,23.981) (20,24.123) (21,24.359) (22,24.170) (23,24.359) (24,24.596) (25,24.312) (26,24.076) (27,23.981) (28,24.218) (29,24.218) (30,23.887) (31,24.028) (32,23.839) (33,23.650) (34,23.414) (35,23.650) (36,23.887) (37,24.076) (38,24.643) (39,24.691) (40,24.265) (41,24.028) (42,24.832) (43,24.927) (44,24.738) (45,24.738) (46,24.596) (47,24.974) (48,25.589) (49,25.163) (50,25.022) (51,25.211) (52,26.015) (53,26.346) (54,25.636) (55,26.251) (56,24.927) (57,24.643) (58,23.981) (59,24.076) (60,24.501) (61,24.076) (62,24.265) (63,23.839)};
\addplot[pxBest, line width=1.5pt, dashed] coordinates {(0,17.808) (1,20.882) (2,21.517) (3,19.398) (4,23.425) (5,20.563) (6,21.836) (7,21.941) (8,22.154) (9,23.955) (10,23.743) (11,24.166) (12,22.154) (13,21.623) (14,25.757) (15,26.074) (16,22.471) (17,19.928) (18,23.319) (19,24.803) (20,24.485) (21,17.703) (22,21.093) (23,20.352) (24,19.292) (25,21.941) (26,21.306) (27,22.577) (28,22.260) (29,23.001) (30,25.120) (31,25.651) (32,24.697) (33,24.590) (34,20.352) (35,21.836) (36,20.458) (37,15.688) (38,17.384) (39,19.398) (40,13.569) (41,14.841) (42,14.311) (43,14.523) (44,15.476) (45,16.112) (46,11.343) (47,13.993) (48,16.536) (49,17.066) (50,15.476) (51,14.947) (52,15.688) (53,13.993) (54,11.979) (55,10.601) (56,13.463) (57,13.145) (58,12.509) (59,11.025) (60,14.204) (61,12.720) (62,13.674) (63,13.674)};
\addplot[pxGT, line width=1.5pt] coordinates {(0,19.853) (1,20.272) (2,21.026) (3,21.193) (4,21.612) (5,20.942) (6,19.769) (7,21.528) (8,19.937) (9,20.858) (10,20.356) (11,21.863) (12,21.863) (13,21.528) (14,21.193) (15,22.199) (16,22.450) (17,22.701) (18,24.544) (19,24.293) (20,23.120) (21,25.214) (22,25.884) (23,24.376) (24,24.628) (25,26.136) (26,26.219) (27,26.554) (28,23.036) (29,24.293) (30,22.115) (31,22.785) (32,21.780) (33,21.528) (34,21.612) (35,21.863) (36,23.539) (37,22.617) (38,21.863) (39,21.361) (40,19.602) (41,19.769) (42,20.021) (43,19.183) (44,17.675) (45,16.502) (46,16.083) (47,16.837) (48,16.921) (49,14.241) (50,14.157) (51,13.570) (52,13.570) (53,13.403) (54,13.235) (55,13.319) (56,14.911) (57,12.649) (58,13.152) (59,13.403) (60,12.398) (61,10.974) (62,12.398) (63,13.403)};
\addplot[pxBolt, line width=1.7pt] coordinates {(0,21.108) (1,21.641) (2,21.962) (3,22.254) (4,22.268) (5,22.534) (6,22.671) (7,23.017) (8,23.474) (9,23.935) (10,24.345) (11,24.796) (12,25.078) (13,25.430) (14,25.448) (15,25.581) (16,25.861) (17,26.299) (18,26.654) (19,27.063) (20,27.243) (21,27.567) (22,27.991) (23,28.124) (24,28.149) (25,28.382) (26,28.551) (27,28.663) (28,28.288) (29,28.253) (30,28.246) (31,28.517) (32,28.872) (33,29.280) (34,29.522) (35,29.669) (36,29.642) (37,29.577) (38,29.060) (39,28.481) (40,28.021) (41,27.579) (42,27.050) (43,26.701) (44,26.236) (45,25.914) (46,25.602) (47,25.201) (48,24.925) (49,24.808) (50,24.581) (51,24.647) (52,24.418) (53,24.431) (54,24.376) (55,24.505) (56,24.620) (57,24.854) (58,24.923) (59,24.856) (60,24.565) (61,24.639) (62,24.380) (63,23.857)};
\addplot[pxRag, line width=1.7pt] coordinates {(0,21.738) (1,22.419) (2,22.774) (3,23.261) (4,23.392) (5,23.722) (6,24.011) (7,24.425) (8,25.027) (9,25.542) (10,25.906) (11,26.508) (12,26.938) (13,27.330) (14,27.444) (15,27.544) (16,28.043) (17,28.451) (18,28.775) (19,29.290) (20,29.541) (21,29.881) (22,30.220) (23,30.498) (24,30.701) (25,30.902) (26,31.153) (27,31.396) (28,31.124) (29,31.267) (30,31.320) (31,31.556) (32,31.967) (33,32.443) (34,32.717) (35,32.920) (36,32.938) (37,33.034) (38,32.491) (39,31.945) (40,31.507) (41,31.026) (42,30.556) (43,30.181) (44,29.731) (45,29.373) (46,28.965) (47,28.532) (48,28.271) (49,28.104) (50,27.926) (51,28.006) (52,27.766) (53,27.745) (54,27.738) (55,27.895) (56,27.946) (57,28.128) (58,28.116) (59,27.885) (60,27.604) (61,27.683) (62,27.383) (63,26.724)};
\draw[pxInk, -stealth, line width=2.2pt] (rel axis cs:0,0) -- (rel axis cs:1,0);
\end{axis}
\pgfresetboundingbox
\path[use as bounding box] (-1.75cm,0) rectangle (16.3cm,4.3cm);
\end{tikzpicture}}%
\ifdefined\prexpFs\else\newlength\prexpFs\fi\setlength\prexpFs{0.02271\textwidth}%
\ifdefined\prexpVa\else\newlength\prexpVa\fi\ifdefined\prexpVb\else\newlength\prexpVb\fi\pgfmathsetmacro\prexpRa{\ht0/\wd0}%
\pgfmathsetlength\prexpVa{0pt}\pgfmathsetlength\prexpVb{0.03*\prexpRa*\textwidth}%
\par%
\noindent\makebox[\textwidth]{\scalebox{2.0}{\begin{tikzpicture}[baseline=(current bounding box.south)]
\fill[pxGreen] (0,0) rectangle (0.36,0.14);
\node[anchor=west, inner sep=2pt, font=\tiny] at (0.36,0.07) {w/ RAG worse};
\fill[pxRed] (1.9,0) rectangle (2.26,0.14);
\node[anchor=west, inner sep=2pt, font=\tiny] at (2.26,0.07) {w/ RAG better};
\node[anchor=west, inner sep=2pt, font=\tiny] at (3.8,0.07) {\prexpDelta{} / \prexpdelta{} = MSE / MAE reduction};
\end{tikzpicture}}}\par\vspace{\prexpVa}%
\noindent\resizebox{\textwidth}{!}{\box0}\par\vspace{\prexpVb}%
\noindent\makebox[\textwidth][l]{\fontsize{\the\prexpFs}{\the\dimexpr1.2\prexpFs\relax}\selectfont\bfseries \hspace*{0.2436\textwidth}\makebox[0pt]{(a) TS-RAG MSE / MAE changes}\hspace{0.5127\textwidth}\makebox[0pt]{(b) Cross-RAG MSE / MAE changes}}\par\vspace{22pt}%
\noindent\makebox[\textwidth]{{\fontsize{\the\prexpFs}{\the\dimexpr1.2\prexpFs\relax}\selectfont\bfseries\begin{tabular}{@{}l@{\quad}l@{\quad}l@{}}\tikz[baseline=-0.6ex]\draw[pxHist, line width=0.8pt] (0,0) -- (0.45,0);\,history & \tikz[baseline=-0.6ex]\draw[pxGT, line width=1pt] (0,0) -- (0.45,0);\,ground truth & \tikz[baseline=-0.6ex]\draw[pxBolt, line width=1.1pt] (0,0) -- (0.45,0);\,Chronos-Bolt\\\tikz[baseline=-0.6ex]\draw[pxRag, line width=1.1pt] (0,0) -- (0.45,0);\,TS-RAG & \tikz[baseline=-0.6ex]\draw[pxBest, line width=1pt, dashed] (0,0) -- (0.45,0);\,best retrieved neighbor & \tikz[baseline=-0.6ex]\draw[pxNbr, line width=0.6pt] (0,0) -- (0.45,0);\,other retrieved neighbors\end{tabular}}}\par\vspace{4pt}%
\noindent\resizebox{\textwidth}{!}{\box2}\par\vspace{14pt}%
\noindent\makebox[\textwidth][l]{\fontsize{\the\prexpFs}{\the\dimexpr1.2\prexpFs\relax}\selectfont\bfseries \hspace*{0.2964\textwidth}\makebox[0pt]{(c) Neighbors used well}\hspace{0.5042\textwidth}\makebox[0pt]{(d) Neighbors not exploited}}\par%
\end{minipage}}\par%
\endgroup%

%% file: related_work.tex
\section{Related Work}

\paragraph{Time Series Foundation Models.}
Time Series Foundation Models (TSFMs) learn transferable temporal
patterns from large, heterogeneous corpora, supporting forecasting
with limited or no task-specific adaptation
\citep{liang2024foundation}. Chronos \citep{ansari2024chronos}
discretizes numerical observations into tokens and adopts
language modeling objectives, whereas TimesFM
\citep{das2024decoder} operates on continuous-valued patches
with a decoder-only architecture. Moirai \citep{woo2024unified}
addresses heterogeneity in frequency, dimensionality, and
marginal distributions. Other developments include
mixture-of-experts modeling in Moirai-MoE
\citep{liu2025moirai}, masked modeling in MOMENT
\citep{goswami2024moment}, generative probabilistic forecasting
in Sundial \citep{liu2025sundial}, and scalable pretraining
in Timer-S1 \citep{liu2026timer}.

\paragraph{Retrieval-Augmented Forecasting.}
Retrieval augmentation supplies forecasters with historical
examples relevant to the current input. RAF
\citep{tireretrieval} studies retrieval augmentation for
pretrained forecasters, while RAFT \citep{han2025retrieval}
incorporates historical sequences and their continuations
into a multi-resolution forecasting framework.
TimeRAF \citep{zhang2025timeraf} supervises candidate ranking
with forecasting feedback and integrates retrieved
representations through channel prompting.
TS-RAG \citep{ningts} adaptively combines query and retrieval
information, while Cross-RAG \citep{lee2026cross} uses
cross-attention between the query and retrieved representations
to incorporate relevant information and reduce interference
from irrelevant neighbors.

\paragraph{Forecast Refinement and Example Reweighting.}
Retrieved evidence also supports refinement in prediction
space. PFRP \citep{du2026predicting} adaptively combines
predictions from a global historical memory with local
model forecasts. RATL \citep{he2026ratl} freezes the base
forecaster and learns to combine retrieved historical
residuals, training its router with prediction loss
and oracle supervision derived from candidate correction
errors. Both methods are developed and evaluated under
dataset-specific supervised training settings.
Example reweighting provides a complementary approach
to allocating training effort. Focal loss downweights
well-classified instances \citep{lin2017focal}, while
meta-learning methods learn example weights using
validation objectives \citep{ren2018learning}. 

In contrast, our method combines prediction-space refinement
with opportunity-aware reweighting across source-training
queries for zero-shot transfer.

%% file: our_method.tex
\section{Our Method}
\label{sec:method}

We introduce a \textbf{R}etrieval \textbf{O}pportunity-\textbf{A}ware
\textbf{R}efinement framework (\textbf{ROAR}) that refines
a base forecast using retrieved historical continuations.
It allocates additional training emphasis across queries based on
base-forecast difficulty and the relative improvement offered
by retrieved candidates.
Fig.~\ref{fig:framework} provides an overview of ROAR's
prediction framework and two-stage training procedure.
Stage I keeps the base forecaster fixed and learns candidate
aggregation and correction strength using opportunity-aware weighting.
Stage II retains the learned aggregation and jointly calibrates
the trainable forecasting module and gate, using a reference
penalty to anchor their combined prediction to the Stage-I
refined output.

\input{fig-framework}

\paragraph{Problem Setup.}
\label{sec:problem_setup}
For a single channel, let $\mathbf{x}_i\in\mathbb{R}^{L}$ denote
an observed context and $\mathbf{y}_i\in\mathbb{R}^{H}$ its future.
Each query retrieves $K$ historical examples
$\mathcal{N}_i=\{(\mathbf{x}_{ij}^{r},\mathbf{y}_{ij}^{r})\}_{j=1}^{K}$,
consisting of contexts and their observed continuations.
Retrieval uses only continuations available at the forecast origin
and excludes each source-training query's own target segment.
The retrieval-augmented base forecaster combines a frozen
foundation backbone with parameters $\theta$ and a trainable
forecasting module with parameters $\eta$.
We denote its point forecast by
\begin{equation}
\label{eq:base_forecaster}
\widehat{\mathbf{y}}_i^0(\eta)
=f_{\theta,\eta}(\mathbf{x}_i,\mathcal{N}_i).
\end{equation}
Let $\eta^R$ denote the forecasting module's source-trained
initialization before refinement and
$\widehat{\mathbf{y}}_i^{0,R}
=\widehat{\mathbf{y}}_i^0(\eta^R)$ its point forecast.
The parameters $\theta$ remain frozen throughout.
All trainable modules are optimized on source data and applied
to target datasets without parameter updates. We denote the per-query forecasting loss by $\ell_i$.
Its stage-specific form is given in each stage below.

\subsection{Overall Framework: Refinement with Historical Candidates}
\label{sec:refinement_framework}
We first describe the prediction framework shared by both
training stages.
ROAR aligns retrieved continuations to the query and aggregates
the resulting candidates into a proposal.
A learned gate determines how far to adjust the base forecast
toward this proposal.
The two-stage optimization of this framework is described in
Sections~\ref{sec:opportunity_learning}
and~\ref{sec:joint_calibration}.

\paragraph{Context-Aligned Historical Candidates.}
\label{sec:candidate_construction}
Retrieved continuations can differ from the query in level and
scale. We construct two aligned versions of each continuation,
using the contextual mean and the latest observation as alternative
reference levels. Let $\mu_i$ and $s_i>0$ denote the temporal mean and
stabilized standard deviation of $\mathbf{x}_i$.
The corresponding statistics of the retrieved contexts are
$\mu_{ij}^{r}$ and $s_{ij}^{r}>0$
(where $i$ indexes queries and $j$ indexes neighbors
retrieved for each query). The candidates are
\begin{equation}
\label{eq:candidate_alignment}
\mathbf{Z}_{ij}^{\mathrm{lev}}
=\mu_i+s_i
\frac{\mathbf{y}_{ij}^{r}-\mu_{ij}^{r}}{s_{ij}^{r}},
\qquad
\mathbf{Z}_{ij}^{\mathrm{anc}}
=x_{i,L}+s_i
\frac{\mathbf{y}_{ij}^{r}-x_{ij,L}^{r}}{s_{ij}^{r}},
\end{equation}
where $x_{i,L}$ and $x_{ij,L}^{r}$ are the last observed values
of the query and retrieved contexts. Level alignment transfers
deviations from the historical mean, while anchor alignment
transfers changes relative to the last historical observation.
We retain both aligned versions of each retrieved continuation,
yielding $J=2K$ candidates indexed by $k$:
$\mathcal{Z}_i=\{\mathbf{Z}_{ik}\}_{k=1}^{J}$.
Their values depend only on observed contexts and historical
continuations.

\paragraph{Candidate Aggregation and Correction Gating.}
\label{sec:forecast_refinement}
A candidate scorer determines the contribution of each historical
forecast, while a separate gate controls correction strength.
For candidate $j$ of query $i$, we construct a feature vector
$\mathbf{e}_{ij}^R$ from its retrieval distance, normalized
deviation from the reference base forecast, and alignment type.
The gate uses query-level features
$\boldsymbol{\phi}_i^R$ summarizing reference-forecast uncertainty,
retrieval distance, and candidate dispersion. These features
are computed from the frozen reference branch and retrieved history.
The aggregation weights and correction strength are
\begin{equation}
\label{eq:refinement_weights}
p_{ij}
=\frac{\exp\!\left(s_{\psi_c}(\mathbf{e}_{ij}^{R})\right)}
{\sum_{k=1}^{J}\exp\!\left(s_{\psi_c}(\mathbf{e}_{ik}^{R})\right)},
\qquad
u_i=\sigma\!\left(b_{\psi_u}(\boldsymbol{\phi}_i^{R})\right),
\end{equation}
where $\psi_c$ and $\psi_u$ parameterize the scorer and gate,
respectively, and $\sigma$ is the sigmoid function.
The aggregated proposal is
$\mathbf{r}_i=\sum_{k=1}^{J}p_{ik}\mathbf{Z}_{ik}$. We adjust the base forecast toward this proposal:
\begin{equation}
\label{eq:roar_prediction}
\boxed{
\widehat{\mathbf{y}}_i
=\widehat{\mathbf{y}}_i^0(\eta)
+u_i\left(\mathbf{r}_i-\widehat{\mathbf{y}}_i^0(\eta)\right).
}
\end{equation}
The resulting forecast is a convex combination of the base
prediction and historical candidates, with weights $1-u_i$
and $u_ip_{ij}$.

In our quantile-output implementation, the base forecaster
produces $\mathbf{q}_i^{0,\tau}(\eta)$ at quantile level $\tau$.
We use the median ($\tau=0.5$) as the base point forecast
$\widehat{\mathbf{y}}_i^0(\eta)$ and apply the same refinement
correction to every quantile:
\begin{equation}
\label{eq:shared_quantile_correction}
\widehat{\mathbf{q}}_i^\tau
=\mathbf{q}_i^{0,\tau}(\eta)
+\widehat{\mathbf{y}}_i-\widehat{\mathbf{y}}_i^0(\eta).
\end{equation}

Candidate aggregation operates within each query.
The opportunity-aware objective below determines the additional
training emphasis assigned across queries.

\subsection{Stage \texorpdfstring{$\mathrm{I}$}{I}: Opportunity-Aware Learning}
\label{sec:opportunity_learning}
A large base error indicates room for improvement, but does not
establish that an individual retrieved candidate offers a better
forecast. Stage I therefore learns the scorer and gate with
$\eta=\eta^R$ fixed, assigning additional training emphasis to
queries that are both difficult and supported by beneficial
historical candidates.

\paragraph{Query-Level Improvement Opportunity.}
Let $\Omega_i$ be the set of valid forecasting entries for query
$i$ and $n_i=\max\{|\Omega_i|,1\}$.
For a point-forecast vector $\mathbf{v}=(v_1,\ldots,v_H)^\top$,
define its error as
\begin{equation}
\label{eq:forecast_loss}
\delta_i(\mathbf{v})
=\frac{1}{n_i}
\sum_{t\in\Omega_i}(v_t-y_{i,t})^2.
\end{equation}
Write $\delta_i^0=\delta_i(\widehat{\mathbf{y}}_i^{0,R})$
for the reference base error and
$\delta_i^{r,\min}=\min_{1\leq k\leq J}\delta_i(\mathbf{Z}_{ik})$
for the best individual candidate error.
A threshold $t_h$ on the normalized reference error identifies
difficult queries:
$h_i=\mathbf{1}[\delta_i^0/s_i^2>t_h]$.
The opportunity score and training weight are
\begin{equation}
\label{eq:opportunity_score}
g_i
=\frac{[\delta_i^0-\delta_i^{r,\min}]_+}
{\delta_i^0+\epsilon s_i^2},
\qquad
w_i=s_i^2(1+\beta h_i g_i),
\end{equation}
where $[a]_+=\max(a,0)$, $\epsilon>0$, and $\beta\geq0$.
The score $g_i$ measures the relative error reduction available
from the best individual candidate and is zero when no candidate
improves on the base forecast.
The factor $s_i^2$ sets the baseline weighting of the raw-scale
forecasting loss, and
$1+\beta h_i g_i$ supplies the opportunity-dependent multiplier.

\paragraph{Learning the Refined Forecast.}
For a source-training mini-batch $\mathcal{B}$, we optimize the
opportunity-weighted forecasting loss with auxiliary gate supervision:
\begin{equation}
\label{eq:roar_objective}
\boxed{
\min_{\psi_c,\psi_u}\mathcal{L}_{\mathrm{init}}
=\frac{\sum_{i\in\mathcal{B}}
\operatorname{sg}(w_i)
\ell_i}
{\sum_{i\in\mathcal{B}}\operatorname{sg}(w_i)}
+\frac{\lambda_{\mathrm{gate}}}{|\mathcal{B}|}
\sum_{i\in\mathcal{B}}
\operatorname{BCE}\!\left(u_i,\operatorname{sg}(z_i)\right).
}
\end{equation}
In our implementation, $\ell_i$ is twice the mean pinball loss
over quantile levels and valid forecast entries, evaluated on
the refined quantiles in Eq.~\ref{eq:shared_quantile_correction}.
Here, $\operatorname{BCE}$ is binary cross-entropy,
$\lambda_{\mathrm{gate}}\geq0$ controls the auxiliary term,
and $\operatorname{sg}$ denotes stop-gradient.
The normalized weight $w_i/\sum_{k\in\mathcal{B}}w_k$ determines
query $i$'s forecasting-loss contribution to learning both
candidate aggregation and correction strength.
The auxiliary target compares the current aggregated proposal
($u_i=1$) with the reference base forecast ($u_i=0$):
$z_i=\mathbf{1}[\delta_i(\mathbf{r}_i)<\delta_i^0]$.

The label supervises a preference between the two endpoints.
The combined objective determines the continuous correction
strength.
The score $g_i$ instead evaluates the best individual candidate.
Opportunity weights and preference labels are constructed from
source-training outcomes and are used only during training.

\subsection{Stage \texorpdfstring{$\mathrm{II}$}{II}: Reference-Guided Joint Calibration}
\label{sec:joint_calibration}
Stage I fixes the base forecast to maintain a stable criterion
for candidate opportunity while learning the refiner.
Stage II jointly adapts the forecasting module and gate under
a shared prediction objective while retaining the learned
retrieval proposal. The Stage-I refined prediction provides
a common reference for these updates.

The Stage-I parameters $(\eta^R,\psi_c^R,\psi_u^R)$ define
the reference proposal $\mathbf{r}_i^R$, gate $u_i^R$,
and refined prediction
\begin{equation}
\label{eq:reference_prediction}
\widehat{\mathbf{y}}_i^R
=(1-u_i^R)\widehat{\mathbf{y}}_i^{0,R}
+u_i^R\mathbf{r}_i^R.
\end{equation}
We freeze the reference model and scorer $\psi_c^R$.
The aggregation weights $p_{ij}^R$, proposal $\mathbf{r}_i^R$,
and feature vectors $\boldsymbol{\phi}_i^R$ therefore remain
fixed for each query throughout calibration.
The gate parameters $\psi_u$ remain trainable on these
query-specific features.
Starting from $(\eta,\psi_u)=(\eta^R,\psi_u^R)$,
we jointly update $\eta$ and $\psi_u$, yielding
\begin{equation}
\label{eq:calibrated_prediction}
\widehat{\mathbf{y}}_i
=(1-u_i)\widehat{\mathbf{y}}_i^0(\eta)
+u_i\mathbf{r}_i^R.
\end{equation}
The objective combines a uniformly averaged forecasting loss
with a squared penalty on changes to the reference prediction:
\begin{equation}
\label{eq:roar_calibration}
\boxed{
\min_{\eta,\psi_u}\mathcal{L}_{\mathrm{cal}}
=\frac{1}{|\mathcal{B}|}\sum_{i\in\mathcal{B}}
\left[
\ell_i
+\lambda_{\mathrm{ref}}\,
\frac{\left\|\widehat{\mathbf{y}}_i
-\operatorname{sg}(\widehat{\mathbf{y}}_i^R)
\right\|_{\Omega_i}^{2}}{n_i}
\right].
}
\end{equation}
Here, $\ell_i=\delta_i(\widehat{\mathbf{y}}_i)$ is point-forecast
MSE on the original scale. The norm is restricted to $\Omega_i$,
and $\lambda_{\mathrm{ref}}\geq0$ controls the reference penalty.
Thus, we jointly calibrate the forecasting module and gate
while regularizing their combined prediction toward the
Stage-I refined output.

\subsection{Theoretical Analysis}
\label{sec:refinement_analysis}
We analyze how candidate opportunity translates into forecasting
gains and guides training allocation across queries.
We then characterize the additional corrections available through
joint calibration under a local linearization on source data.

\paragraph{From Candidate Opportunity to Forecast Improvement.}
For Stage I with the base forecaster fixed, define the weighted
candidate regret
\begin{equation}
\label{eq:aggregation_regret}
R_i^{\mathrm{agg}}
=\sum_{k=1}^{J}p_{ik}\delta_i(\mathbf{Z}_{ik})
-\delta_i^{r,\min}\geq0.
\end{equation}
For any query with $g_i>0$, the point-forecast MSE reduction
admits the exact decomposition
\begin{equation}
\label{eq:realized_refinement_gain}
\delta_i^0-\delta_i(\widehat{\mathbf{y}}_i)
=u_i\big[(\delta_i^0+\epsilon s_i^2)g_i-R_i^{\mathrm{agg}}\big]
+\frac{u_i}{n_i}\sum_{k=1}^{J}p_{ik}\|\mathbf{Z}_{ik}-\mathbf{r}_i\|_{\Omega_i}^{2}
+\frac{u_i(1-u_i)}{n_i}\|\mathbf{r}_i-\widehat{\mathbf{y}}_i^{0,R}\|_{\Omega_i}^{2}.
\end{equation}
The two squared-norm terms quantify nonnegative reductions
from convex mixing relative to the corresponding weighted
component errors.
Consequently, $u_i>0$ and
$R_i^{\mathrm{agg}}/(\delta_i^0+\epsilon s_i^2)<g_i$
are sufficient for a strict improvement over the base forecast.
The proof is given in Appendix~\ref{app:opportunity_realization}.

\paragraph{Training Allocation Across Queries.}
For a source mini-batch $\mathcal{B}$, define the baseline
scale weights $q_i=s_i^2/\sum_{k\in\mathcal{B}}s_k^2$,
the difficulty-filtered opportunity $a_i=h_i g_i$,
and its weighted mean $m=\sum_{i\in\mathcal{B}}q_i a_i$.
When $m>0$, let $q_i^{\mathrm{opp}}=q_i a_i/m$.
The forecasting term in Eq.~\ref{eq:roar_objective} admits
the exact decomposition
\begin{equation}
\label{eq:opportunity_allocation_summary}
\frac{\sum_{i\in\mathcal{B}}w_i\ell_i}{\sum_{i\in\mathcal{B}}w_i}
=(1-\alpha)\sum_{i\in\mathcal{B}}q_i\ell_i
+\alpha\sum_{i\in\mathcal{B}}q_i^{\mathrm{opp}}\ell_i,
\qquad \alpha=\frac{\beta m}{1+\beta m}.
\end{equation}
Thus opportunity weighting combines the baseline loss with
an additional component supported on difficult queries whose
candidates improve on the base forecast.
All queries retain positive forecasting-loss weights.
When $m=0$, the forecasting term reduces to the scale-weighted
baseline. The derivation is given in
Appendix~\ref{app:opportunity_allocation}.

\paragraph{Additional Corrections from Joint Calibration.}
For squared-error calibration on a fixed source set, assume
the prediction map is differentiable at the Stage-I reference.
Let $\mathbf{e}$ be the stacked difference between the targets
and the Stage-I refined predictions, restricted to valid entries
and scaled so that $\|\mathbf{e}\|^2$ equals the source-averaged MSE.
Let $\mathcal{V}_\eta$ and $\mathcal{V}_u$ be the ranges of the
Jacobians of the correspondingly scaled prediction map with
respect to the forecasting-module and gate parameters,
evaluated at the reference.
Linearized joint updates span
$\mathcal{V}_{\mathrm{joint}}=\mathcal{V}_\eta+\mathcal{V}_u$.
For an output correction $\boldsymbol{\Delta}\in\mathcal{V}$,
the linearized objective is
$\|\mathbf{e}-\boldsymbol{\Delta}\|^2
+\lambda\|\boldsymbol{\Delta}\|^2$,
where $\lambda=\lambda_{\mathrm{ref}}\geq0$.
Writing $\mathbf{P}_{\mathcal{V}}$ for orthogonal projection
onto $\mathcal{V}$, its optimal correction and the resulting
forecasting MSE reduction are
\begin{equation}
\label{eq:calibration_optimum_summary}
\boldsymbol{\Delta}_{\mathcal{V}}^\star
=\frac{\mathbf{P}_{\mathcal{V}}\mathbf{e}}{1+\lambda},
\qquad
G(\mathcal{V})
=\|\mathbf{e}\|^2-\|\mathbf{e}-\boldsymbol{\Delta}_{\mathcal{V}}^\star\|^2
=\frac{1+2\lambda}{(1+\lambda)^2}\|\mathbf{P}_{\mathcal{V}}\mathbf{e}\|^2.
\end{equation}
Comparing the optima of the linearized objectives at the same
reference and penalty strength, for either
$\mathcal{U}\in\{\mathcal{V}_\eta,\mathcal{V}_u\}$,
\begin{equation}
\label{eq:joint_calibration_gain_summary}
G(\mathcal{V}_{\mathrm{joint}})-G(\mathcal{U})
=\frac{1+2\lambda}{(1+\lambda)^2}
\left\|(\mathbf{P}_{\mathcal{V}_{\mathrm{joint}}}
-\mathbf{P}_{\mathcal{U}})\mathbf{e}\right\|^2
\geq0.
\end{equation}
The gain over $\mathcal{U}$ is strict precisely when the
reference residual has a nonzero component in
$\mathcal{V}_{\mathrm{joint}}\cap\mathcal{U}^{\perp}$.
The derivation and an error analysis for nonlinear updates
are provided in Appendices~\ref{app:joint_calibration}
and~\ref{app:calibration_errors}.

%% file: fig-framework.tex
\usetikzlibrary{arrows.meta,calc,shapes.geometric}

\definecolor{cBase}{RGB}{52,101,164}
\definecolor{cRet}{RGB}{214,120,26}
\definecolor{cGate}{RGB}{34,135,100}
\definecolor{cOut}{RGB}{186,48,84}
\definecolor{cSone}{RGB}{112,74,160}
\definecolor{cStwo}{RGB}{0,128,150}

\begin{figure}[t]
\centering
\resizebox{\linewidth}{!}{%
\begin{tikzpicture}[
  font=\small,
  >={Stealth[length=2mm]},
  flow/.style={->,line width=0.75pt,draw=black!70},
  thin/.style={->,line width=0.6pt,draw=black!55},
  card/.style={draw=black!35,fill=white,line width=0.5pt},
  op/.style={circle,draw=black!75,fill=white,line width=0.8pt,inner sep=0pt,minimum size=0.5cm},
  lab/.style={font=\small,align=center},
  cnote/.style={font=\footnotesize\itshape,align=center},
  snote/.style={font=\scriptsize\itshape,text=black!65,align=center},
  group/.style={draw=black!35,densely dashed,rounded corners=5pt,line width=0.6pt},
  gtitle/.style={font=\small\bfseries,text=black!85},
  gap/.style={preaction={draw=white,line width=3pt}},
]
\newcommand{\shadowcard}[4]{%
  \fill[black!12] ($(#1,#2)+(0.06,-0.06)$) rectangle ++(#3,#4);
  \draw[card] (#1,#2) rectangle ++(#3,#4);}
\newcommand{\mlp}[3]{%
  \begin{scope}[shift={(#1,#2)}]
    \foreach \a in {-0.22,0,0.22} \foreach \b in {-0.3,-0.1,0.1,0.3}
      \draw[black!30,line width=0.3pt] (-0.35,\a) -- (0,\b);
    \foreach \b in {-0.3,-0.1,0.1,0.3} \foreach \c in {-0.11,0.11}
      \draw[black!30,line width=0.3pt] (0,\b) -- (0.35,\c);
    \foreach \a in {-0.22,0,0.22} \fill[#3!55] (-0.35,\a) circle (0.065);
    \foreach \b in {-0.3,-0.1,0.1,0.3} \fill[#3!80] (0,\b) circle (0.065);
    \foreach \c in {-0.11,0.11} \fill[#3] (0.35,\c) circle (0.065);
  \end{scope}}
\newcommand{\fire}[1]{\tikz[baseline=-0.5ex,x=0.12em,y=0.12em]{\fill[#1] (0,0) .. controls (-4,1) and (-3,4) .. (-1,5) .. controls (-1,3) and (0,3) .. (0,3) .. controls (2,5) and (1,7) .. (1,8) .. controls (5,4) and (4,1) .. (2,0) -- cycle;}}
\newcommand{\frz}{\tikz[baseline=-0.5ex,x=0.12em,y=0.12em]{\foreach \a in {0,60,...,300}{\draw[cBase!80,line width=0.45pt,rotate=\a] (0,0)--(0,4) (-1,2.5)--(0,1.5)--(1,2.5);}}}

\draw[group] (2.45,2.75) rectangle (10.0,6.35);
\node[gtitle,anchor=north west] at (2.55,6.3) {(a) Retrieval-augmented base forecaster};
\draw[group] (2.45,-0.25) rectangle (10.0,2.6);
\node[gtitle,anchor=north east] at (9.9,2.55) {(b) Complementary candidates};
\draw[group] (10.3,-0.25) rectangle (17.1,6.35);
\node[gtitle,anchor=north west] at (10.4,6.3) {(c) Input-dependent refinement};

\node[anchor=north west,font=\footnotesize,align=left,inner sep=1pt] at (17.4,6.35)
  {\frz\ frozen\\[2pt] \fire{cSone}\ trained in Stage I\\[2pt] \fire{cStwo}\ trained in Stage II};

\shadowcard{0}{3.35}{2.1}{1.3}
\fill[black!4] (1.45,3.4) rectangle (2.05,4.6);
\draw[cBase,line width=1.1pt] plot[smooth] coordinates
  {(0.1,3.75)(0.35,4.15)(0.6,3.7)(0.85,4.05)(1.1,4.35)(1.3,3.95)(1.45,4.07)};
\node[black!45] at (1.75,4.0) {?};
\node[lab] at (1.05,3.05) {Query $\mathbf{x}_i$};

\node[cylinder,shape border rotate=90,aspect=0.25,draw=black!55,fill=black!5,
      minimum width=1.2cm,minimum height=1.15cm,line width=0.6pt] (db) at (1.05,1.1) {};
\foreach \h in {0.85,1.25} \draw[black!35] (0.45,\h) arc (180:360:0.6 and 0.11);
\node[lab] at (1.05,0.15) {History};
\draw[flow] (1.05,2.8) -- (db.top);
\node[font=\footnotesize\itshape,align=right,anchor=east] at (0.98,2.2) {retrieve\\[-2pt] top-$K$};

\foreach \k in {2,1,0}{
  \pgfmathsetmacro{\ox}{0.1*\k} \pgfmathsetmacro{\oy}{0.1*\k}
  \shadowcard{2.75+\ox}{0.62+\oy}{1.55}{1.05}
}
\draw[black!55,line width=0.8pt] plot[smooth] coordinates
  {(2.82,0.95)(3.07,1.3)(3.32,1.0)(3.57,1.25)(3.85,1.2)};
\draw[cRet,line width=1.1pt] plot[smooth] coordinates
  {(3.85,1.2)(3.98,1.47)(4.11,1.05)(4.23,1.25)};
\draw[black!30,densely dotted] (3.85,0.67) -- (3.85,1.62);
\node[lab] at (3.55,0.15) {Neighbors $\mathcal{N}_i$};
\draw[flow] (db.east) -- (2.75,1.15);

\draw[fill=cBase!10,draw=cBase,line width=0.9pt]
  (2.8,3.25) -- (2.8,4.75) -- (4.2,4.4) -- (4.2,3.6) -- cycle;
\node[align=center] at (3.45,4.0) {TSFM\\ $f_{\theta}$};
\node at (2.83,4.72) {\frz};
\draw[flow] (2.1,4.0) -- (2.8,4.0);

\node[draw=cBase,fill=cBase!10,line width=0.9pt,rounded corners=2pt,
      minimum height=1.0cm,align=center,inner sep=4pt] (fus) at (5.7,4.0)
  {Forecasting\\ module $\eta$};
\node at (fus.north east) {\fire{cStwo}};
\draw[flow] (4.2,4.0) -- (fus.west);

\shadowcard{7.2}{3.35}{2.4}{1.3}
\fill[cBase!15] plot[smooth] coordinates {(7.3,4.25)(7.85,4.45)(8.4,4.1)(8.95,4.35)(9.5,4.2)}
   -- plot[smooth] coordinates {(9.5,3.65)(8.95,3.8)(8.4,3.55)(7.85,3.85)(7.3,3.7)} -- cycle;
\draw[cBase,line width=1.1pt] plot[smooth] coordinates
  {(7.3,3.98)(7.85,4.12)(8.4,3.8)(8.95,4.05)(9.5,3.9)};
\node[lab] at (8.4,3.05) {Base forecast $\widehat{\mathbf{y}}^{0}_i$};
\draw[flow] (fus.east) -- (7.2,4.0);

\draw[flow] (3.45,1.87) -- (3.45,2.68) -| (fus.south);

\node[draw=cRet,fill=cRet!10,line width=0.9pt,rounded corners=2pt,
      minimum width=1.3cm,minimum height=1.2cm,align=center,font=\footnotesize] (aln) at (5.7,1.15)
  {Level /\\ Anchor\\ align};
\draw[flow] (4.52,1.15) -- (aln.west);

\draw[thin,gap] (1.95,3.35) -- (1.95,1.97) -- (5.7,1.97) -- (aln.north);
\node[font=\footnotesize\bfseries\itshape,text=black!75,align=center,anchor=south,inner sep=1pt] at (4.55,1.99) {query\\[-3pt] statistics};

\shadowcard{7.2}{0.5}{2.4}{1.3}
\draw[black!30,densely dashed] (7.25,1.05) -- (9.55,1.05);
\draw[cBase,line width=1pt] plot[smooth] coordinates
  {(7.28,0.9)(7.55,1.25)(7.8,0.85)(8.05,1.15)(8.25,1.4)};
\foreach \o in {-0.22,0.0,0.2}
  \draw[cRet!55,line width=0.8pt] plot[smooth] coordinates
    {(8.25,1.05+\o)(8.6,1.2+\o)(8.95,0.95+\o)(9.52,1.1+\o)};
\foreach \o in {-0.25,0.0,0.22}
  \draw[cRet,line width=0.9pt] plot[smooth] coordinates
    {(8.25,1.4)(8.6,1.46+0.5*\o)(8.95,1.3+\o)(9.52,1.4+1.2*\o)};
\draw[black!30,densely dotted] (8.25,0.55) -- (8.25,1.75);
\node[lab] at (8.4,0.15) {Candidates $\mathcal{Z}_i$};
\draw[flow] (aln.east) -- (7.2,1.15);

\node[draw=cBase!65,fill=cBase!5,line width=0.65pt,
      rounded corners=2pt,minimum height=0.68cm,
      inner xsep=5pt,inner ysep=2pt,align=center,font=\small] (refFeatures) at (12.75,4.62)
  {Reference-based features};

\fill[cBase!70] (4.45,4.0) circle (1.1pt);
\draw[->,line width=0.65pt,draw=cBase!70,densely dashed]
  (4.45,4.0) -- (4.45,5.15) -- (13.3,5.15) -- (13.3,4.96);
\node[font=\footnotesize\bfseries\itshape,text=cBase!80!black,anchor=south,inner sep=1pt] at (6.6,5.19)
  {\frz\ reference branch $\eta^{R}$};

\node[draw=cGate,fill=cGate!6,line width=0.9pt,rounded corners=3pt,
      minimum width=1.2cm,minimum height=1.05cm] (gt) at (11.7,2.75) {};
\mlp{11.7}{2.75}{cGate}
\node[lab,anchor=south] at (11.45,3.3) {Gate $b_{\psi_u}$};
\node at ([xshift=-0.02cm]gt.north east) {\fire{cSone}};
\node at ([xshift=0.2cm]gt.north east) {\fire{cStwo}};

\node[draw=cRet,fill=cRet!6,line width=0.9pt,rounded corners=3pt,
      minimum width=1.2cm,minimum height=1.05cm] (sc) at (11.7,1.15) {};
\mlp{11.7}{1.15}{cRet}
\node at (sc.north east) {\fire{cSone}};
\node[lab] at (11.9,0.25) {Scorer $s_{\psi_c}$};

\draw[draw=black!70,line width=0.75pt]
  (refFeatures.west) -- (10.55,4.62) -- (10.55,2.75);
\fill[black!70] (10.55,2.75) circle (1pt);
\draw[flow] (10.55,2.75) -- (gt.west);
\draw[flow] (10.55,2.75) -- (10.55,1.15) -- (sc.west);
\node[font=\footnotesize,anchor=south west,inner sep=1pt] at (10.57,2.77) {$\boldsymbol\phi_i^R$};
\node[font=\footnotesize,anchor=south west,inner sep=1pt] at (10.57,1.17) {$\mathbf e_{ij}^R$};

\draw[thin,gap] (9.6,1.55) -- (10.15,1.55) -- (10.15,4.5) -- ($(refFeatures.west)+(0,-0.12)$);

\node[op,draw=cGate,text=cGate] (u) at (13.1,2.75) {$\sigma$};
\draw[flow] (gt.east) -- (u.west);
\node[cnote,text=cGate,anchor=north] at (13.1,2.43) {correction\\[-2pt] strength $u_i$};

\begin{scope}[shift={(13.3,0.7)}]
  \draw[black!40] (0,0) -- (1.15,0);
  \foreach \x/\h in {0.1/0.25,0.3/0.7,0.5/0.4,0.7/0.12,0.9/0.55}
    \fill[cRet!80] (\x,0) rectangle ++(0.14,\h);
\end{scope}
\draw[flow] (sc.east) -- (13.3,1.15);
\node[font=\scriptsize,anchor=south,inner sep=1pt] at (12.8,1.17) {softmax};
\node[lab] at (13.87,0.42) {$p_{ij}$};

\node[op] (sum) at (14.95,1.15) {$\Sigma$};
\draw[flow] (14.5,1.15) -- (sum.west);
\draw[cRet,line width=0.75pt] (9.6,0.75) -- (9.85,0.75) -- (9.85,-0.05) -- (14.95,-0.05);
\draw[flow,cRet] (14.95,-0.05) -- (sum.south);
\node[font=\footnotesize,text=cRet,anchor=south] at (15.38,-0.05) {$\mathbf{Z}_{ij}$};

\node[op] (mt) at (15.85,4.0) {$\times$};
\node[op] (mb) at (15.85,1.15) {$\times$};
\node[op] (add) at (16.62,2.6) {$+$};

\draw[flow,cBase,gap]
  (9.6,4.0) -- (mt.west);
\draw[flow,cRet] (sum.east) -- (mb.west) node[midway,above,font=\footnotesize] {$\mathbf{r}_i$};
\node[cnote,text=cRet,anchor=north] at (15.5,0.9) {proposal};
\draw[flow,cGate] (u.north) |- (15.85,3.45) -- (mt.south);
\node[font=\footnotesize,text=cGate,anchor=south] at (15.2,3.45) {$1-u_i$};
\draw[flow,cGate] (u.east) -- (14.3,2.75) |- (15.5,1.95) -| (mb.north);
\node[font=\footnotesize,text=cGate,anchor=south] at (15.2,1.95) {$u_i$};
\draw[flow] (mt.east) -| (add.north);
\draw[flow] (mb.east) -| (add.south);

\shadowcard{17.45}{1.95}{2.1}{1.3}
\fill[cOut!15] plot[smooth] coordinates {(17.55,2.9)(18.0,3.1)(18.5,2.8)(19.0,3.05)(19.45,2.95)}
   -- plot[smooth] coordinates {(19.45,2.4)(19.0,2.5)(18.5,2.2)(18.0,2.5)(17.55,2.35)} -- cycle;
\draw[cBase,densely dashed,line width=0.8pt] plot[smooth] coordinates
  {(17.55,2.45)(18.0,2.6)(18.5,2.3)(19.0,2.55)(19.45,2.4)};
\draw[cOut,line width=1.2pt] plot[smooth] coordinates
  {(17.55,2.65)(18.0,2.83)(18.5,2.53)(19.0,2.78)(19.45,2.67)};
\draw[flow] (add.east) -- (17.45,2.6);
\node[lab,anchor=north] at (18.5,1.7)
  {\textbf{Refined forecast}\\[1pt]
   $\begin{aligned}\widehat{\mathbf{y}}_i&=(1-u_i)\,\widehat{\mathbf{y}}^{0}_i\\[-2pt]&\quad+u_i\,\mathbf{r}_i\end{aligned}$};

\draw[black!20,line width=0.45pt]
  (0,-0.42) -- (19.55,-0.42);

\fill[cSone!6,draw=cSone!35,line width=0.6pt,rounded corners=5pt]
  (-0.15,-0.55) rectangle (9.38,-2.4);
\fill[cStwo!6,draw=cStwo!35,line width=0.6pt,rounded corners=5pt]
  (10.0,-0.55) rectangle (19.6,-2.4);

\begin{scope}[
  shift={(0,-0.87)},
  trainlabel/.style={
    anchor=west,font=\footnotesize,text=black!55
  },
  traintag/.style={
    rounded corners=2pt,draw=cSone!25,fill=cSone!8,
    minimum height=0.43cm,inner xsep=5pt,inner ysep=2pt,
    font=\footnotesize,text=cSone!65!black,align=center
  }
]

  \node[rounded corners=2pt,fill=cSone,inner xsep=6pt,
    inner ysep=3pt,font=\footnotesize\bfseries,text=white]
    at (0.78,0) {Stage I};
  \node[anchor=west,font=\small\bfseries,text=cSone!65!black]
    at (1.65,0) {Opportunity-aware learning};

  \node[trainlabel] at (0,-0.64) {Update};
  \node[traintag,minimum width=2.95cm]
    at (2.93,-0.64) {Candidate scorer $\psi_c$};
  \node[traintag,minimum width=1.55cm]
    at (5.3,-0.64) {Gate $\psi_u$};

  \node[anchor=east,font=\scriptsize\bfseries,text=cSone!75!black]
    at (9.2,0) {Across-query weighting};
  \node[anchor=east,draw=cSone!60,fill=white,rounded corners=2pt,
    inner xsep=4pt,inner ysep=2.5pt,font=\footnotesize,line width=0.6pt]
    at (9.2,-0.64) {$w_i=s_i^{2}\,(1+\beta\,h_i\,g_i)$};

  \node[trainlabel] at (0,-1.20) {Objective};
  \node[anchor=west,font=\footnotesize,text=black!85]
    at (1.43,-1.20)
    {Opportunity-weighted prediction $+$ gate supervision};
\end{scope}

\draw[-{Stealth[length=2mm]},draw=black!45,line width=0.85pt]
  (9.44,-0.87) -- (9.96,-0.87);

\begin{scope}[
  shift={(10.15,-0.87)},
  trainlabel/.style={
    anchor=west,font=\footnotesize,text=black!55
  },
  traintag/.style={
    rounded corners=2pt,draw=cStwo!25,fill=cStwo!8,
    minimum height=0.43cm,inner xsep=5pt,inner ysep=2pt,
    font=\footnotesize,text=cStwo!65!black,align=center
  }
]

  \node[rounded corners=2pt,fill=cStwo,inner xsep=6pt,
    inner ysep=3pt,font=\footnotesize\bfseries,text=white]
    at (0.78,0) {Stage II};
  \node[anchor=west,font=\small\bfseries,text=cStwo!65!black]
    at (1.65,0) {Reference-guided joint calibration};

  \node[trainlabel] at (0,-0.64) {Update};
  \node[traintag,minimum width=2.95cm]
    at (2.93,-0.64) {Forecasting module $\eta$};
  \node[traintag,minimum width=1.55cm]
    at (5.3,-0.64) {Gate $\psi_u$};

  \node[trainlabel] at (0,-1.20) {Objective};
  \node[anchor=west,font=\footnotesize,text=black!85]
    at (1.43,-1.20)
    {Prediction MSE $+$ reference anchoring};
\end{scope}

\end{tikzpicture}}%
\caption{Overview of ROAR.
(a--c) Aligned historical continuations refine the base forecast
through learned candidate aggregation and input-dependent gating.
The TSFM backbone remains frozen, and the scorer and gate
use frozen reference features.
Stage~I applies opportunity-aware weights $w_i$ across queries,
emphasizing difficult queries with useful candidates.
Stage~II jointly calibrates the forecasting module and gate,
anchoring their combined prediction to the Stage-I refined output.
}
\label{fig:framework}
\end{figure}

%% file: experiment.tex
\section{Experiments}

\subsection{Experimental Settings}

\textbf{Pretraining and Evaluation Datasets  }
For pretraining, we adopt the Chronos pretraining dataset \citep{ansari2024chronos} where 50M data points are uniformly sampled following \citep{ningts,lee2026cross} and we adopt the same in-domain retrieval knowledge base following \citep{ningts} which is constructed by a subset of 5M data points.
We partition both the pretraining dataset and the retrieval knowledge base into segments using a fixed context window, yielding approximately 26 million pretraining pairs and 2.8 million knowledge base entries.

We adopt 7 benchmark time series forecasting datasets for zero-shot forecasting evaluation, including ETTh1, ETTh2, ETTm1, ETTm2 \citep{zhou2021informer}, Weather, Electricity \citep{wutimesnet}, and Exchange Rate \citep{lai2018modeling}. We conduct zero-shot evaluation on the test set of each dataset. Details of each dataset are presented in Table \ref{tab:dataset} in Appendix \ref{app_exp}.

\textbf{Baselines and Implementation.}
We compare ROAR with 7 TSFMs: Chronos~\citep{ansari2024chronos}, Chronos-Bolt~\citep{ansari2024chronos}, MOMENT~\citep{goswami2024moment}, TTM~\citep{ekambaram2024tiny}, Moirai~\citep{woo2024unified}, TimesFM~\citep{das2024decoder}, and Time-MoE~\citep{shitime}. We also compare with 5 RAG-based TSF methods: RAF~\citep{tireretrieval}, TS-RAG~\citep{ningts}, Cross-RAG~\citep{lee2026cross}, PFRP~\citep{du2026predicting}, and RATL~\citep{he2026ratl}. PFRP and RATL are adapted to the zero-shot setting. All RAG-based methods, including ROAR, use the frozen Chronos-Bolt backbone and share the same training and inference protocols. Following~\citep{ningts}, the input length and forecasting horizon are set as 512 and 64, and the pre-training corpus and retrieval settings follow TS-RAG with $K=10$ retrieved examples. After training the forecasting module, ROAR is trained for 2,500 steps in Stage~I and 2,000 steps in Stage~II. We set $\beta=0.5$, $\lambda_{\mathrm{gate}}=1$, and $\lambda_{\mathrm{ref}}=15$. All methods are evaluated with MSE and MAE. We report the averaged scores of 5 independent runs for our ROAR and all reproduced methods. Further implementation details and hyper-parameters are provided in Appendix~\ref{app_exp_set}.

\begin{table*}[!t]
\input{Tables/result_TSFMs}
\vspace{8pt}
\input{Tables/result_TS_RAG}
\end{table*}

\subsection{Zero-shot Forecasting Performance}
We compare \baby with 7 TSFMs and 5 RAG-based TSF methods. Baseline results come from reproductions using the official source code and from the results reported in~\citep{lee2026cross}.

\textbf{Comparison with TSFMs.}
As shown in Table~\ref{tab:tsfms}, \baby achieves the best performance in most settings among all TSFMs.
Compared with its Chronos-Bolt backbone, the second-best model on average, \baby reduces the average MSE from 0.201 to 0.187 (7.0\%). The largest reduction is on ETTm1, from 0.311 to 0.274 (11.9\%).
On Electricity and Exchange, \baby is slightly behind the best MAE (0.205 vs.\ 0.200 and 0.176 vs.\ 0.172) while keeping the best MSE. This is consistent with Stage~II being calibrated with point-forecast MSE.
These results show that retrieved historical continuations add useful information that pretrained backbones alone do not provide.

\input{Tables/result_backbone}

\textbf{Comparison with RAG-based TSF methods.}
We present the comparison with RAF, TS-RAG, Cross-RAG, PFRP, and RATL in Table~\ref{tab:rag_tsf}. PFRP and RATL were originally trained per dataset. We adapt them to the zero-shot setting by training once on the same pre-training corpus over a frozen Chronos-Bolt backbone. This protocol is same for all methods.
We can find that, \baby achieves the best MSE on six of the seven datasets and the best average MSE and MAE.
Compared with Cross-RAG, the second-best method in average MSE, \baby reduces the average MSE from 0.193 to 0.187 and the average MAE from 0.255 to 0.250. The largest reduction is on ETTm1 (0.290 to 0.274).
Existing methods learn how to use candidates within each query, but weight all training queries
equally regardless of whether their candidates can improve the base forecast.
\baby instead emphasizes training queries where retrieval offers an opportunity and learns how
strongly to correct toward it, making its use of retrieved information more effective.

\subsection{Effectiveness Examination of ROAR}

\input{Tables/result_ablation}

\paragraph{Ablation Study}
We examine each training stage of \baby by removing them one at a time (Table~\ref{tab:ablation_stages}).
``w/o Stage~II'' uses the Stage-I prediction $\widehat{\mathbf{y}}_i^R$ without joint calibration, and ``w/o Stage~I \& II'' reduces to the base forecaster $\widehat{\mathbf{y}}_i^{0,R}$; Stage~I is not removed alone since Stage~II builds on it.
All variants share the same frozen backbone, retrieval, and zero-shot protocol.
Removing either stage degrades performance.
Stage~I contributes most, and removing it raises the average MSE from 0.1884 to 0.1950, showing that opportunity-aware refinement turns retrieved candidates into useful corrections.
Stage~II further lowers MSE on every dataset, with smaller and mixed effects on MAE since it is trained with MSE.
Overall, Stage~I learns where and how much to trust retrieved candidates, and Stage~II calibrates the base forecast around this refinement.

\paragraph{Plugging ROAR into RAG-based TSF Methods}
We test whether ROAR works as a plug-in refiner for existing retrieval-augmented forecasters.
We apply it to three methods: RAF, TS-RAG, and Cross-RAG. Each method keeps its own backbone, retrieval pipeline, and hyper-parameters, and its forecast serves as the base forecast $\widehat{\mathbf{y}}_i^{0}$ that ROAR refines. Only the ROAR modules are added.
Fig.~\ref{fig:plugin} reports the relative MSE reduction.
ROAR lowers the average MSE by 4.7\%, 4.5\%, and 1.6\%, respectively. It improves 18 of the 21 Base--dataset pairs and degrades only RAF on Exchange.
Cross-RAG gains the least. A likely reason is that its cross-attention fusion already uses the retrieved series more effectively, so less improvement is left for refinement.
Overall, ROAR is compatible with different retrieval-augmented designs and consistently improves.

\paragraph{Generalization across Backbones}
To examine whether ROAR generalizes beyond different backbones,
we replace the backbone with Moirai2.0~\citep{liu2025moirai2} and TimesFM2.5~\citep{das2024decoder}, and keep the same retrieval procedure and zero-shot protocol.
Table~\ref{tab:zeroshot_backbones} shows that ROAR improves both backbones on all seven datasets and on both metrics. The average MSE falls by 8.5\% for Moirai2.0 and 5.0\% for TimesFM2.5.
The largest gain is on Electricity (20.5\% with Moirai2.0), whose recurring daily and weekly patterns make retrieved continuations especially informative. Gains are smaller on Weather (1.8\%), where historical segments offer less transferable information.
Overall, ROAR is backbone-agnostic and complements the representations learned by different TSFMs.

\begin{figure}[t]
\centering
{\phantomsubcaption\label{fig:opportunity_gain}}
{\phantomsubcaption\label{fig:weather-ablation}}
{\phantomsubcaption\label{fig:weather-combined}}
\resizebox{\linewidth}{!}{\input{plot/experiments/visualization_main2}}
\caption{(a) MSE improvement over Chronos-Bolt over the best-candidate opportunity $g_i$.
(b) Forecasting visualization between ROAR and the ablative version ``ROAR w/o Stage-I \& II''.
(c) Forecasting visualization between ROAR and RAG-based TSF methods TS-RAG and Cross-RAG.}
\label{fig:opp_weather}
\end{figure}

\subsection{Visualizations}

\paragraph{Retrieval Opportunity Gains Visualization}
We examine whether ROAR's gains concentrate on queries with beneficial candidates.
We compute $g_i$ (Eq.~\ref{eq:opportunity_score}) with Chronos-Bolt as a common base over the top-10
level-aligned candidates shared by TS-RAG and ROAR, and bin queries into $g_i=0$ and within-dataset
quartiles $B_1$--$B_4$ of $g_i>0$.
To avoid selecting bins on the evaluated errors, $g_i$ and the improvement are computed on opposite halves
of the horizon, averaged over both directions; $g_i>0$ thus indicates a better candidate on the binning half.
As shown in Fig.~\ref{fig:opp_weather}(a), gains of all methods grow with $g_i$, and ROAR grows fastest:
it reaches $+17.0\%$ in $B_4$ versus $+5.9\%$ (TS-RAG) and $+9.0\%$ (Cross-RAG), and leads in every $g_i>0$ bin.
When no candidate is better, all methods lose slightly ($-1.2\%$ to $-2.9\%$).
Thus, ROAR converts retrieval opportunity into forecast improvement more effectively than existing
retrieval-augmented forecasters.

\paragraph{Forecasting Visualizations}
We present the forecasting visualizations between ROAR and the ablative version and RAG-based TSF methods in Fig.~\ref{fig:opp_weather}(b,c).
We can find that our ROAR consistently outperforms both the ablative version and the baseline methods in terms of trend fidelity and continuity with the history, directly indicating the effectiveness of our proposal. More visualizations are provided in Appendix~\ref{app_vis}.

%% file: Tables/result_TSFMs.tex
\centering
\caption{\textbf{Zero-shot forecasting results across TSFMs.} Best results are in \textbf{boldface}, and the second-best results are \underline{underlined}.}
\label{tab:tsfms}
\renewcommand{\arraystretch}{1.25}
\setlength{\tabcolsep}{3.5pt}
\resizebox{\linewidth}{!}{
\begin{tabular}{l cc cc cc cc cc cc cc cc}
\toprule
\multirow{2}{*}{\textbf{Methods}}
& \multicolumn{2}{c}{\textbf{\baby}}
& \multicolumn{2}{c}{\textbf{Chronos-bolt}}
& \multicolumn{2}{c}{\textbf{MOMENT}}
& \multicolumn{2}{c}{\textbf{TTM}}
& \multicolumn{2}{c}{\textbf{Moirai}}
& \multicolumn{2}{c}{\textbf{TimesFM}}
& \multicolumn{2}{c}{\textbf{Chronos}}
& \multicolumn{2}{c}{\textbf{Time-MoE}} \\
& \multicolumn{2}{c}{(Ours)}
& \multicolumn{2}{c}{(TMLR 2024)}
& \multicolumn{2}{c}{(ICML 2024)}
& \multicolumn{2}{c}{(NeurIPS 2024)}
& \multicolumn{2}{c}{(ICML 2024)}
& \multicolumn{2}{c}{(ICML 2024)}
& \multicolumn{2}{c}{(TMLR 2024)}
& \multicolumn{2}{c}{(ICLR 2025)} \\
\cmidrule(lr){2-3}\cmidrule(lr){4-5}\cmidrule(lr){6-7}\cmidrule(lr){8-9}
\cmidrule(lr){10-11}\cmidrule(lr){12-13}\cmidrule(lr){14-15}\cmidrule(lr){16-17}
\textbf{Metric}
& MSE & MAE & MSE & MAE & MSE & MAE & MSE & MAE
& MSE & MAE & MSE & MAE & MSE & MAE & MSE & MAE \\
\midrule
ETTh1
& \textbf{0.339} & \textbf{0.363}
& \underline{0.362} & \underline{0.365}
& 0.392 & 0.411
& \underline{0.362} & 0.371
& 0.369 & 0.384
& 0.425 & 0.383
& 0.422 & 0.381
& \underline{0.362} & 0.367 \\
ETTh2
& \textbf{0.235} & \textbf{0.295}
& \underline{0.252} & \underline{0.299}
& 0.274 & 0.333
& 0.253 & 0.303
& 0.255 & 0.305
& 0.289 & 0.323
& 0.266 & 0.314
& \underline{0.252} & 0.322 \\
ETTm1
& \textbf{0.274} & \textbf{0.309}
& \underline{0.311} & \underline{0.319}
& 0.351 & 0.383
& 0.315 & 0.325
& 0.540 & 0.432
& 0.332 & 0.333
& 0.394 & 0.370
& 0.321 & 0.334 \\
ETTm2
& \textbf{0.141} & \textbf{0.222}
& \underline{0.149} & \underline{0.224}
& 0.170 & 0.258
& 0.151 & 0.241
& 0.196 & 0.269
& 0.170 & 0.255
& 0.166 & 0.252
& 0.157 & 0.254 \\
Weather
& \textbf{0.141} & \textbf{0.178}
& 0.153 & \underline{0.183}
& 0.180 & 0.238
& 0.154 & 0.189
& 0.171 & 0.191
& --- & ---
& 0.190 & 0.211
& \underline{0.149} & 0.184 \\
Electricity
& \textbf{0.112} & 0.205
& \underline{0.113} & \textbf{0.200}
& 0.197 & 0.303
& 0.172 & 0.264
& 0.183 & 0.281
& --- & ---
& 0.146 & 0.224
& 0.114 & \underline{0.203} \\
Exchange
& \textbf{0.066} & 0.176
& \underline{0.067} & 0.178
& 0.098 & 0.206
& \textbf{0.066} & \underline{0.173}
& \textbf{0.066} & \textbf{0.172}
& 0.070 & 0.180
& 0.083 & 0.188
& 0.085 & 0.206 \\
\midrule
Average
& \textbf{0.187} & \textbf{0.250}
& \underline{0.201} & \underline{0.253}
& 0.237 & 0.304
& 0.210 & 0.267
& 0.254 & 0.291
& --- & ---
& 0.238 & 0.277
& 0.206 & 0.267 \\
\bottomrule
\end{tabular}
}

%% file: Tables/result_TS_RAG.tex
\centering
\caption{\textbf{Zero-shot forecasting results across RAG-based TSF methods.} Best results are in \textbf{boldface}, and the second-best results are \underline{underlined}.}
\label{tab:rag_tsf}
\renewcommand{\arraystretch}{1.25}
\setlength{\tabcolsep}{3.5pt}
\resizebox{\linewidth}{!}{
\begin{tabular}{l cc cc cc cc cc cc cc | cc}
\toprule
\multirow{2}{*}{\textbf{Methods}}
& \multicolumn{2}{c}{\textbf{ETTh1}}
& \multicolumn{2}{c}{\textbf{ETTh2}}
& \multicolumn{2}{c}{\textbf{ETTm1}}
& \multicolumn{2}{c}{\textbf{ETTm2}}
& \multicolumn{2}{c}{\textbf{Weather}}
& \multicolumn{2}{c}{\textbf{Electricity}}
& \multicolumn{2}{c}{\textbf{Exchange}}
& \multicolumn{2}{|c}{\textbf{Average}} \\
\cmidrule(lr){2-3}\cmidrule(lr){4-5}\cmidrule(lr){6-7}\cmidrule(lr){8-9}
\cmidrule(lr){10-11}\cmidrule(lr){12-13}\cmidrule(lr){14-15}\cmidrule(lr){16-17}
& MSE & MAE & MSE & MAE & MSE & MAE & MSE & MAE
& MSE & MAE & MSE & MAE & MSE & MAE & MSE & MAE \\
\midrule
PFRP$_{\text{Chronos-bolt}}$
& 0.361 & 0.365 & 0.249 & 0.299 & 0.311 & 0.319 & 0.149 & 0.224
& 0.153 & 0.184 & 0.114 & 0.201 & 0.068 & 0.179
& 0.201 & 0.253 \\
RATL$_{\text{Chronos-bolt}}$
& 0.362 & 0.365 & 0.251 & 0.299 & 0.310 & 0.318 & 0.148 & 0.224
& 0.152 & 0.182 & 0.113 & \underline{0.200} & 0.067 & 0.178
& 0.201 & 0.252 \\
RAF$_{\text{Chronos-bolt}}$
& 0.366 & 0.371 & 0.252 & 0.305 & 0.306 & 0.329 & 0.148 & 0.228
& 0.178 & 0.206 & 0.118 & 0.211 & \textbf{0.063} & \textbf{0.172}
& 0.204 & 0.260 \\
TS-RAG
& 0.359 & \underline{0.364} & 0.248 & \underline{0.299} & 0.300 & \underline{0.313} & 0.147 & \underline{0.222}
& 0.150 & 0.181 & \underline{0.113} & \textbf{0.200} & 0.067 & 0.176
& 0.198 & \underline{0.251} \\
Cross-RAG
& \underline{0.344} & 0.371 & \underline{0.244} & 0.302 & \underline{0.290} & 0.319 & \underline{0.146} & 0.226
& \underline{0.145} & \underline{0.180} & 0.113 & 0.205 & 0.070 & 0.183
& \underline{0.193} & 0.255 \\
\baby (Ours)
& \textbf{0.339} & \textbf{0.363} & \textbf{0.235} & \textbf{0.295} & \textbf{0.274} & \textbf{0.309} & \textbf{0.141} & \textbf{0.222}
& \textbf{0.141} & \textbf{0.178} & \textbf{0.112} & 0.205 & \underline{0.066} & \underline{0.176}
& \textbf{0.187} & \textbf{0.250} \\
\bottomrule
\end{tabular}
}

%% file: Tables/result_backbone.tex
\definecolor{cblue}{HTML}{74A9C5}
\definecolor{cyellow}{HTML}{EDDDAB}
\definecolor{cred}{HTML}{DD7389}
\newcommand{\dd}[1]{$-#1$}
\newsavebox{\roartabbox}
\newlength{\roartabH}

\begin{figure}[t]
\centering
\captionsetup{position=top, skip=3pt}
\sbox{\roartabbox}{%
\setlength{\tabcolsep}{2.5pt}%
\renewcommand{\arraystretch}{1.1}%
\resizebox{0.49\textwidth}{!}{%
\begin{tabular}{l ccc ccc}
\toprule
\multirow{2}{*}{\textbf{Data}}
& \multicolumn{3}{c}{\textbf{Moirai 2.0}}
& \multicolumn{3}{c}{\textbf{TimesFM 2.5}} \\
\cmidrule(lr){2-4}\cmidrule(lr){5-7}
& Base & \textbf{+ROAR} & $\Delta\%$
& Base & \textbf{+ROAR} & $\Delta\%$ \\
\midrule
ETTh1   & .446/.451 & \textbf{.412}/\textbf{.428} & \cellcolor{black!22}\dd{7.5}/\dd{5.2}   & .404/.421 & \textbf{.390}/\textbf{.411} & \cellcolor{black!10}\dd{3.4}/\dd{2.3} \\
ETTh2   & .274/.342 & \textbf{.258}/\textbf{.331} & \cellcolor{black!18}\dd{6.1}/\dd{3.1}   & .268/.336 & \textbf{.252}/\textbf{.326} & \cellcolor{black!18}\dd{5.9}/\dd{3.0} \\
ETTm1   & .379/.398 & \textbf{.352}/\textbf{.380} & \cellcolor{black!22}\dd{7.3}/\dd{4.5}   & .348/.376 & \textbf{.332}/\textbf{.367} & \cellcolor{black!13}\dd{4.5}/\dd{2.5} \\
ETTm2   & .189/.275 & \textbf{.177}/\textbf{.266} & \cellcolor{black!20}\dd{6.5}/\dd{3.2}   & .187/.272 & \textbf{.175}/\textbf{.264} & \cellcolor{black!19}\dd{6.5}/\dd{3.0} \\
Weather & .157/.205 & \textbf{.154}/\textbf{.202} & \cellcolor{black!5}\dd{1.8}/\dd{1.4}    & .156/.202 & \textbf{.153}/\textbf{.201} & \cellcolor{black!5}\dd{1.8}/\dd{0.6} \\
Elec.   & .244/.351 & \textbf{.194}/\textbf{.304} & \cellcolor{black!60}\dd{20.5}/\dd{13.6} & .184/.293 & \textbf{.168}/\textbf{.275} & \cellcolor{black!27}\dd{9.0}/\dd{6.3} \\
Exch.   & .092/.215 & \textbf{.085}/\textbf{.207} & \cellcolor{black!26}\dd{8.6}/\dd{3.9}   & .094/.216 & \textbf{.088}/\textbf{.209} & \cellcolor{black!17}\dd{5.8}/\dd{3.2} \\
\midrule
\textbf{Avg.} & .255/.320 & \textbf{.233}/\textbf{.303} & \cellcolor{black!25}\dd{8.5}/\dd{5.4} & .234/.303 & \textbf{.223}/\textbf{.293} & \cellcolor{black!15}\dd{5.0}/\dd{3.0} \\
\bottomrule
\end{tabular}}}%
\setlength{\roartabH}{\ht\roartabbox}%
\addtolength{\roartabH}{\dp\roartabbox}%
\begin{minipage}[t]{0.46\textwidth}
\vspace{0pt}
\captionof{figure}{\textbf{ROAR as a plug-in across retrieval-augmented methods.} Relative MSE reduction of each method equipped with ROAR (higher is better).}
\label{fig:plugin}
\centering
\resizebox{\linewidth}{!}{%
\begin{tikzpicture}
\begin{axis}[
    ybar, bar width=3.4pt,
    scale only axis,
    width=\linewidth-0.9cm, height=\roartabH-0.55cm,
    ymin=-6, ymax=15,
    symbolic x coords={ETTh1,ETTh2,ETTm1,ETTm2,Weather,Elec.,Exch.,Avg.},
    xtick=data,
    x tick label style={font=\scriptsize, rotate=30, anchor=north east, inner sep=1pt},
    y tick label style={font=\scriptsize},
    ytick={-6,-3,0,3,6,9,12},
    ylabel={MSE reduction (\%)},
    ylabel style={font=\footnotesize, yshift=-4pt},
    enlarge x limits=0.07,
    axis x line*=bottom, axis y line*=left,
    tick align=outside, tick style={black},
    extra y ticks={0}, extra y tick labels={},
    extra y tick style={grid=major, grid style={gray, dashed, thin}},
    legend style={font=\scriptsize, at={(0.5,1.0)}, anchor=north, legend columns=3,
      draw=none, fill=none, legend cell align=left, column sep=4pt, inner sep=1pt},
    legend image code/.code={\fill[#1] (0cm,-0.06cm) rectangle (0.22cm,0.1cm);},
    nodes near coords={\pgfmathprintnumber[fixed,fixed zerofill,precision=1]{\pgfplotspointmeta}},
    point meta=y,
    visualization depends on={y \as \rawy},
    every node near coord/.append style={font=\tiny, rotate=90, inner sep=1pt,
      anchor={\ifdim \rawy pt<0pt east\else west\fi}},
  ]
\addplot[fill=cblue, draw=none] coordinates {(ETTh1,9.5) (ETTh2,6.5) (ETTm1,0.0) (ETTm2,2.1) (Weather,7.3) (Elec.,4.9) (Exch.,-2.9) (Avg.,4.7)};
\addplot[fill=cyellow, draw=none] coordinates {(ETTh1,4.2) (ETTh2,2.4) (ETTm1,7.7) (ETTm2,3.4) (Weather,4.7) (Elec.,0.9) (Exch.,6.0) (Avg.,4.5)};
\addplot[fill=cred, draw=none] coordinates {(ETTh1,1.2) (ETTh2,0.4) (ETTm1,1.4) (ETTm2,1.4) (Weather,2.8) (Elec.,0.0) (Exch.,7.1) (Avg.,1.6)};
\legend{RAF, TS-RAG, Cross-RAG}
\end{axis}
\end{tikzpicture}}
\end{minipage}\hfill
\begin{minipage}[t]{0.49\textwidth}
\vspace{0pt}
\captionof{table}{\textbf{ROAR across TSFM backbones.} $\Delta\%$: relative change of MSE\,/\,MAE over the backbone; deeper shading means larger MSE reduction.}
\label{tab:zeroshot_backbones}
\centering
\vspace{6pt}
\usebox{\roartabbox}
\end{minipage}
\end{figure}

%% file: Tables/result_ablation.tex
\newcommand{\ablneg}[1]{\raisebox{-0.5ex}{\scriptsize$-$#1}}
\newcommand{\ablpos}[1]{\raisebox{-0.5ex}{\scriptsize$+$#1}}
\newcommand{\ablzero}{\raisebox{-0.5ex}{\scriptsize$\pm$0.0000}}

\begin{wraptable}{r}{0.58\textwidth}
\vspace{-12pt}
\centering
\caption{\textbf{Ablation of ROAR's two stages} (Chronos-Bolt, zero-shot). Subscripts: absolute change caused by removing one more stage, i.e., w/o Stage~II vs.\ ROAR, and w/o Stage~I \& II vs.\ w/o Stage~II.}
\label{tab:ablation_stages}
\vspace{2pt}
\setlength{\tabcolsep}{2.5pt}
\renewcommand{\arraystretch}{1.1}
\resizebox{\linewidth}{!}{%
\begin{tabular}{l cc cc cc}
\toprule
\multirow{2}{*}{\textbf{Data}}
& \multicolumn{2}{c}{\textbf{ROAR}}
& \multicolumn{2}{c}{\textbf{w/o Stage II}}
& \multicolumn{2}{c}{\textbf{w/o Stage I \& II}} \\
\cmidrule(lr){2-3}\cmidrule(lr){4-5}\cmidrule(lr){6-7}
& MSE & MAE & MSE & MAE & MSE & MAE \\
\midrule
ETTh1   & \textbf{0.3385} & \textbf{0.3626} & 0.3432\ablpos{0.0047} & 0.3645\ablpos{0.0019} & 0.3527\ablpos{0.0095} & 0.3645\ablzero \\
ETTh2   & \textbf{0.2351} & 0.2953 & 0.2361\ablpos{0.0010} & \textbf{0.2952}\ablneg{0.0001} & 0.2413\ablpos{0.0052} & 0.2959\ablpos{0.0007} \\
ETTm1   & \textbf{0.2735} & \textbf{0.3087} & 0.2778\ablpos{0.0043} & 0.3092\ablpos{0.0005} & 0.2949\ablpos{0.0171} & 0.3132\ablpos{0.0040} \\
ETTm2   & \textbf{0.1408} & \textbf{0.2221} & 0.1418\ablpos{0.0010} & 0.2222\ablpos{0.0001} & 0.1459\ablpos{0.0041} & 0.2230\ablpos{0.0008} \\
Weather & \textbf{0.1405} & 0.1779 & 0.1418\ablpos{0.0013} & \textbf{0.1776}\ablneg{0.0003} & 0.1476\ablpos{0.0058} & 0.1809\ablpos{0.0033} \\
Elec.   & \textbf{0.1119} & 0.2047 & 0.1121\ablpos{0.0002} & 0.2034\ablneg{0.0013} & 0.1133\ablpos{0.0012} & \textbf{0.2028}\ablneg{0.0006} \\
Exch.   & \textbf{0.0660} & \textbf{0.1761} & 0.0662\ablpos{0.0002} & 0.1765\ablpos{0.0004} & 0.0692\ablpos{0.0030} & 0.1803\ablpos{0.0038} \\
\midrule
\textbf{Avg.} & \textbf{0.1866} & \textbf{0.2496} & 0.1884\ablpos{0.0018} & 0.2498\ablpos{0.0002} & 0.1950\ablpos{0.0066} & 0.2515\ablpos{0.0017} \\
\bottomrule
\end{tabular}}
\vspace{-10pt}
\end{wraptable}

%% file: plot/experiments/visualization_main2.tex
\definecolor{ogTs}{HTML}{4E79A7}
\definecolor{ogCr}{HTML}{59A14F}
\definecolor{ogRo}{HTML}{E15759}
\definecolor{ogAb}{HTML}{B07AA1}
\pgfplotsset{
  rowpanel/.style={
    scale only axis, width=3.4cm, height=2.7cm,
    title style={font=\small, yshift=-1mm},
    label style={font=\footnotesize},
    tick label style={font=\scriptsize},
    axis line style={black!60}, tick style={black!60},
    grid style={black!10},
    legend style={font=\footnotesize, draw=none, fill=none, legend columns=-1,
                  /tikz/every even column/.append style={column sep=8pt}},
    legend cell align=left,
  },
  wcmp/.style={
    xlabel={Time step}, ylabel={Value}, grid=major, no markers,
    yticklabel style={/pgf/number format/1000 sep={}},
    axis background/.style={fill=gray!5},
  },
  histstyle/.style={gray!70, line width=0.9pt},
  gtstyle/.style={black, dashed, line width=1.2pt, line join=round},
  ablstyle/.style={ogAb, line width=1.1pt},
  tsragstyle/.style={ogTs, line width=1.0pt},
  crossragstyle/.style={ogCr, line width=1.0pt},
  roarstyle/.style={ogRo, line width=1.4pt},
}
\begin{tikzpicture}
\begin{groupplot}[group style={group size=3 by 1, horizontal sep=1.75cm}, rowpanel]

\nextgroupplot[title={(a) Gain vs.\ opportunity $g_i$}, clip=false,
  xmin=-0.35, xmax=4.45, ymin=-5, ymax=20, ytick={-5,0,5,10,15,20}, xtick={0,...,4},
  xticklabels={{$g{=}0$\\\tiny(48\%)},{$B_1$\\\tiny(13\%)},{$B_2$\\\tiny(13\%)},{$B_3$\\\tiny(13\%)},{$B_4$\\\tiny(13\%)}},
  xticklabel style={align=center}, ymajorgrids,
  xlabel={Opportunity $g_i$ ($g_i{>}0$: quartiles)}, ylabel={MSE improvement (\%)}]
\draw[black!55, line width=0.6pt] (axis cs:-0.35,0) -- (axis cs:4.45,0);
\draw[black!40, line width=1.2pt, dash pattern=on 4pt off 2.5pt] (axis cs:0.5,-5) -- (axis cs:0.5,20);
\node[anchor=north, align=center, font=\tiny, text=black!65, inner sep=1pt] at (axis cs:0.075,19.6) {No\\better};
\node[anchor=north west, align=left, font=\tiny, text=black!65, inner sep=1pt] at (axis cs:0.62,19.6) {At least one\\better candidate};
\addplot[tsragstyle, mark=square*, mark size=1.6pt] coordinates {(0,-2.11) (1,-0.49) (2,0.66) (3,2.54) (4,5.94)};
\addplot[crossragstyle, mark=triangle*, mark size=1.9pt] coordinates {(0,-1.19) (1,1.18) (2,2.76) (3,5.05) (4,9.02)};
\addplot[roarstyle, mark=*, mark size=1.9pt] coordinates {(0,-2.92) (1,1.81) (2,4.54) (3,8.59) (4,17.03)};
\node[anchor=west, font=\tiny, text=ogTs, inner sep=1pt] at (axis cs:4.1,5.94) {+5.9};
\node[anchor=west, font=\tiny, text=ogCr, inner sep=1pt] at (axis cs:4.1,9.02) {+9.0};
\node[anchor=west, font=\tiny, text=ogRo, inner sep=1pt] at (axis cs:4.1,17.03) {+17.0};

\nextgroupplot[title={(b) Weather: ablation}, wcmp]
\addplot[histstyle] coordinates {(-80,1262.1) (-79,1265) (-78,1266.5) (-77,1266.9) (-76,1266.2) (-75,1266.8) (-74,1267.7) (-73,1267.1) (-72,1268) (-71,1266.6) (-70,1265) (-69,1267.1) (-68,1268.3) (-67,1269) (-66,1270.1) (-65,1269.3) (-64,1269.3) (-63,1269.1) (-62,1267.9) (-61,1268.7) (-60,1270.7) (-59,1271.4) (-58,1271.9) (-57,1270.4) (-56,1270.2) (-55,1270.7) (-54,1272.2) (-53,1273.7) (-52,1273.2) (-51,1273) (-50,1272.1) (-49,1271.2) (-48,1273.2) (-47,1274.5) (-46,1274) (-45,1274.6) (-44,1274.4) (-43,1272.1) (-42,1271.2) (-41,1271.4) (-40,1271.9) (-39,1272.4) (-38,1271.5) (-37,1270.8) (-36,1272.5) (-35,1273.4) (-34,1272.5) (-33,1272.6) (-32,1273) (-31,1271.5) (-30,1271.6) (-29,1271.8) (-28,1271.9) (-27,1270.8) (-26,1269.3) (-25,1268.8) (-24,1267.3) (-23,1268.2) (-22,1266) (-21,1264.1) (-20,1261.2) (-19,1260) (-18,1260.2) (-17,1259) (-16,1254.8) (-15,1252.8) (-14,1251.7) (-13,1251.1) (-12,1249.5) (-11,1248.2) (-10,1246.4) (-9,1244.1) (-8,1241.5) (-7,1238.4) (-6,1236.5) (-5,1233.3) (-4,1229.7) (-3,1226.8) (-2,1223.5) (-1,1222.4)};
\addplot[gtstyle] coordinates {(-1,1222.4) (0,1221.2) (1,1219.7) (2,1218) (3,1215.9) (4,1214.7) (5,1214) (6,1212.7) (7,1210.1) (8,1208.2) (9,1207.6) (10,1209) (11,1209.9) (12,1210.4) (13,1209) (14,1207.2) (15,1209.4) (16,1211.3) (17,1212.1) (18,1213.6) (19,1214.6) (20,1215.5) (21,1217) (22,1217.5) (23,1218.3) (24,1219.5) (25,1221.3) (26,1222.5) (27,1223.6) (28,1224.3) (29,1226.9) (30,1228.4) (31,1228.9) (32,1228.7) (33,1229.3) (34,1231.1) (35,1232.3) (36,1233.1) (37,1233.7) (38,1233.9) (39,1235.7) (40,1237.8) (41,1240.8) (42,1242.5) (43,1241.9) (44,1241.7) (45,1241) (46,1240.8) (47,1241.6) (48,1243.6) (49,1244.8) (50,1245.1) (51,1245.5) (52,1245.1) (53,1247) (54,1247.2) (55,1247.4) (56,1247.5) (57,1248.2) (58,1249.4) (59,1250.8) (60,1251.9) (61,1253.2) (62,1253.2) (63,1252.6)};
\addplot[ablstyle] coordinates {(-1,1222.4) (0,1222.2) (1,1220.1) (2,1218.7) (3,1216.7) (4,1214.9) (5,1215.2) (6,1213.5) (7,1212.8) (8,1211.7) (9,1211.4) (10,1211.4) (11,1211.5) (12,1212) (13,1212.6) (14,1212.6) (15,1212.6) (16,1212.8) (17,1212.5) (18,1215.2) (19,1215.7) (20,1216.7) (21,1217.5) (22,1219.4) (23,1221.2) (24,1222.3) (25,1223.6) (26,1225.4) (27,1227.3) (28,1228.2) (29,1229.7) (30,1230.5) (31,1232.3) (32,1234) (33,1235.2) (34,1236.3) (35,1238) (36,1239.1) (37,1241) (38,1241.4) (39,1241.9) (40,1243.2) (41,1244.6) (42,1244.3) (43,1246.8) (44,1248.3) (45,1248.9) (46,1249.5) (47,1250.4) (48,1251) (49,1250.7) (50,1252.5) (51,1253.6) (52,1253.2) (53,1254.7) (54,1255.9) (55,1256.1) (56,1256.3) (57,1256.7) (58,1256.9) (59,1257.6) (60,1257.8) (61,1258.1) (62,1259.6) (63,1259.1)};
\addplot[roarstyle] coordinates {(-1,1222.4) (0,1222.6) (1,1220.9) (2,1219.6) (3,1218.1) (4,1216.4) (5,1216.6) (6,1215.8) (7,1214.8) (8,1213.7) (9,1213.1) (10,1213.1) (11,1212.8) (12,1213.3) (13,1213.5) (14,1212.5) (15,1212.7) (16,1212.9) (17,1212.6) (18,1214.1) (19,1214.9) (20,1215.6) (21,1216.3) (22,1217.4) (23,1218.6) (24,1219.1) (25,1220.2) (26,1221.8) (27,1223.2) (28,1223.4) (29,1224.4) (30,1225.3) (31,1227.4) (32,1228.9) (33,1230) (34,1231) (35,1232.7) (36,1233.3) (37,1234.8) (38,1235.4) (39,1236.1) (40,1237.1) (41,1238.4) (42,1238.3) (43,1240.6) (44,1241.6) (45,1242) (46,1242.8) (47,1243.8) (48,1244.4) (49,1244.3) (50,1246) (51,1246.5) (52,1246.5) (53,1248.1) (54,1249.4) (55,1249.6) (56,1249.8) (57,1250.3) (58,1250.6) (59,1251.1) (60,1251.3) (61,1251.8) (62,1253.1) (63,1253.8)};

\nextgroupplot[title={(c) Weather: baselines}, wcmp, legend to name=rowlegend]
\addplot[histstyle] coordinates {(-80,1262.1) (-79,1265) (-78,1266.5) (-77,1266.9) (-76,1266.2) (-75,1266.8) (-74,1267.7) (-73,1267.1) (-72,1268) (-71,1266.6) (-70,1265) (-69,1267.1) (-68,1268.3) (-67,1269) (-66,1270.1) (-65,1269.3) (-64,1269.3) (-63,1269.1) (-62,1267.9) (-61,1268.7) (-60,1270.7) (-59,1271.4) (-58,1271.9) (-57,1270.4) (-56,1270.2) (-55,1270.7) (-54,1272.2) (-53,1273.7) (-52,1273.2) (-51,1273) (-50,1272.1) (-49,1271.2) (-48,1273.2) (-47,1274.5) (-46,1274) (-45,1274.6) (-44,1274.4) (-43,1272.1) (-42,1271.2) (-41,1271.4) (-40,1271.9) (-39,1272.4) (-38,1271.5) (-37,1270.8) (-36,1272.5) (-35,1273.4) (-34,1272.5) (-33,1272.6) (-32,1273) (-31,1271.5) (-30,1271.6) (-29,1271.8) (-28,1271.9) (-27,1270.8) (-26,1269.3) (-25,1268.8) (-24,1267.3) (-23,1268.2) (-22,1266) (-21,1264.1) (-20,1261.2) (-19,1260) (-18,1260.2) (-17,1259) (-16,1254.8) (-15,1252.8) (-14,1251.7) (-13,1251.1) (-12,1249.5) (-11,1248.2) (-10,1246.4) (-9,1244.1) (-8,1241.5) (-7,1238.4) (-6,1236.5) (-5,1233.3) (-4,1229.7) (-3,1226.8) (-2,1223.5) (-1,1222.4)};
\addplot[gtstyle] coordinates {(-1,1222.4) (0,1221.2) (1,1219.7) (2,1218) (3,1215.9) (4,1214.7) (5,1214) (6,1212.7) (7,1210.1) (8,1208.2) (9,1207.6) (10,1209) (11,1209.9) (12,1210.4) (13,1209) (14,1207.2) (15,1209.4) (16,1211.3) (17,1212.1) (18,1213.6) (19,1214.6) (20,1215.5) (21,1217) (22,1217.5) (23,1218.3) (24,1219.5) (25,1221.3) (26,1222.5) (27,1223.6) (28,1224.3) (29,1226.9) (30,1228.4) (31,1228.9) (32,1228.7) (33,1229.3) (34,1231.1) (35,1232.3) (36,1233.1) (37,1233.7) (38,1233.9) (39,1235.7) (40,1237.8) (41,1240.8) (42,1242.5) (43,1241.9) (44,1241.7) (45,1241) (46,1240.8) (47,1241.6) (48,1243.6) (49,1244.8) (50,1245.1) (51,1245.5) (52,1245.1) (53,1247) (54,1247.2) (55,1247.4) (56,1247.5) (57,1248.2) (58,1249.4) (59,1250.8) (60,1251.9) (61,1253.2) (62,1253.2) (63,1252.6)};
\addplot[tsragstyle] coordinates {(-1,1222.4) (0,1224.1) (1,1223.1) (2,1221.7) (3,1221.1) (4,1220.8) (5,1218.1) (6,1215.2) (7,1214.7) (8,1216.7) (9,1216.8) (10,1213.6) (11,1213.3) (12,1214.3) (13,1215.2) (14,1216.1) (15,1216.9) (16,1215.8) (17,1218.4) (18,1221.4) (19,1222.6) (20,1224.8) (21,1223.9) (22,1225.7) (23,1227.6) (24,1230.1) (25,1229.4) (26,1232.3) (27,1233) (28,1234.4) (29,1236.6) (30,1234.8) (31,1234.7) (32,1236.3) (33,1238.3) (34,1238.4) (35,1238) (36,1240.3) (37,1241.4) (38,1244.5) (39,1245.1) (40,1246) (41,1248.1) (42,1250.2) (43,1251.3) (44,1253.4) (45,1250.8) (46,1252.2) (47,1253) (48,1253.1) (49,1253.6) (50,1256.8) (51,1256.4) (52,1256.9) (53,1257) (54,1258.3) (55,1257.9) (56,1259.5) (57,1260.4) (58,1258.4) (59,1257.9) (60,1257.4) (61,1259.6) (62,1260.4) (63,1260.4)};
\addplot[crossragstyle] coordinates {(-1,1222.4) (0,1225.8) (1,1224.1) (2,1222.7) (3,1221.4) (4,1221.1) (5,1220.8) (6,1219.6) (7,1218.6) (8,1217.6) (9,1217.2) (10,1216.6) (11,1215.8) (12,1215.9) (13,1216.1) (14,1216.5) (15,1216.7) (16,1217) (17,1217.4) (18,1218.8) (19,1220.1) (20,1220.9) (21,1221.7) (22,1222.3) (23,1224.2) (24,1225.6) (25,1226.8) (26,1227.6) (27,1229.1) (28,1230.5) (29,1232.2) (30,1233.2) (31,1233.6) (32,1235.1) (33,1236.7) (34,1237.7) (35,1237.8) (36,1238.8) (37,1239.8) (38,1240.6) (39,1241.8) (40,1242.8) (41,1243.8) (42,1244.8) (43,1246.3) (44,1247.8) (45,1248.8) (46,1249.2) (47,1250.8) (48,1251.9) (49,1252.3) (50,1253) (51,1254.2) (52,1255.1) (53,1256.1) (54,1256.7) (55,1257.9) (56,1259.4) (57,1259.8) (58,1259.5) (59,1260.4) (60,1259.7) (61,1260.7) (62,1262.3) (63,1263.4)};
\addlegendimage{ablstyle}
\addplot[roarstyle] coordinates {(-1,1222.4) (0,1222.6) (1,1220.9) (2,1219.6) (3,1218.1) (4,1216.4) (5,1216.6) (6,1215.8) (7,1214.8) (8,1213.7) (9,1213.1) (10,1213.1) (11,1212.8) (12,1213.3) (13,1213.5) (14,1212.5) (15,1212.7) (16,1212.9) (17,1212.6) (18,1214.1) (19,1214.9) (20,1215.6) (21,1216.3) (22,1217.4) (23,1218.6) (24,1219.1) (25,1220.2) (26,1221.8) (27,1223.2) (28,1223.4) (29,1224.4) (30,1225.3) (31,1227.4) (32,1228.9) (33,1230) (34,1231) (35,1232.7) (36,1233.3) (37,1234.8) (38,1235.4) (39,1236.1) (40,1237.1) (41,1238.4) (42,1238.3) (43,1240.6) (44,1241.6) (45,1242) (46,1242.8) (47,1243.8) (48,1244.4) (49,1244.3) (50,1246) (51,1246.5) (52,1246.5) (53,1248.1) (54,1249.4) (55,1249.6) (56,1249.8) (57,1250.3) (58,1250.6) (59,1251.1) (60,1251.3) (61,1251.8) (62,1253.1) (63,1253.8)};
\legend{,Ground Truth,TS-RAG,Cross-RAG,ROAR w/o Stage I \& II,ROAR (ours)}

\end{groupplot}
\path (group c1r1.outer south west) -- (group c3r1.outer south east) node[midway, below=2pt] {\pgfplotslegendfromname{rowlegend}};
\end{tikzpicture}

%% file: appendix.tex
\section*{Appendices: ROAR: Retrieval Opportunity-Aware Refinement for Zero-Shot Time Series Forecasting}

\section{Preliminary Study: Are Retrieved Neighbors Fully Exploited?}
\label{app:preliminary}

This section examines whether existing retrieval-augmented forecasters fully exploit the
information carried by retrieved neighbors. We study three questions:
(\textbf{Q1}) Whether retrieval contributes positively and whether the contribution stems from the
retrieved content;
(\textbf{Q2}) How much of the improvement offered by the retrieved candidates is realized; and
(\textbf{Q3}) Whether retrieval similarity indicates which candidates are useful.
The overall finding is that retrieval is beneficial, but existing methods realize only a small
part of its potential: the value of retrieved neighbors is highly uneven across samples, and
similarity cannot reveal it.

\subsection{Experimental Setup}
\label{app:pre_setup}

\paragraph{Models and data.}
We conduct the examination on two state-of-the-art retrieval-augmented zero-shot forecasting
methods, TS-RAG~\citep{ningts} and Cross-RAG~\citep{lee2026cross}, both reproduced with their
official configurations. We use the benchmarks of the main experiments and follow the official
evaluation procedure of both methods. Following their official code, each variable retrieves
only from its own training split of the target dataset. Unless stated otherwise, results are
equal-weight averages over ETTh1, ETTh2, ETTm1, ETTm2, Weather, Exchange, and Electricity.

\paragraph{Candidate forecasts and sample groups.}
For a test sample $i$ with retrieved neighbors $\{(\mathbf{x}^r_{ij},\mathbf{y}^r_{ij})\}_{j=1}^{K}$,
we take the level-aligned continuation $\mathbf{Z}^{\mathrm{lev}}_{ij}$ of Eq.~(2) as the $j$-th
candidate forecast. Let $\ell^0_i$ denote the MSE of the base forecast and
$\ell^{r,\min}_i=\min_j \ell_i(\mathbf{Z}^{\mathrm{lev}}_{ij})$ that of the best candidate.
We compute the opportunity score $g_i=[\ell^0_i-\ell^{r,\min}_i]_+/\ell^0_i$, which has the
form of Eq.~(7), and split the samples into
$\mathcal{G}^{+}=\{i: g_i>0\}$, where at least one candidate outperforms the base forecast,
and $\mathcal{G}^{-}=\{i: g_i=0\}$, where none does.

\paragraph{Metrics.}
MSE and MAE follow the official evaluation. For a set of samples, the improvement of a forecast
over the base forecast is $\Delta=(e_0-e)/e_0$, where $e_0$ and $e$ are the mean errors of the
base and the compared forecast on that set. Positive values indicate lower error.
The groups $\mathcal{G}^{\pm}$ and the best candidate are always defined by MSE; MAE is
reported on the same samples and candidates.

\subsection{Does Retrieval Contribute Positively?}
\label{app:pre_positive}

\paragraph{Method.}
We use two controls.
\emph{Candidate quality:} for each test sample, we draw $K$ training samples uniformly from its
retrieval library (three draws) and compare the share of samples that have a candidate better
than the base forecast with that obtained from the retrieved top-$K$ neighbors.
\emph{Neighbor shuffling:} we replace the neighbor set of each test sample with that of a
randomly chosen test sample of the same variable, keeping the model and all other inputs
fixed, and measure the overall improvement of each method.

\paragraph{Results.}
Table~\ref{tab:pre_positive} reports both controls.
TS-RAG's retrieved candidates outperform random ones on all seven datasets. Cross-RAG's
candidates outperform random ones on the ETT datasets and Electricity but fall behind on Weather
and Exchange, so on average they are on par with random draws.
After shuffling, the MSE improvement of both methods turns negative and MAE degrades further.
With the original neighbors, Cross-RAG improves MSE but leaves MAE essentially unchanged.

\begin{table}[h]
\centering
\caption{Contribution of retrieval. Top: share of samples with a candidate better than the base
forecast. Bottom: overall improvement $\Delta$ over the base forecast with the original and the
shuffled neighbor sets (seed 2021). Positive values indicate lower error.}
\label{tab:pre_positive}
\small
\setlength{\tabcolsep}{6pt}
\begin{tabular}{lcccc}
\toprule
& \multicolumn{2}{c}{TS-RAG} & \multicolumn{2}{c}{Cross-RAG} \\
\cmidrule(lr){2-3}\cmidrule(lr){4-5}
& MSE & MAE & MSE & MAE \\
\midrule
\multicolumn{5}{l}{\textit{Share of samples with a better candidate}} \\
\quad Retrieved top-$K$ & \multicolumn{2}{c}{46.3\%} & \multicolumn{2}{c}{35.3\%} \\
\quad Random draws from the library & \multicolumn{2}{c}{30.6\%} & \multicolumn{2}{c}{35.1\%} \\
\midrule
\multicolumn{5}{l}{\textit{Overall improvement $\Delta$}} \\
\quad Original neighbors & $+1.4\%$ & $+0.7\%$ & $+3.3\%$ & $-0.4\%$ \\
\quad Shuffled neighbors & $-0.7\%$ & $-1.2\%$ & $-12.9\%$ & $-14.0\%$ \\
\bottomrule
\end{tabular}
\end{table}

\paragraph{Conclusion.}
Retrieval contributes positively, and the contribution originates from the retrieved content
rather than from the model architecture.

\subsection{Is the Retrieval Gain Fully Realized?}
\label{app:pre_realized}

\paragraph{Method.}
On $\mathcal{G}^{+}$, we measure the improvement offered by the best candidate and define the
\emph{realization rate} as the error reduction achieved by the retrieval-augmented method
divided by that offered by the best candidate. Within $\mathcal{G}^{+}$ and $\mathcal{G}^{-}$,
we compare the retrieval-augmented and base forecasts sample by sample, and we measure the
share of the total available improvement contributed by the top $10\%$ of samples.
Because $\mathcal{G}^{\pm}$ are defined relative to the base forecast, grouping and evaluating
on the same horizon steps share the same noise and induce a regression-to-the-mean effect.
We therefore also define the groups on one half of the horizon (steps 1--32 or 33--64) and
evaluate on the other half, averaging both directions.

\paragraph{Results.}
Table~\ref{tab:pre_realized} summarizes the results (three seeds, averaged).
The best candidate reduces MSE by about half and MAE by about a third on $\mathcal{G}^{+}$,
yet TS-RAG and Cross-RAG realize only $5.1\%$ and $20.0\%$ of the MSE reduction
($5.9\%$ and $15.8\%$ for MAE). Even when a better candidate exists, retrieval worsens
$38\%$ and $27\%$ of the forecasts in MSE. The available improvement is highly concentrated:
the top $10\%$ of samples account for $80\%$ and $85\%$ of it.
Table~\ref{tab:pre_split} shows that, across halves, the degradation on $\mathcal{G}^{-}$
shrinks by about two thirds to three quarters in MSE and by about half to two thirds in MAE.
A large part of this degradation thus reflects regression to the mean, and we report it only
as a descriptive result.

\begin{table}[h]
\centering
\caption{Realization of the available improvement (three seeds, averaged).
$\Delta$ denotes the improvement over the base forecast; positive values indicate lower error.}
\label{tab:pre_realized}
\small
\setlength{\tabcolsep}{6pt}
\begin{tabular}{lcccc}
\toprule
& \multicolumn{2}{c}{TS-RAG} & \multicolumn{2}{c}{Cross-RAG} \\
\cmidrule(lr){2-3}\cmidrule(lr){4-5}
& MSE & MAE & MSE & MAE \\
\midrule
Share of $\mathcal{G}^{+}$ & \multicolumn{2}{c}{46.3\%} & \multicolumn{2}{c}{35.3\%} \\
$\Delta$ of the best candidate on $\mathcal{G}^{+}$ & $+51.2\%$ & $+31.1\%$ & $+46.1\%$ & $+27.2\%$ \\
Realization rate & 5.1\% & 5.9\% & 20.0\% & 15.8\% \\
Worsened samples in $\mathcal{G}^{+}$ & 38\% & 40\% & 27\% & 34\% \\
Worsened samples in $\mathcal{G}^{-}$ & 55\% & 56\% & 64\% & 66\% \\
Share of available gain from the top 10\% of samples & 80\% & 66\% & 85\% & 75\% \\
\bottomrule
\end{tabular}
\end{table}

\begin{table}[h]
\centering
\caption{Split-horizon control on $\mathcal{G}^{-}$. Improvement $\Delta$ of the
retrieval-augmented forecast when the groups are defined and evaluated on the same half of the
horizon or on different halves (three seeds, averaged).}
\label{tab:pre_split}
\small
\setlength{\tabcolsep}{6pt}
\begin{tabular}{lcccc}
\toprule
& \multicolumn{2}{c}{TS-RAG} & \multicolumn{2}{c}{Cross-RAG} \\
\cmidrule(lr){2-3}\cmidrule(lr){4-5}
Grouping and evaluation & MSE & MAE & MSE & MAE \\
\midrule
Same half & $-7.5\%$ & $-4.5\%$ & $-13.8\%$ & $-9.6\%$ \\
Different halves & $-2.4\%$ & $-1.6\%$ & $-3.5\%$ & $-5.1\%$ \\
\bottomrule
\end{tabular}
\end{table}

\paragraph{Conclusion.}
Existing methods realize only a small part of the improvement available from retrieval, and
this improvement is concentrated in a small share of samples.

\subsection{Can Retrieval Similarity Identify Useful Candidates?}
\label{app:pre_similarity}

\paragraph{Method.}
We examine how often the most similar candidate is the best one, compared with random choice,
and the Spearman correlation between the retrieval ranking and the candidate MSE ranking.
We further compute the AUROC of retrieval similarity and of other observable quantities
(candidate dispersion, deviation of the candidate mean from the base forecast, context
volatility, and the base MSE of the preceding sample) for identifying $\mathcal{G}^{+}$.
Finally, we measure the improvement obtained by replacing the base forecast with the most
similar candidate.

\paragraph{Results.}
As shown in Table~\ref{tab:pre_similarity}, the most similar candidate is the best one no more
often than a random choice, and the retrieval ranking is nearly uncorrelated with candidate
error. None of the observable quantities separates $\mathcal{G}^{+}$ from $\mathcal{G}^{-}$
(AUROC close to $0.5$), and using the most similar candidate directly degrades both metrics
severely.

\begin{table}[h]
\centering
\caption{Relation between retrieval similarity and candidate usefulness. Values in parentheses
are the expected rates under random choice.}
\label{tab:pre_similarity}
\small
\setlength{\tabcolsep}{6pt}
\begin{tabular}{lcccc}
\toprule
& \multicolumn{2}{c}{TS-RAG} & \multicolumn{2}{c}{Cross-RAG} \\
\cmidrule(lr){2-3}\cmidrule(lr){4-5}
& MSE & MAE & MSE & MAE \\
\midrule
Most similar candidate is the best & \multicolumn{2}{c}{10.5\% (10.0\%)} & \multicolumn{2}{c}{6.4\% (6.7\%)} \\
Spearman correlation (rank vs.\ MSE) & \multicolumn{2}{c}{0.04} & \multicolumn{2}{c}{0.04} \\
AUROC of observables for $\mathcal{G}^{+}$ & \multicolumn{2}{c}{0.45--0.53} & \multicolumn{2}{c}{0.46--0.53} \\
$\Delta$ of the most similar candidate & $-157\%$ & $-64.2\%$ & $-392\%$ & $-78.6\%$ \\
\bottomrule
\end{tabular}
\end{table}

\paragraph{Conclusion.}
Retrieval similarity indicates neither which candidate is useful nor which samples admit an
improvement, which limits how existing methods exploit the candidate set.

\subsection{Summary}
\label{app:pre_summary}

Retrieval contributes positively to forecasting, and the contribution stems from the retrieved
content (Section~\ref{app:pre_positive}). However, existing methods realize only a small
fraction of the improvement offered by the retrieved candidates, which is concentrated in a
small share of samples (Section~\ref{app:pre_realized}). Because retrieval similarity cannot
reveal which candidates are useful (Section~\ref{app:pre_similarity}), a refiner should learn
from the improvement each candidate offers over the base forecast, which motivates ROAR.
Two limitations remain. Cross-RAG's retrieved candidates are on average no better than random
draws, so its positive contribution rests mainly on the shuffling control. Moreover, no
observable feature identifies samples with an opportunity at inference time; whether a learned
gate can adapt to the input is examined in the main experiments.

\section{Proofs and Additional Analysis}
\label{app:roar_theory}

\subsection{From Candidate Opportunity to Forecast Improvement}
\label{app:opportunity_realization}
Fix a Stage-I query, its reference base forecast, and the
candidate weights and gate produced by the refiner.
The weights satisfy $p_{ik}\geq0$ and $\sum_kp_{ik}=1$,
and the gate satisfies $u_i\in[0,1]$.
Write $\mathbf{r}_i=\sum_kp_{ik}\mathbf{Z}_{ik}$ and
$\widehat{\mathbf{y}}_i
=(1-u_i)\widehat{\mathbf{y}}_i^{0,R}+u_i\mathbf{r}_i$.
Throughout this subsection, $\delta_i$ is the squared
point-forecast error summed over $\Omega_i$ and divided by $n_i$.

Expanding the squared errors around $\mathbf{r}_i$ and using
$\sum_kp_{ik}(\mathbf{Z}_{ik}-\mathbf{r}_i)=0$ gives
\begin{equation}
\label{eq:candidate_variance_identity}
\begin{aligned}
\delta_i(\mathbf{r}_i)
&=\sum_{k=1}^{J}p_{ik}\delta_i(\mathbf{Z}_{ik})
-\frac{1}{n_i}\sum_{k=1}^{J}p_{ik}
\|\mathbf{Z}_{ik}-\mathbf{r}_i\|_{\Omega_i}^{2}\\
&=\delta_i^{r,\min}+R_i^{\mathrm{agg}}
-\frac{1}{n_i}\sum_{k=1}^{J}p_{ik}
\|\mathbf{Z}_{ik}-\mathbf{r}_i\|_{\Omega_i}^{2}.
\end{aligned}
\end{equation}
Applying the same squared-error identity to the mixture of
the base forecast and proposal yields
\begin{equation}
\label{eq:base_proposal_mixture_identity}
\delta_i(\widehat{\mathbf{y}}_i)
=(1-u_i)\delta_i^0+u_i\delta_i(\mathbf{r}_i)
-\frac{u_i(1-u_i)}{n_i}
\|\mathbf{r}_i-\widehat{\mathbf{y}}_i^{0,R}\|_{\Omega_i}^{2}.
\end{equation}
Combining the two identities gives, for every query,
\begin{equation}
\label{eq:general_refinement_gain}
\delta_i^0-\delta_i(\widehat{\mathbf{y}}_i)
=u_i\big[\delta_i^0-\delta_i^{r,\min}-R_i^{\mathrm{agg}}\big]
+\frac{u_i}{n_i}\sum_{k=1}^{J}p_{ik}\|\mathbf{Z}_{ik}-\mathbf{r}_i\|_{\Omega_i}^{2}
+\frac{u_i(1-u_i)}{n_i}\|\mathbf{r}_i-\widehat{\mathbf{y}}_i^{0,R}\|_{\Omega_i}^{2}.
\end{equation}
When $g_i>0$, its definition implies
$\delta_i^0-\delta_i^{r,\min}
=(\delta_i^0+\epsilon s_i^2)g_i$.
Substitution proves Eq.~\ref{eq:realized_refinement_gain}.
Dropping the two nonnegative squared-norm terms gives
\begin{equation}
\label{eq:opportunity_realization_bound}
\frac{\delta_i^0-\delta_i(\widehat{\mathbf{y}}_i)}
{\delta_i^0+\epsilon s_i^2}
\geq
u_i\left(g_i-
\frac{R_i^{\mathrm{agg}}}{\delta_i^0+\epsilon s_i^2}\right).
\end{equation}
The denominator is positive because $s_i>0$ and $\epsilon>0$.
The stated sufficient condition makes the right-hand side
strictly positive.
The condition is not necessary, since the mixing terms can
provide additional gains.

\subsection{Training Allocation Across Queries}
\label{app:opportunity_allocation}
Fix a nonempty source mini-batch $\mathcal{B}$.
Recall that $s_i>0$, $\beta\geq0$, and $a_i=h_i g_i\geq0$.
Let $S_{\mathcal{B}}=\sum_{i\in\mathcal{B}}s_i^2$.
Since $w_i=s_i^2(1+\beta a_i)$, we have
$\sum_{i\in\mathcal{B}}w_i=S_{\mathcal{B}}(1+\beta m)$.
The normalized forecasting-loss weights therefore satisfy
\begin{equation}
\label{eq:normalized_query_weights}
\widetilde q_i
=\frac{w_i}{\sum_{k\in\mathcal{B}}w_k}
=\frac{q_i(1+\beta a_i)}{1+\beta m}.
\end{equation}
When $m>0$, $q_i^{\mathrm{opp}}=q_i a_i/m$ defines a
probability distribution. Substituting
$\alpha=\beta m/(1+\beta m)$ gives
$\widetilde q_i=(1-\alpha)q_i+\alpha q_i^{\mathrm{opp}}$.
Multiplying by $\ell_i$ and summing over $i$ proves
Eq.~\ref{eq:opportunity_allocation_summary}.
If $m=0$, positivity of $q_i$ and nonnegativity of $a_i$
imply $a_i=0$ for every query, so $\widetilde q_i=q_i$.
For all $m\geq0$, $\widetilde q_i>0$.
When $m>0$, $q_i^{\mathrm{opp}}>0$ exactly for queries
with $h_i g_i>0$.

The redistribution does not decrease the weighted mean opportunity:
\begin{equation}
\label{eq:opportunity_mass_shift}
\begin{aligned}
\sum_{i\in\mathcal{B}}\widetilde q_i a_i-m
&=\frac{\beta\left(
\sum_{i\in\mathcal{B}}q_i a_i^2-m^2\right)}{1+\beta m}\\
&=\frac{\beta\,\operatorname{Var}_{q}(a)}{1+\beta m}
\geq0.
\end{aligned}
\end{equation}
The increase is strict when $\beta>0$ and the opportunity
values are not constant across the batch.
These identities hold for any per-query forecasting loss
$\ell_i$. They apply to the forecasting term in
Eq.~\ref{eq:roar_objective}, while the auxiliary gate loss
retains its uniform averaging.

\subsection{Joint Calibration in the Linearized Prediction Space}
\label{app:joint_calibration}

\paragraph{Coordinates and Correction Spaces.}
Fix a nonempty source calibration set $\mathcal{S}$ and write
$\boldsymbol{\vartheta}=(\eta,\psi_u)$,
$\boldsymbol{\vartheta}^R=(\eta^R,\psi_u^R)$, and
$\lambda=\lambda_{\mathrm{ref}}\geq0$.
Define the scaled prediction map and target vector by
\begin{equation}
\label{eq:calibration_weighted_coordinates}
\mathbf{f}(\boldsymbol{\vartheta})
=\left[
\frac{\widehat y_{i,t}(\boldsymbol{\vartheta})}
{\sqrt{|\mathcal{S}|n_i}}
\right]_{i\in\mathcal{S},\,t\in\Omega_i},
\qquad
\mathbf{y}_{\mathcal{S}}
=\left[
\frac{y_{i,t}}{\sqrt{|\mathcal{S}|n_i}}
\right]_{i\in\mathcal{S},\,t\in\Omega_i}.
\end{equation}
Empty valid-entry sets contribute no coordinates and zero loss.
With $\mathbf{e}=\mathbf{y}_{\mathcal{S}}
-\mathbf{f}(\boldsymbol{\vartheta}^R)$, the calibration objective
on $\mathcal{S}$ is exactly
\begin{equation}
\label{eq:calibration_source_objective}
\mathcal{L}_{\mathcal{S}}(\boldsymbol{\vartheta})
=\|\mathbf{y}_{\mathcal{S}}-\mathbf{f}(\boldsymbol{\vartheta})\|^2
+\lambda\|\mathbf{f}(\boldsymbol{\vartheta})
-\mathbf{f}(\boldsymbol{\vartheta}^R)\|^2.
\end{equation}

Assume that $\mathbf{f}$ is differentiable at the reference.
Let $\mathbf{J}_\eta$ and $\mathbf{J}_u$ denote its Jacobian
blocks with respect to $\eta$ and $\psi_u$ at
$\boldsymbol{\vartheta}^R$.
Because the proposal and reference features remain fixed,
the per-query blocks before masking and scaling are
\begin{equation}
\label{eq:calibration_jacobian_blocks}
\begin{aligned}
\mathbf{J}_{\eta,i}
&=(1-u_i^R)
\left.\frac{\partial\widehat{\mathbf{y}}_i^0(\eta)}
{\partial\eta}\right|_{\eta^R},\\
\mathbf{J}_{u,i}
&=(\mathbf{r}_i^R-\widehat{\mathbf{y}}_i^{0,R})
\left(\left.\nabla_{\psi_u}u_i\right|_{\psi_u^R}\right)^\top.
\end{aligned}
\end{equation}
Restricting these blocks to $\Omega_i$, dividing by
$\sqrt{|\mathcal{S}|n_i}$, and stacking over queries yields
$\mathbf{J}_\eta$ and $\mathbf{J}_u$.
Thus forecasting-module updates change the base prediction
with sensitivity scaled by $1-u_i^R$, while gate updates act
along the fixed proposal direction for each query.

Writing $\mathbf{J}=[\mathbf{J}_\eta\;\mathbf{J}_u]$,
the linearized output correction for a parameter increment
$\mathbf{d}$ is $\boldsymbol{\Delta}=\mathbf{J}\mathbf{d}$.
The available correction spaces are
\begin{equation}
\label{eq:calibration_tangent_spaces}
\mathcal{V}_\eta=\operatorname{range}(\mathbf{J}_\eta),
\qquad
\mathcal{V}_u=\operatorname{range}(\mathbf{J}_u),
\qquad
\mathcal{V}_{\mathrm{joint}}
=\operatorname{range}(\mathbf{J})
=\mathcal{V}_\eta+\mathcal{V}_u.
\end{equation}

\paragraph{Optimal Correction and Forecasting Gain.}
For any of these spaces $\mathcal{V}$, define
\begin{equation}
\mathcal{Q}_{\mathcal{V}}(\boldsymbol{\Delta})
=\|\mathbf{e}-\boldsymbol{\Delta}\|^2
+\lambda\|\boldsymbol{\Delta}\|^2,
\qquad \boldsymbol{\Delta}\in\mathcal{V}.
\end{equation}
Since $\mathbf{e}-\mathbf{P}_{\mathcal{V}}\mathbf{e}$ is
orthogonal to $\mathcal{V}$, completing the square gives
\begin{equation}
\label{eq:calibration_geometry}
\mathcal{Q}_{\mathcal{V}}(\boldsymbol{\Delta})
=\|\mathbf{e}\|^2
-\frac{\|\mathbf{P}_{\mathcal{V}}\mathbf{e}\|^2}{1+\lambda}
+(1+\lambda)
\left\|\boldsymbol{\Delta}
-\frac{\mathbf{P}_{\mathcal{V}}\mathbf{e}}{1+\lambda}
\right\|^2.
\end{equation}
The unique optimal output correction is therefore
$\boldsymbol{\Delta}_{\mathcal{V}}^\star
=\mathbf{P}_{\mathcal{V}}\mathbf{e}/(1+\lambda)$.
Its forecasting MSE reduction is
\begin{equation}
\label{eq:linearized_calibration_gain}
\begin{aligned}
G(\mathcal{V})
&=2\langle\mathbf{e},\boldsymbol{\Delta}_{\mathcal{V}}^\star\rangle
-\|\boldsymbol{\Delta}_{\mathcal{V}}^\star\|^2\\
&=c_\lambda\|\mathbf{P}_{\mathcal{V}}\mathbf{e}\|^2,
\qquad
c_\lambda=\frac{1+2\lambda}{(1+\lambda)^2}.
\end{aligned}
\end{equation}
This proves Eq.~\ref{eq:calibration_optimum_summary}.

For $\mathcal{U}\subseteq\mathcal{V}_{\mathrm{joint}}$,
the operator
$\mathbf{P}_{\mathcal{V}_{\mathrm{joint}}}
-\mathbf{P}_{\mathcal{U}}$
is the orthogonal projector onto
$\mathcal{V}_{\mathrm{joint}}\cap\mathcal{U}^{\perp}$.
Orthogonality gives
\begin{equation}
\label{eq:nested_projection_identity}
\|\mathbf{P}_{\mathcal{V}_{\mathrm{joint}}}\mathbf{e}\|^2
=\|\mathbf{P}_{\mathcal{U}}\mathbf{e}\|^2
+\|(\mathbf{P}_{\mathcal{V}_{\mathrm{joint}}}
-\mathbf{P}_{\mathcal{U}})\mathbf{e}\|^2.
\end{equation}
Multiplying by $c_\lambda>0$ proves
Eq.~\ref{eq:joint_calibration_gain_summary} for either
$\mathcal{U}\in\{\mathcal{V}_\eta,\mathcal{V}_u\}$.
The gain is strict exactly when the residual projects
nontrivially onto the added orthogonal directions.

\subsection{Optimization Error and Nonlinear Updates}
\label{app:calibration_errors}
To quantify departures from the linearized optima, assume
that $\mathbf{f}$ is continuously differentiable in a convex
neighborhood of $\boldsymbol{\vartheta}^R$ and its Jacobian
is $L_J$-Lipschitz there in operator norm.
For a correction space $\mathcal{V}$, consider a parameter
increment $\mathbf{d}_{\mathcal{V}}$ whose segment from the
reference lies in this neighborhood.
For a single-component update, set the other parameter block
to zero.
The actual output correction satisfies
\begin{equation}
\label{eq:calibration_remainder}
\begin{aligned}
\boldsymbol{\Delta}_{\mathrm{act}}
(\mathbf{d}_{\mathcal{V}})
&=\mathbf{f}(\boldsymbol{\vartheta}^R+\mathbf{d}_{\mathcal{V}})
-\mathbf{f}(\boldsymbol{\vartheta}^R)\\
&=\mathbf{J}\mathbf{d}_{\mathcal{V}}
+\mathbf{T}(\mathbf{d}_{\mathcal{V}}),\\
\|\mathbf{T}(\mathbf{d}_{\mathcal{V}})\|
&\leq\frac{L_J}{2}\|\mathbf{d}_{\mathcal{V}}\|^2.
\end{aligned}
\end{equation}
The remainder bound follows by integrating the Jacobian
difference along the segment from the reference.

Suppose the increment has linearized objective gap
\begin{equation}
\label{eq:calibration_optimization_error}
\mathcal{Q}_{\mathcal{V}}(\mathbf{J}\mathbf{d}_{\mathcal{V}})
-\mathcal{Q}_{\mathcal{V}}(
\boldsymbol{\Delta}_{\mathcal{V}}^\star)
\leq\xi_{\mathcal{V}},
\qquad \xi_{\mathcal{V}}\geq0.
\end{equation}
Equation~\ref{eq:calibration_geometry} makes the left-hand
side equal to
$(1+\lambda)\|\mathbf{J}\mathbf{d}_{\mathcal{V}}
-\boldsymbol{\Delta}_{\mathcal{V}}^\star\|^2$.
Combining this identity with the remainder bound yields
\begin{equation}
\label{eq:calibration_output_error}
\|\boldsymbol{\Delta}_{\mathrm{act}}(\mathbf{d}_{\mathcal{V}})
-\boldsymbol{\Delta}_{\mathcal{V}}^\star\|
\leq\rho_{\mathcal{V}},
\qquad
\rho_{\mathcal{V}}
=\sqrt{\frac{\xi_{\mathcal{V}}}{1+\lambda}}
+\frac{L_J}{2}\|\mathbf{d}_{\mathcal{V}}\|^2.
\end{equation}

Define the actual forecasting gain as
\begin{equation}
G_{\mathrm{act}}(\mathbf{d}_{\mathcal{V}})
=\|\mathbf{e}\|^2
-\|\mathbf{e}-\boldsymbol{\Delta}_{\mathrm{act}}
(\mathbf{d}_{\mathcal{V}})\|^2.
\end{equation}
Let $\mathbf{b}_{\mathcal{V}}
=\boldsymbol{\Delta}_{\mathrm{act}}(\mathbf{d}_{\mathcal{V}})
-\boldsymbol{\Delta}_{\mathcal{V}}^\star$.
Expanding the squared error gives
\begin{equation}
G_{\mathrm{act}}(\mathbf{d}_{\mathcal{V}})-G(\mathcal{V})
=2\langle\mathbf{e}-\boldsymbol{\Delta}_{\mathcal{V}}^\star,
\mathbf{b}_{\mathcal{V}}\rangle
-\|\mathbf{b}_{\mathcal{V}}\|^2.
\end{equation}
Cauchy--Schwarz and Eq.~\ref{eq:calibration_output_error} imply
\begin{equation}
\label{eq:calibration_actual_gain_bound}
|G_{\mathrm{act}}(\mathbf{d}_{\mathcal{V}})-G(\mathcal{V})|
\leq E_{\mathcal{V}},
\qquad
E_{\mathcal{V}}
=2\|\mathbf{e}-\boldsymbol{\Delta}_{\mathcal{V}}^\star\|
\rho_{\mathcal{V}}+\rho_{\mathcal{V}}^2.
\end{equation}
Consequently, for joint and single-component increments
satisfying these conditions at the same reference and penalty
strength, and either
$\mathcal{U}\in\{\mathcal{V}_\eta,\mathcal{V}_u\}$,
\begin{equation}
\label{eq:nonlinear_joint_advantage}
G_{\mathrm{act}}(\mathbf{d}_{\mathcal{V}_{\mathrm{joint}}})
-G_{\mathrm{act}}(\mathbf{d}_{\mathcal{U}})
\geq c_\lambda
\|(\mathbf{P}_{\mathcal{V}_{\mathrm{joint}}}
-\mathbf{P}_{\mathcal{U}})\mathbf{e}\|^2
-E_{\mathcal{V}_{\mathrm{joint}}}-E_{\mathcal{U}}.
\end{equation}
Thus the joint-calibration advantage persists when the
additional projected-residual gain exceeds the combined
optimization and linearization error bounds.

\subsection{Correction from Observable Features}
\label{app:observable_gate}
Fix the reference base forecaster and learned candidate
aggregator. Let
$\mathbf{e}_i^{0,R}
=\mathbf{y}_i-\widehat{\mathbf{y}}_i^{0,R}$ and
$\mathbf{d}_i^R
=\mathbf{r}_i^R-\widehat{\mathbf{y}}_i^{0,R}$
denote the base residual and proposal direction.
Under a fixed query distribution, assume their squared norms
over valid entries have finite expectations after division
by $n_i$.
For the observable gate features $\boldsymbol{\phi}_i^R$,
define
\begin{equation}
\label{eq:conditional_correction_statistics}
\begin{aligned}
A(\boldsymbol{\phi})
&=\mathbb{E}\!\left[
\frac{\|\mathbf{d}_i^R\|_{\Omega_i}^{2}}{n_i}
\,\middle|\,
\boldsymbol{\phi}_i^R=\boldsymbol{\phi}
\right],\\
B(\boldsymbol{\phi})
&=\mathbb{E}\!\left[
\frac{\langle\mathbf{e}_i^{0,R},\mathbf{d}_i^R\rangle_{\Omega_i}}
{n_i}
\,\middle|\,
\boldsymbol{\phi}_i^R=\boldsymbol{\phi}
\right].
\end{aligned}
\end{equation}
For the prediction
$\widehat{\mathbf{y}}_i^{0,R}+u\mathbf{d}_i^R$,
the conditional MSE relative to the base forecast is
\begin{equation}
\label{eq:conditional_gate_risk}
\mathbb{E}\!\left[
\delta_i(\widehat{\mathbf{y}}_i^{0,R}+u\mathbf{d}_i^R)
-\delta_i^0
\,\middle|\,
\boldsymbol{\phi}_i^R=\boldsymbol{\phi}
\right]
=u^2 A(\boldsymbol{\phi})-2uB(\boldsymbol{\phi}).
\end{equation}
Minimizing this quadratic over $u\in[0,1]$ yields
\begin{equation}
\label{eq:optimal_observable_gate}
u^\star(\boldsymbol{\phi})
=\operatorname{clip}_{[0,1]}\!\left(
\frac{B(\boldsymbol{\phi})}{A(\boldsymbol{\phi})}
\right)
\end{equation}
when $A(\boldsymbol{\phi})>0$.
If $A(\boldsymbol{\phi})=0$, then $B(\boldsymbol{\phi})=0$,
and $u^\star=0$ is one minimizer.
For $A(\boldsymbol{\phi})>0$, a gate value that strictly
improves conditional MSE exists if and only if
$B(\boldsymbol{\phi})>0$.
This characterizes the MSE-optimal correction among gates
that are arbitrary measurable functions of the available
features. It provides a benchmark for the learned gate.

\newpage
\section{Additional Experiment Settings and Results}
\label{app_exp}

\subsection{Experimental Settings}
\label{app_exp_set}

\textbf{Evaluation Datasets} Statistics of evaluation datasets are presented in Table \ref{tab:dataset}.

\input{Tables/table_dataset}

\input{Tables/table_para}

\textbf{Comparing Baselines  }
We set two sets of baseline methods for comparison, including 7 TSFMs, 
Chronos \citep{ansari2024chronos}, Chronos-Bolt \citep{ansari2024chronos}, MOMENT \citep{goswami2024moment}, TTM \citep{ekambaram2024tiny}, Moirai \citep{woo2024unified}, TimesFM \citep{das2024decoder} and Time-MoE \citep{shitime}; and 5 RAG-based time series forecasting (TSF) methods, RAF \citep{tireretrieval}, TS-RAG \citep{ningts}, Cross-RAG \citep{lee2026cross}, PFRP \citep{du2026predicting} and RATL \citep{he2026ratl}. For RAG-based TSF methods, we use the same Chronos-Bolt backbone. PFRP and RATL were not designed for the zero-shot setting, so we adapt them to it. Both adapted methods use the same backbone and the same training and inference protocols as our method and the other baselines. We evaluate all methods with MSE and MAE.

\textbf{Implementation Details.}
We adopt Chronos-Bolt~\citep{ansari2024chronos} as the frozen backbone. Following~\citep{ningts}, the input length and forecasting horizon are set as 512 and 64, and the pre-training corpus, retrieval database, and remaining retrieval settings are the same as TS-RAG. The number of retrieved examples $K$ is set as 10, yielding $J=20$ candidates. The forecasting module is first trained for 10,000 steps, followed by 2,500 steps for Stage~I and 2,000 steps for Stage~II. For Chronos-Bolt, $\ell_i$ in Stage~I is the pinball loss over its quantile outputs, each shifted by the same correction. We use AdamW with a constant learning rate, which is set as $3\times10^{-4}$ for the scorer and gate, and $3\times10^{-5}$ for the forecasting module in Stage~II. The weight decay is set as 0.01, the dropout rate as 0.2, the batch size as 256, and the gradient clipping threshold as 1.0. We set $\beta=0.5$, $\lambda_{\mathrm{gate}}=1$, and $\lambda_{\mathrm{ref}}=15$. The threshold $t_h$ is set as the median normalized base error over 12,800 source-training queries before training. The sensitivity to $K$ and $\lambda_{\mathrm{ref}}$ is reported in Appendix~\ref{app:sensitivity}. All experiments are conducted on 2 GeForce RTX 4090 GPUs on an Ubuntu platform with 128GB of memory. More hyperparameter settings are listed in Table~\ref{tab:hyperparameters} in Appendix~\ref{app_exp}.

\subsection{Sensitivity Analysis}
\label{app:sensitivity}
In this section, we examine the parameter sensitivity on three key hyper-parameters: the number of retrieved sequences $K$, the reference penalty $\lambda_{\mathrm{ref}}$, and the gate weight $\lambda_{\text{gate}}$.
We follow the zero-shot protocol of the main experiments with the Chronos-Bolt backbone, and evaluate on ETTh1, ETTm1, Weather and Exchange. Specifically, we vary $K \in \{1,3,\dots,15\}$, $\lambda_{\mathrm{ref}} \in \{1,5,10,15,20\}$ and $\lambda_{\text{gate}} \in \{0.1,0.3,0.5,0.7,0.9,1\}$. All other settings, including the retrieval database and training configuration, are kept identical to those in the main experiments.
The results are reported in Fig.~\ref{fig:sensitivity}, where the upper and lower panels of each subplot show MAE and MSE, respectively.

\textbf{Number of retrieved sequences $K$.} As shown in Fig.~\ref{fig:sensitivity}(1), the forecasting error generally decreases as $K$ increases from $1$, since aggregating more neighbors provides a more reliable reference for the backbone. Compared with $K{=}1$, the best MSE is reduced by $3.2\%$, $3.2\%$, $4.2\%$ and $7.5\%$ on ETTh1, ETTm1, Weather and Exchange, respectively. Most of this gain is obtained by $K{=}5$ on ETTh1 and ETTm1 and by $K{=}7$ on Weather, whereas Exchange keeps improving until $K{=}11$. Beyond the best value, the error fluctuates slightly and rises by up to $2.1\%$ on Exchange, possibly because additional neighbors of lower relevance introduce noise.

\textbf{Reference penalty $\lambda_{\mathrm{ref}}$.} As shown in Fig.~\ref{fig:sensitivity}(2), ROAR is insensitive to $\lambda_{\mathrm{ref}}$ on ETTh1 and ETTm1, where the MSE varies by less than $0.5\%$ across the whole range. On Weather and Exchange, a too small penalty ($\lambda_{\mathrm{ref}}{=}1$) leads to noticeably higher errors, e.g., the MAE on Weather rises from $0.180$ to $0.185$ and the MSE on Exchange from $0.069$ to $0.076$, indicating that without sufficient regularization the fused prediction may deviate excessively from the reference. For $\lambda_{\mathrm{ref}} \geq 5$, the performance is stable on ETTh1, ETTm1 and Weather, while Exchange benefits slightly from a larger penalty.

\textbf{Gate weight $\lambda_{\text{gate}}$.} As shown in Fig.~\ref{fig:sensitivity}(3), increasing $\lambda_{\text{gate}}$ yields a mild but monotonic reduction in MSE on all datasets, with the largest relative gain of about $1.2\%$ on ETTm1, while the MAE remains almost unchanged. This suggests that the gating term mainly suppresses large errors, and that ROAR is robust to its exact weight.

Overall, ROAR maintains stable performance across a broad range of values for all three hyper-parameters. The only setting to avoid is an overly small reference penalty, and moderate choices of $K$ and $\lambda_{\mathrm{ref}}$ are sufficient to obtain near-optimal results, which indicates that ROAR does not rely on delicate hyper-parameter tuning.

\input{plot/experiments/sensitivity_final}   %

\newpage
\subsection{Efficiency analysis.}
To assess the computational overhead introduced by ROAR, we compare it with the RAG-based baselines in terms of per-iteration latency, FLOPs, total parameters and trainable parameters. All methods share the same Chronos-Bolt backbone and retrieval setting, and the latency is averaged over ETTh1, ETTm1 and Weather under identical hardware and batch size. We further report the two intermediate variants of ROAR to isolate the cost of each stage. The results are summarized in Table~\ref{tab:efficiency}.

Several observations can be made. First, the base variant of ROAR (w/o Stage~I \& II) is already lightweight: it has the fewest total parameters among all methods, and its trainable parameters (2.27M) are less than half of those of TS-RAG (4.78M) and about a quarter of those of Cross-RAG (8.71M). Second, Stage~I introduces almost no overhead. It keeps the FLOPs and total parameters unchanged, increases the latency by only 0.31\,ms, and involves merely 162 trainable parameters, since the backbone and fusion module are frozen and only two small MLPs are optimized. Third, the full ROAR roughly doubles the latency, FLOPs and total parameters compared with the other methods, because Stage~II requires an additional forward pass through the reference branch. Nevertheless, the number of trainable parameters remains 2.27M, and the absolute latency stays below 16\,ms per iteration.

Overall, the performance gains of ROAR come with a moderate and predictable inference cost, while its trainable parameters remain substantially fewer than those of existing RAG-based methods, making it practical for deployment in the zero-shot setting.

\begin{table}[h]
\centering
\caption{Efficiency comparisons across RAG-based TSF methods. Iteration time is averaged over ETTh1, ETTm1 and Weather.}
\footnotesize
\label{tab:efficiency}
\setlength{\tabcolsep}{5pt}
\renewcommand{\arraystretch}{1.15}
\begin{tabular}{l ccccc}
\toprule
\textbf{Method}
& \makecell{Avg Iter\\Time (ms)$\downarrow$}
& \makecell{FLOPs\\(G)$\downarrow$}
& \makecell{Total\\Params (M)$\downarrow$}
& \makecell{Trainable\\Params (M)$\downarrow$}
& \makecell{Trainable\\Ratio (\%)$\downarrow$} \\
\midrule
TS-RAG                    & 8.14  & 6.96  & 210.07 & 4.7754              & 2.273 \\
Cross-RAG                 & 7.42  & 6.96  & 214.01 & 8.7122              & 4.071 \\
\midrule
ROAR w/o Stage I \& II    & 7.69  & 6.92  & 207.56 & 2.2654              & 1.091 \\
ROAR w/o Stage II         & 8.00  & 6.92  & 207.56 & $1.6\times10^{-4}$  & $7.8\times10^{-5}$ \\
ROAR (full)               & 15.68 & 13.84 & 415.12 & 2.2655              & 0.546 \\
\bottomrule
\end{tabular}
\end{table}

\subsection{Visualizations}
\label{app_vis}

We present more detailed forecasting visualizations between our ROAR and the ablative version ``ROAR w/o Stages I \& II'' and baseline methods TS-RAG and Cross-RAG in Figs.\ref{fig:app-vis-ablation} and \ref{fig:app-vis-baselines}.

\input{plot/experiments/appendix_vis_ablation}
\input{plot/experiments/appendix_vis_baselines}

%% file: Tables/table_dataset.tex
\begin{table*}[ht]
\centering
\footnotesize
\caption{Statistics of evaluation datasets. ``\# Channels'' denotes the number of channels.}
\label{tab:dataset}
\setlength{\tabcolsep}{4.5pt}
\begin{tabular}{l|p{40pt}<{\centering}p{38pt}<{\centering}p{38pt}<{\centering}p{38pt}<{\centering}p{38pt}<{\centering}p{38pt}<{\centering}p{38pt}<{\centering}p{38pt}<{\centering}}
\toprule
\textbf{Dataset} & \textbf{ETTh1} & \textbf{ETTh2} & \textbf{ETTm1} & \textbf{ETTm2} & \textbf{Weather} & \textbf{\scriptsize Electricity} & \textbf{Exchange} \\
\midrule
\# Channels & 7 & 7 & 7 & 7 & 21 & 321 & 8 \\
Timestamps  & 17420 & 17420 & 69680 & 69680 & 52696  & 26304 & 7588\\
Frequency & 1 hour & 1 hour & 15 mins & 15 mins & 10 mins & 1 hour & 1 day  \\
Split ratio & 6:2:2 & 6:2:2 & 6:2:2 & 6:2:2 & 7:1:2 & 7:1:2 & 7:1:2 \\
\bottomrule
\end{tabular}
\end{table*}

%% file: Tables/table_para.tex
\begin{table}[h]
\centering
\caption{\textbf{Hyperparameter settings of ROAR} (Chronos-Bolt backbone).}
\label{tab:hyperparameters}
\footnotesize
\renewcommand{\arraystretch}{1.1}
\setlength{\tabcolsep}{5pt}
\begin{tabular}{l ccc}
\toprule
& \textbf{Base training} & \textbf{Stage I} & \textbf{Stage II} \\
\midrule
Trainable parameters & $\eta$ & $\psi_c,\psi_u$ & $\eta,\psi_u$ \\
Loss & Pinball & Eq.~\ref{eq:roar_objective} & Eq.~\ref{eq:roar_calibration} \\
Optimizer & AdamW & AdamW & AdamW \\
Learning rate & $3\times10^{-4}$ & $3\times10^{-4}$ & $\eta$: $3\times10^{-5}$;\ $\psi_u$: $3\times10^{-4}$ \\
LR schedule & -- & Constant & Constant \\
Weight decay & 0.01 & 0.01 & 0.01 \\
Batch size & 256 & 256 & 256 \\
Training steps & 10,000 & 2,500 & 2,000 \\
Dropout rate & 0.2 & 0.2 & 0.2 \\
Gradient clipping & 1.0 & 1.0 & 1.0 \\
\midrule
\multicolumn{4}{l}{\textit{Shared settings}} \\
Input length $L$ / horizon $H$ & \multicolumn{3}{c}{512 / 64} \\
Retrieved examples $K$  & \multicolumn{3}{c}{10} \\
\midrule
\multicolumn{4}{l}{\textit{ROAR-specific settings}} \\
Opportunity weight $\beta$ & \multicolumn{3}{c}{0.5} \\
Difficulty threshold $t_h$ & \multicolumn{3}{c}{Median normalized base error (0.1546)} \\
Gate supervision weight $\lambda_{\mathrm{gate}}$ & \multicolumn{3}{c}{1} \\
Reference penalty $\lambda_{\mathrm{ref}}$ & \multicolumn{3}{c}{15} \\
Scorer / gate architecture & \multicolumn{3}{c}{MLP ($3\!\to\!16\!\to\!1$)} \\
Initial gate value $u_i$ & \multicolumn{3}{c}{$\approx 0.1$ (bias $-2.197$)} \\
IQR quantiles in $\boldsymbol{\phi}_i^R$ & \multicolumn{3}{c}{0.2 / 0.8} \\
\bottomrule
\end{tabular}
\end{table}

%% file: plot/experiments/sensitivity_final.tex
\definecolor{colETTh1}{HTML}{E377C2}
\definecolor{colETTm1}{HTML}{2CA02C}
\definecolor{colWeather}{HTML}{1F9E9E}
\definecolor{colExchange}{HTML}{7B7BC9}

\newlength{\brkgap}\setlength{\brkgap}{4pt}
\newcommand{\brkmark}[2]{%
  \draw[black, line width=0.6pt] ($(#1)+(-2.5pt,-1.5pt)$) -- ++(5pt,3pt);
  \draw[black, line width=0.6pt] ($(#2)+(-2.5pt,-1.5pt)$) -- ++(5pt,3pt);}

\pgfplotsset{
  rbase/.style={
    width=5.2cm, height=1.0cm, scale only axis,
    axis y line*=left,
    tick align=outside, tick style={black}, major tick length=2pt,
    grid=major, grid style={gray!20},
    enlarge x limits=0.05,
    tick label style={font=\scriptsize},
    label style={font=\scriptsize},
    ylabel style={yshift=-4pt},
    yticklabel style={/pgf/number format/fixed, /pgf/number format/precision=3,
                      /pgf/number format/fixed zerofill},
    scaled y ticks=false,
    max space between ticks=16,
    clip=false,
  },
  rtop/.style={rbase,
    axis x line*=bottom, x axis line style={draw=none},
    xticklabels={}, x tick style={draw=none},
    ylabel={MAE},
    title style={font=\footnotesize\bfseries, yshift=-2mm},
  },
  rbot/.style={rbase, axis x line*=bottom, ylabel={MSE}},
  msestyle/.style={solid, mark=*, line width=1.1pt, mark size=1.4pt},
  maestyle/.style={dashed, mark=*, mark options={solid}, line width=0.9pt, mark size=1.2pt, opacity=0.6},
}

\begin{figure}[p]
\centering
\begin{subfigure}{\textwidth}
\centering
\begin{tikzpicture}
\begin{axis}[rtop, name=ETTh1top, at={(0cm,0cm)}, anchor=south west,
  title={ETTh1}, xtick={1,3,5,7,9,11,13,15}, ymin=0.36224, ymax=0.37736]
\addplot[maestyle, colETTh1] coordinates { (1,0.3726) (3,0.3698) (5,0.3684) (7,0.3687) (9,0.3670) (11,0.3674) (13,0.3676) (15,0.3676) };
\end{axis}
\begin{axis}[rbot, name=ETTh1bot, at={(ETTh1top.south west)}, anchor=north west, yshift=-\brkgap,
  xtick={1,3,5,7,9,11,13,15}, ymin=0.34154, ymax=0.35666]
\addplot[msestyle, colETTh1] coordinates { (1,0.3547) (3,0.3490) (5,0.3473) (7,0.3468) (9,0.3450) (11,0.3443) (13,0.3449) (15,0.3435) };
\end{axis}
\brkmark{ETTh1top.south west}{ETTh1bot.north west}
\begin{axis}[rtop, name=ETTm1top, at={(6.7cm,0cm)}, anchor=south west,
  title={ETTm1}, xtick={1,3,5,7,9,11,13,15}, ymin=0.30551, ymax=0.31779]
\addplot[maestyle, colETTm1] coordinates { (1,0.3140) (3,0.3114) (5,0.3098) (7,0.3095) (9,0.3095) (11,0.3093) (13,0.3101) (15,0.3105) };
\end{axis}
\begin{axis}[rbot, name=ETTm1bot, at={(ETTm1top.south west)}, anchor=north west, yshift=-\brkgap,
  xtick={1,3,5,7,9,11,13,15}, ymin=0.27191, ymax=0.28419]
\addplot[msestyle, colETTm1] coordinates { (1,0.2826) (3,0.2779) (5,0.2750) (7,0.2746) (9,0.2738) (11,0.2735) (13,0.2756) (15,0.2745) };
\end{axis}
\brkmark{ETTm1top.south west}{ETTm1bot.north west}
\begin{axis}[rtop, name=Weathertop, at={(0cm,-3.25cm)}, anchor=south west,
  title={Weather}, xtick={1,3,5,7,9,11,13,15}, ymin=0.17817, ymax=0.18654]
\addplot[maestyle, colWeather] coordinates { (1,0.1845) (3,0.1822) (5,0.1820) (7,0.1802) (9,0.1813) (11,0.1802) (13,0.1803) (15,0.1806) };
\end{axis}
\begin{axis}[rbot, name=Weatherbot, at={(Weathertop.south west)}, anchor=north west, yshift=-\brkgap,
  xtick={1,3,5,7,9,11,13,15}, ymin=0.13982, ymax=0.14819]
\addplot[msestyle, colWeather] coordinates { (1,0.1471) (3,0.1441) (5,0.1433) (7,0.1416) (9,0.1420) (11,0.1412) (13,0.1409) (15,0.1412) };
\end{axis}
\brkmark{Weathertop.south west}{Weatherbot.north west}
\begin{axis}[rtop, name=Exchangetop, at={(6.7cm,-3.25cm)}, anchor=south west,
  title={Exchange}, xtick={1,3,5,7,9,11,13,15}, ymin=0.17398, ymax=0.18262]
\addplot[maestyle, colExchange] coordinates { (1,0.1815) (3,0.1792) (5,0.1776) (7,0.1772) (9,0.1761) (11,0.1751) (13,0.1765) (15,0.1767) };
\end{axis}
\begin{axis}[rbot, name=Exchangebot, at={(Exchangetop.south west)}, anchor=north west, yshift=-\brkgap,
  xtick={1,3,5,7,9,11,13,15}, ymin=0.06353, ymax=0.07217]
\addplot[msestyle, colExchange] coordinates { (1,0.0705) (3,0.0685) (5,0.0668) (7,0.0669) (9,0.0661) (11,0.0652) (13,0.0660) (15,0.0666) };
\end{axis}
\brkmark{Exchangetop.south west}{Exchangebot.north west}
\end{tikzpicture}
\caption{Number of retrieved sequences $K$}
\label{fig:sens-topk}
\end{subfigure}
\vspace{2pt}

\begin{subfigure}{\textwidth}
\centering
\begin{tikzpicture}
\begin{axis}[rtop, name=ETTh1top, at={(0cm,0cm)}, anchor=south west,
  title={ETTh1}, xtick={1,5,10,15,20}, ymin=0.36105, ymax=0.36645]
\addplot[maestyle, colETTh1] coordinates { (1,0.3635) (5,0.3634) (10,0.3637) (15,0.3639) (20,0.3641) };
\end{axis}
\begin{axis}[rbot, name=ETTh1bot, at={(ETTh1top.south west)}, anchor=north west, yshift=-\brkgap,
  xtick={1,5,10,15,20}, ymin=0.33680, ymax=0.34220]
\addplot[msestyle, colETTh1] coordinates { (1,0.3393) (5,0.3391) (10,0.3395) (15,0.3397) (20,0.3399) };
\end{axis}
\brkmark{ETTh1top.south west}{ETTh1bot.north west}
\begin{axis}[rtop, name=ETTm1top, at={(6.7cm,0cm)}, anchor=south west,
  title={ETTm1}, xtick={1,5,10,15,20}, ymin=0.30665, ymax=0.31205]
\addplot[maestyle, colETTm1] coordinates { (1,0.3099) (5,0.3088) (10,0.3090) (15,0.3092) (20,0.3093) };
\end{axis}
\begin{axis}[rbot, name=ETTm1bot, at={(ETTm1top.south west)}, anchor=north west, yshift=-\brkgap,
  xtick={1,5,10,15,20}, ymin=0.27170, ymax=0.27710]
\addplot[msestyle, colETTm1] coordinates { (1,0.2738) (5,0.2743) (10,0.2747) (15,0.2749) (20,0.2750) };
\end{axis}
\brkmark{ETTm1top.south west}{ETTm1bot.north west}
\begin{axis}[rtop, name=Weathertop, at={(0cm,-3.25cm)}, anchor=south west,
  title={Weather}, xtick={1,5,10,15,20}, ymin=0.17698, ymax=0.18562]
\addplot[maestyle, colWeather] coordinates { (1,0.1845) (5,0.1796) (10,0.1786) (15,0.1782) (20,0.1781) };
\end{axis}
\begin{axis}[rbot, name=Weatherbot, at={(Weathertop.south west)}, anchor=north west, yshift=-\brkgap,
  xtick={1,5,10,15,20}, ymin=0.13723, ymax=0.14587]
\addplot[msestyle, colWeather] coordinates { (1,0.1419) (5,0.1412) (10,0.1412) (15,0.1412) (20,0.1412) };
\end{axis}
\brkmark{Weathertop.south west}{Weatherbot.north west}
\begin{axis}[rtop, name=Exchangetop, at={(6.7cm,-3.25cm)}, anchor=south west,
  title={Exchange}, xtick={1,5,10,15,20}, ymin=0.17477, ymax=0.19043]
\addplot[maestyle, colExchange] coordinates { (1,0.1884) (5,0.1795) (10,0.1776) (15,0.1770) (20,0.1768) };
\end{axis}
\begin{axis}[rbot, name=Exchangebot, at={(Exchangetop.south west)}, anchor=north west, yshift=-\brkgap,
  xtick={1,5,10,15,20}, ymin=0.06322, ymax=0.07888]
\addplot[msestyle, colExchange] coordinates { (1,0.0756) (5,0.0687) (10,0.0672) (15,0.0667) (20,0.0665) };
\end{axis}
\brkmark{Exchangetop.south west}{Exchangebot.north west}
\end{tikzpicture}
\caption{Reference penalty $\lambda_{\mathrm{ref}}$}
\label{fig:sens-dev}
\end{subfigure}
\vspace{2pt}

\begin{subfigure}{\textwidth}
\centering
\begin{tikzpicture}
\begin{axis}[rtop, name=ETTh1top, at={(0cm,0cm)}, anchor=south west,
  title={ETTh1}, xtick={0.1,0.3,0.5,0.7,0.9,1}, ymin=0.36040, ymax=0.36580]
\addplot[maestyle, colETTh1] coordinates { (0.1,0.3628) (0.3,0.3629) (0.5,0.3631) (0.7,0.3632) (0.9,0.3633) (1,0.3634) };
\end{axis}
\begin{axis}[rbot, name=ETTh1bot, at={(ETTh1top.south west)}, anchor=north west, yshift=-\brkgap,
  xtick={0.1,0.3,0.5,0.7,0.9,1}, ymin=0.33715, ymax=0.34255]
\addplot[msestyle, colETTh1] coordinates { (0.1,0.3406) (0.3,0.3400) (0.5,0.3397) (0.7,0.3394) (0.9,0.3392) (1,0.3391) };
\end{axis}
\brkmark{ETTh1top.south west}{ETTh1bot.north west}
\begin{axis}[rtop, name=ETTm1top, at={(6.7cm,0cm)}, anchor=south west,
  title={ETTm1}, xtick={0.1,0.3,0.5,0.7,0.9,1}, ymin=0.30620, ymax=0.31160]
\addplot[maestyle, colETTm1] coordinates { (0.1,0.3090) (0.3,0.3089) (0.5,0.3088) (0.7,0.3088) (0.9,0.3088) (1,0.3088) };
\end{axis}
\begin{axis}[rbot, name=ETTm1bot, at={(ETTm1top.south west)}, anchor=north west, yshift=-\brkgap,
  xtick={0.1,0.3,0.5,0.7,0.9,1}, ymin=0.27325, ymax=0.27865]
\addplot[msestyle, colETTm1] coordinates { (0.1,0.2776) (0.3,0.2763) (0.5,0.2755) (0.7,0.2749) (0.9,0.2745) (1,0.2743) };
\end{axis}
\brkmark{ETTm1top.south west}{ETTm1bot.north west}
\begin{axis}[rtop, name=Weathertop, at={(0cm,-3.25cm)}, anchor=south west,
  title={Weather}, xtick={0.1,0.3,0.5,0.7,0.9,1}, ymin=0.17705, ymax=0.18245]
\addplot[maestyle, colWeather] coordinates { (0.1,0.1799) (0.3,0.1798) (0.5,0.1797) (0.7,0.1796) (0.9,0.1796) (1,0.1796) };
\end{axis}
\begin{axis}[rbot, name=Weatherbot, at={(Weathertop.south west)}, anchor=north west, yshift=-\brkgap,
  xtick={0.1,0.3,0.5,0.7,0.9,1}, ymin=0.13905, ymax=0.14445]
\addplot[msestyle, colWeather] coordinates { (0.1,0.1423) (0.3,0.1419) (0.5,0.1416) (0.7,0.1414) (0.9,0.1413) (1,0.1412) };
\end{axis}
\brkmark{Weathertop.south west}{Weatherbot.north west}
\begin{axis}[rtop, name=Exchangetop, at={(6.7cm,-3.25cm)}, anchor=south west,
  title={Exchange}, xtick={0.1,0.3,0.5,0.7,0.9,1}, ymin=0.17700, ymax=0.18240]
\addplot[maestyle, colExchange] coordinates { (0.1,0.1799) (0.3,0.1798) (0.5,0.1796) (0.7,0.1796) (0.9,0.1795) (1,0.1795) };
\end{axis}
\begin{axis}[rbot, name=Exchangebot, at={(Exchangetop.south west)}, anchor=north west, yshift=-\brkgap,
  xtick={0.1,0.3,0.5,0.7,0.9,1}, ymin=0.06625, ymax=0.07165]
\addplot[msestyle, colExchange] coordinates { (0.1,0.0692) (0.3,0.0690) (0.5,0.0689) (0.7,0.0688) (0.9,0.0687) (1,0.0687) };
\end{axis}
\brkmark{Exchangetop.south west}{Exchangebot.north west}
\end{tikzpicture}
\caption{Gate weight $\lambda_{\text{gate}}$}
\label{fig:sens-lambda}
\end{subfigure}
\caption{Hyper-parameter sensitivity of ROAR. In each panel, the upper and lower parts show MAE and MSE, respectively.}
\label{fig:sensitivity}
\end{figure}

%% file: plot/experiments/appendix_vis_ablation.tex
\definecolor{apxTs}{HTML}{4E79A7}
\definecolor{apxCr}{HTML}{59A14F}
\definecolor{apxRo}{HTML}{E15759}
\definecolor{apxAb}{HTML}{B07AA1}
\pgfplotsset{
  apxpanel/.style={
    scale only axis, width=3.1cm, height=2.5cm,
    title style={font=\small, yshift=-1mm},
    label style={font=\footnotesize}, tick label style={font=\scriptsize},
    axis line style={black!60}, tick style={black!60}, grid style={black!10},
    xlabel={Time step}, ylabel={Value}, grid=major, no markers, xmin=-80, xmax=64,
    yticklabel style={/pgf/number format/1000 sep={}}, axis background/.style={fill=gray!5},
    legend style={font=\footnotesize, draw=none, fill=none, legend columns=-1,
                  /tikz/every even column/.append style={column sep=8pt}},
    legend cell align=left,
  },
  apxhist/.style={gray!70, line width=0.9pt, forget plot},
  apxgt/.style={black, dashed, line width=1.2pt, line join=round},
  apxabl/.style={apxAb, line width=1.1pt},
  apxts/.style={apxTs, line width=1.0pt},
  apxcr/.style={apxCr, line width=1.0pt},
  apxro/.style={apxRo, line width=1.4pt},
}
\begin{figure}[t]
\centering
\begin{tikzpicture}
\begin{groupplot}[group style={group size=3 by 2, horizontal sep=1.6cm, vertical sep=1.9cm}, apxpanel]
\nextgroupplot[title={ETTh2: OT}, legend to name=apxlegendapp-vis-ablation]
\addplot[apxhist] coordinates {(-80,3.0665) (-79,3.0665) (-78,3.0665) (-77,3.0665) (-76,3.0665) (-75,3.286) (-74,2.627) (-73,1.7485) (-72,1.3085) (-71,0.4295) (-70,-0.0095) (-69,-0.4495) (-68,-0.8885) (-67,-0.8885) (-66,1.089) (-65,5.044) (-64,9.878) (-63,13.833) (-62,15.591) (-61,15.811) (-60,15.151) (-59,14.273) (-58,12.954) (-57,11.196) (-56,9.2185) (-55,7.9005) (-54,7.0215) (-53,6.582) (-52,6.582) (-51,6.582) (-50,6.3625) (-49,6.1425) (-48,5.923) (-47,5.7035) (-46,5.2635) (-45,5.044) (-44,4.3845) (-43,3.9455) (-42,5.2635) (-41,8.3395) (-40,12.514) (-39,15.811) (-38,15.371) (-37,14.492) (-36,15.591) (-35,14.932) (-34,13.613) (-33,12.075) (-32,10.757) (-31,9.878) (-30,9.4385) (-29,8.7795) (-28,8.12) (-27,7.6805) (-26,7.0215) (-25,6.582) (-24,5.7035) (-23,4.8245) (-22,3.9455) (-21,3.0665) (-20,2.1875) (-19,2.1875) (-18,3.9455) (-17,7.241) (-16,10.537) (-15,13.613) (-14,13.613) (-13,14.932) (-12,12.954) (-11,11.416) (-10,9.658) (-9,8.12) (-8,7.241) (-7,6.582) (-6,5.7035) (-5,5.2635) (-4,4.3845) (-3,3.9455) (-2,2.847) (-1,1.968)};
\addplot[apxgt] coordinates {(-1,1.968) (0,1.3085) (1,0.4295) (2,-0.0095008) (3,-0.4495) (4,-0.8885) (5,-1.1085) (6,1.089) (7,5.044) (8,9.4385) (9,12.075) (10,13.833) (11,14.711) (12,14.712) (13,14.053) (14,12.295) (15,10.098) (16,8.7795) (17,8.12) (18,7.6805) (19,7.241) (20,6.3625) (21,5.7035) (22,4.6045) (23,3.7255) (24,3.0665) (25,2.1875) (26,1.7485) (27,1.3085) (28,0.6495) (29,0.4295) (30,2.4075) (31,5.4835) (32,8.999) (33,12.295) (34,13.614) (35,16.25) (36,16.031) (37,14.932) (38,12.954) (39,11.196) (40,9.658) (41,8.34) (42,7.241) (43,6.3625) (44,5.7035) (45,5.044) (46,4.6045) (47,4.165) (48,3.5055) (49,3.0665) (50,2.627) (51,2.1875) (52,1.7485) (53,1.968) (54,3.9455) (55,6.802) (56,9.4385) (57,12.514) (58,13.833) (59,16.031) (60,16.25) (61,15.371) (62,13.394) (63,11.196)};
\addplot[apxabl] coordinates {(-1,1.968) (0,0.7388) (1,-0.28128) (2,-1.0723) (3,-1.4227) (4,-2.1201) (5,-1.8477) (6,0.42689) (7,4.562) (8,8.544) (9,12.025) (10,13.623) (11,14.47) (12,13.891) (13,12.864) (14,10.82) (15,8.8433) (16,7.1367) (17,5.889) (18,5.1901) (19,4.738) (20,4.3149) (21,4.3171) (22,3.9985) (23,3.1545) (24,2.8853) (25,2.46) (26,1.6845) (27,0.93899) (28,0.45377) (29,0.45736) (30,2.2928) (31,5.9201) (32,8.7589) (33,11.941) (34,13.076) (35,13.898) (36,13.526) (37,12.709) (38,10.876) (39,9.1702) (40,7.7226) (41,6.7272) (42,5.868) (43,5.7318) (44,5.4257) (45,5.4086) (46,5.0321) (47,4.543) (48,4.3007) (49,4.0203) (50,2.9508) (51,2.3401) (52,1.8737) (53,2.0961) (54,3.2541) (55,6.5795) (56,9.6853) (57,12.394) (58,13.341) (59,13.782) (60,13.591) (61,12.975) (62,11.475) (63,9.9399)};
\addplot[apxro] coordinates {(-1,1.968) (0,0.64322) (1,-0.33568) (2,-1.0807) (3,-1.4236) (4,-1.9719) (5,-1.5537) (6,0.94638) (7,5.0595) (8,9.232) (9,12.687) (10,14.376) (11,15.371) (12,14.897) (13,13.673) (14,11.523) (15,9.4677) (16,7.7444) (17,6.3774) (18,5.6144) (19,5.1402) (20,4.6686) (21,4.5186) (22,4.0958) (23,3.2605) (24,2.9087) (25,2.3983) (26,1.6737) (27,1.0346) (28,0.52312) (29,0.69002) (30,2.775) (31,6.5587) (32,9.5905) (33,12.8) (34,14.189) (35,15.171) (36,14.762) (37,13.864) (38,11.928) (39,10.113) (40,8.466) (41,7.3118) (42,6.4645) (43,6.1464) (44,5.8096) (45,5.693) (46,5.3212) (47,4.8545) (48,4.5374) (49,4.1845) (50,3.2638) (51,2.653) (52,2.206) (53,2.5504) (54,4.0403) (55,7.3026) (56,10.479) (57,13.238) (58,14.323) (59,14.956) (60,14.77) (61,14.078) (62,12.403) (63,10.908)};
\legend{Ground Truth,ROAR w/o Stage I \& II,ROAR (ours)}
\nextgroupplot[title={ETTm1: LUFL}]
\addplot[apxhist] coordinates {(-80,4.386) (-79,4.477) (-78,4.538) (-77,4.569) (-76,4.569) (-75,4.752) (-74,4.873) (-73,5.087) (-72,5.148) (-71,5.269) (-70,5.148) (-69,5.148) (-68,5.178) (-67,5.208) (-66,5.452) (-65,5.269) (-64,5.178) (-63,5.026) (-62,4.995) (-61,4.843) (-60,4.721) (-59,4.569) (-58,4.447) (-57,4.264) (-56,4.173) (-55,4.021) (-54,3.868) (-53,3.746) (-52,3.503) (-51,3.35) (-50,3.259) (-49,3.076) (-48,2.955) (-47,2.894) (-46,2.772) (-45,2.68) (-44,2.619) (-43,2.559) (-42,2.345) (-41,2.406) (-40,2.376) (-39,2.315) (-38,2.315) (-37,2.284) (-36,2.254) (-35,2.193) (-34,2.224) (-33,2.224) (-32,2.193) (-31,2.193) (-30,2.132) (-29,2.163) (-28,2.132) (-27,2.163) (-26,2.132) (-25,2.193) (-24,2.254) (-23,2.284) (-22,2.254) (-21,2.224) (-20,2.345) (-19,2.345) (-18,2.528) (-17,2.528) (-16,2.68) (-15,2.863) (-14,3.168) (-13,3.442) (-12,3.655) (-11,3.686) (-10,3.807) (-9,3.777) (-8,3.929) (-7,3.868) (-6,3.838) (-5,3.838) (-4,3.96) (-3,4.051) (-2,4.234) (-1,4.508)};
\addplot[apxgt] coordinates {(-1,4.508) (0,4.63) (1,4.691) (2,4.447) (3,4.325) (4,4.203) (5,4.203) (6,4.173) (7,4.325) (8,4.264) (9,4.264) (10,4.264) (11,4.264) (12,4.234) (13,4.325) (14,4.295) (15,4.264) (16,4.234) (17,4.295) (18,4.203) (19,4.325) (20,4.386) (21,4.538) (22,4.721) (23,4.904) (24,4.965) (25,5.026) (26,4.995) (27,5.056) (28,4.995) (29,5.178) (30,5.3) (31,5.3) (32,5.208) (33,5.239) (34,5.087) (35,5.026) (36,4.873) (37,4.782) (38,4.66) (39,4.508) (40,4.264) (41,4.142) (42,4.051) (43,3.746) (44,3.686) (45,3.564) (46,3.29) (47,3.229) (48,3.076) (49,2.985) (50,2.894) (51,2.741) (52,2.711) (53,2.619) (54,2.589) (55,2.528) (56,2.437) (57,2.437) (58,2.437) (59,2.376) (60,2.315) (61,2.284) (62,2.315) (63,2.284)};
\addplot[apxabl] coordinates {(-1,4.508) (0,4.6135) (1,4.5205) (2,4.4186) (3,4.2677) (4,4.2073) (5,4.1788) (6,4.2014) (7,4.2444) (8,4.2217) (9,4.2401) (10,4.2265) (11,4.1856) (12,4.1984) (13,4.2064) (14,4.1814) (15,4.1706) (16,4.2164) (17,4.262) (18,4.3094) (19,4.3495) (20,4.4127) (21,4.5349) (22,4.6922) (23,4.8485) (24,5.0232) (25,5.0384) (26,5.0214) (27,5.0183) (28,5.0437) (29,5.1022) (30,5.1574) (31,5.1908) (32,5.1122) (33,5.0064) (34,4.8648) (35,4.7587) (36,4.6718) (37,4.4953) (38,4.3271) (39,4.1855) (40,4.0442) (41,3.8851) (42,3.7014) (43,3.5379) (44,3.4228) (45,3.2707) (46,3.1704) (47,2.9936) (48,2.963) (49,2.8568) (50,2.7109) (51,2.6313) (52,2.5416) (53,2.52) (54,2.4848) (55,2.4285) (56,2.3658) (57,2.3491) (58,2.3507) (59,2.3401) (60,2.3308) (61,2.2818) (62,2.2815) (63,2.2729)};
\addplot[apxro] coordinates {(-1,4.508) (0,4.6436) (1,4.567) (2,4.4758) (3,4.3196) (4,4.2374) (5,4.206) (6,4.2224) (7,4.2505) (8,4.2292) (9,4.2363) (10,4.2636) (11,4.1914) (12,4.1972) (13,4.2059) (14,4.1785) (15,4.1746) (16,4.1984) (17,4.2357) (18,4.2864) (19,4.3437) (20,4.4003) (21,4.5274) (22,4.6864) (23,4.8569) (24,5.0155) (25,5.0514) (26,5.0468) (27,5.0346) (28,5.05) (29,5.1156) (30,5.1992) (31,5.2348) (32,5.1947) (33,5.1127) (34,5.0061) (35,4.9123) (36,4.8449) (37,4.681) (38,4.5181) (39,4.3753) (40,4.2317) (41,4.0807) (42,3.895) (43,3.7272) (44,3.5986) (45,3.4602) (46,3.3709) (47,3.2141) (48,3.1652) (49,3.0562) (50,2.9167) (51,2.8237) (52,2.7221) (53,2.6838) (54,2.6421) (55,2.5897) (56,2.5498) (57,2.5207) (58,2.5037) (59,2.4852) (60,2.4712) (61,2.4214) (62,2.4178) (63,2.402)};
\nextgroupplot[title={ETTm1: MUFL}]
\addplot[apxhist] coordinates {(-80,10.803) (-79,10.589) (-78,10.696) (-77,11.62) (-76,9.701) (-75,10.625) (-74,10.732) (-73,12.366) (-72,11.62) (-71,11.655) (-70,13.041) (-69,11.584) (-68,12.082) (-67,10.305) (-66,8.742) (-65,8.386) (-64,7.533) (-63,5.828) (-62,0.32) (-61,-4.264) (-60,-3.731) (-59,-6.61) (-58,-9.275) (-57,-9.808) (-56,-11.371) (-55,-11.194) (-54,-11.655) (-53,-12.117) (-52,-13.788) (-51,-12.757) (-50,-7.143) (-49,-15.351) (-48,-15.422) (-47,-14.569) (-46,-15.458) (-45,-11.94) (-44,-2.949) (-43,-12.793) (-42,-11.584) (-41,-10.625) (-40,-9.097) (-39,-7.676) (-38,-6.005) (-37,-4.691) (-36,-1.564) (-35,1.492) (-34,2.701) (-33,3.873) (-32,4.726) (-31,5.543) (-30,6.041) (-29,5.934) (-28,5.543) (-27,5.97) (-26,5.97) (-25,6.076) (-24,6.183) (-23,6.432) (-22,6.361) (-21,6.183) (-20,6.041) (-19,5.97) (-18,5.934) (-17,5.721) (-16,5.508) (-15,5.366) (-14,5.295) (-13,5.543) (-12,7.071) (-11,6.929) (-10,7.356) (-9,6.823) (-8,7.285) (-7,7.64) (-6,7.604) (-5,8.351) (-4,9.346) (-3,10.199) (-2,8.528) (-1,9.097)};
\addplot[apxgt] coordinates {(-1,9.097) (0,9.061) (1,9.061) (2,8.99) (3,9.132) (4,8.351) (5,8.99) (6,9.239) (7,8.99) (8,8.066) (9,7.995) (10,8.884) (11,8.493) (12,8.173) (13,7.711) (14,8.422) (15,8.706) (16,7.995) (17,8.635) (18,8.244) (19,8.671) (20,7.782) (21,8.422) (22,8.742) (23,8.138) (24,8.955) (25,9.239) (26,9.808) (27,15.031) (28,15.067) (29,15.422) (30,14.711) (31,12.544) (32,11.265) (33,10.554) (34,6.823) (35,2.665) (36,-0.533) (37,-1.67) (38,-3.482) (39,-4.513) (40,-5.33) (41,-4.939) (42,-6.361) (43,-6.61) (44,-8.742) (45,-10.234) (46,-11.655) (47,-11.94) (48,-12.295) (49,-9.914) (50,-11.122) (51,-10.305) (52,-11.869) (53,-1.812) (54,-7.391) (55,2.132) (56,-5.153) (57,-3.98) (58,-0.213) (59,-0.711) (60,2.701) (61,6.503) (62,7.64) (63,9.026)};
\addplot[apxabl] coordinates {(-1,9.097) (0,8.5985) (1,8.6509) (2,8.9466) (3,8.774) (4,8.9951) (5,8.6006) (6,8.9352) (7,9.0682) (8,8.7069) (9,8.5514) (10,8.9181) (11,8.83) (12,9.0714) (13,8.8816) (14,8.9884) (15,8.9411) (16,8.429) (17,8.2726) (18,8.5624) (19,8.4578) (20,8.2096) (21,7.5965) (22,8.2069) (23,8.9026) (24,9.599) (25,9.3093) (26,9.4296) (27,10.248) (28,9.671) (29,8.4932) (30,7.4615) (31,6.9296) (32,5.6501) (33,2.8779) (34,-0.99927) (35,-4.531) (36,-5.9353) (37,-8.1783) (38,-10.372) (39,-11.455) (40,-11.863) (41,-12.228) (42,-12.659) (43,-14.019) (44,-14.313) (45,-14.145) (46,-13.083) (47,-13.79) (48,-14.037) (49,-14.018) (50,-13.773) (51,-12.797) (52,-12.595) (53,-12.537) (54,-11.878) (55,-10.81) (56,-9.6439) (57,-8.3133) (58,-6.7028) (59,-5.1181) (60,-1.4122) (61,1.1445) (62,2.7639) (63,3.6386)};
\addplot[apxro] coordinates {(-1,9.097) (0,8.7167) (1,8.6603) (2,8.977) (3,8.7039) (4,8.8452) (5,8.4612) (6,8.8387) (7,9.0749) (8,8.747) (9,8.5153) (10,8.8392) (11,8.9109) (12,9.1635) (13,9.0621) (14,9.0844) (15,9.0681) (16,8.6635) (17,8.6328) (18,9.0061) (19,8.9521) (20,8.6908) (21,8.2185) (22,8.8149) (23,9.3784) (24,9.9333) (25,9.6972) (26,9.8399) (27,10.332) (28,9.6005) (29,8.4641) (30,7.5342) (31,6.7896) (32,5.906) (33,3.5016) (34,0.23017) (35,-3.0265) (36,-4.0182) (37,-5.8926) (38,-7.8466) (39,-8.5164) (40,-8.7494) (41,-9.081) (42,-9.3087) (43,-10.856) (44,-11.233) (45,-10.907) (46,-10.306) (47,-11.12) (48,-11.68) (49,-11.202) (50,-11.187) (51,-10.438) (52,-10.072) (53,-10.497) (54,-9.7438) (55,-8.6704) (56,-7.4759) (57,-6.5075) (58,-5.2361) (59,-3.7454) (60,-0.72795) (61,1.5017) (62,2.8456) (63,3.6468)};
\nextgroupplot[title={ETTm2: OT}]
\addplot[apxhist] coordinates {(-80,34.268) (-79,34.048) (-78,33.389) (-77,32.95) (-76,32.29) (-75,31.851) (-74,30.972) (-73,30.312) (-72,29.653) (-71,28.994) (-70,28.335) (-69,27.676) (-68,27.236) (-67,26.797) (-66,26.357) (-65,25.698) (-64,25.259) (-63,24.819) (-62,24.38) (-61,24.16) (-60,23.721) (-59,23.281) (-58,23.281) (-57,22.842) (-56,22.622) (-55,22.402) (-54,22.402) (-53,22.183) (-52,21.963) (-51,21.963) (-50,21.523) (-49,21.523) (-48,21.303) (-47,21.084) (-46,21.084) (-45,20.865) (-44,20.645) (-43,20.425) (-42,20.205) (-41,20.205) (-40,19.986) (-39,19.766) (-38,19.546) (-37,19.326) (-36,19.326) (-35,19.107) (-34,18.887) (-33,18.887) (-32,18.667) (-31,18.447) (-30,18.447) (-29,18.228) (-28,18.228) (-27,18.008) (-26,18.008) (-25,18.008) (-24,17.788) (-23,17.788) (-22,17.568) (-21,17.568) (-20,17.568) (-19,17.568) (-18,17.568) (-17,17.568) (-16,17.788) (-15,18.228) (-14,18.667) (-13,19.326) (-12,19.986) (-11,21.084) (-10,21.963) (-9,22.842) (-8,23.721) (-7,24.6) (-6,25.698) (-5,26.796) (-4,27.236) (-3,28.555) (-2,28.994) (-1,30.093)};
\addplot[apxgt] coordinates {(-1,30.093) (0,30.752) (1,30.972) (2,31.631) (3,32.07) (4,32.95) (5,33.608) (6,34.268) (7,34.926) (8,35.146) (9,35.146) (10,35.366) (11,35.366) (12,35.585) (13,35.585) (14,35.366) (15,34.927) (16,34.927) (17,34.487) (18,34.048) (19,33.828) (20,33.389) (21,32.729) (22,31.851) (23,31.411) (24,30.752) (25,29.873) (26,29.434) (27,28.555) (28,28.115) (29,27.676) (30,27.236) (31,26.797) (32,26.357) (33,25.918) (34,25.479) (35,25.039) (36,25.039) (37,24.6) (38,24.16) (39,23.941) (40,23.721) (41,23.501) (42,23.281) (43,23.281) (44,23.062) (45,22.842) (46,22.622) (47,22.402) (48,22.402) (49,21.963) (50,21.963) (51,21.743) (52,21.523) (53,21.523) (54,21.523) (55,21.084) (56,21.084) (57,21.084) (58,20.645) (59,20.645) (60,20.645) (61,20.205) (62,20.205) (63,20.205)};
\addplot[apxabl] coordinates {(-1,30.093) (0,31.27) (1,31.496) (2,31.986) (3,32.711) (4,33.213) (5,33.557) (6,34.285) (7,34.819) (8,35.082) (9,35.041) (10,35.483) (11,35.735) (12,35.858) (13,35.683) (14,35.311) (15,35.069) (16,34.789) (17,34.445) (18,33.741) (19,33.356) (20,32.871) (21,32.339) (22,31.173) (23,30.5) (24,30.192) (25,29.653) (26,29.027) (27,28.333) (28,27.633) (29,27.129) (30,26.81) (31,26.413) (32,26.196) (33,25.83) (34,25.382) (35,24.917) (36,24.569) (37,24.341) (38,23.958) (39,23.727) (40,23.617) (41,23.291) (42,23.09) (43,22.551) (44,22.283) (45,22.37) (46,21.98) (47,21.85) (48,21.929) (49,21.924) (50,21.57) (51,21.507) (52,21.388) (53,21.163) (54,21.022) (55,20.84) (56,20.635) (57,20.52) (58,20.36) (59,20.294) (60,19.995) (61,19.754) (62,19.638) (63,19.502)};
\addplot[apxro] coordinates {(-1,30.093) (0,31.315) (1,31.556) (2,32.067) (3,32.712) (4,33.131) (5,33.506) (6,34.184) (7,34.673) (8,34.89) (9,34.991) (10,35.336) (11,35.607) (12,35.696) (13,35.703) (14,35.364) (15,35.083) (16,34.767) (17,34.42) (18,33.885) (19,33.452) (20,32.989) (21,32.494) (22,31.466) (23,30.821) (24,30.475) (25,29.966) (26,29.393) (27,28.596) (28,27.963) (29,27.489) (30,27.198) (31,26.762) (32,26.611) (33,26.18) (34,25.717) (35,25.293) (36,24.939) (37,24.776) (38,24.39) (39,24.108) (40,24.033) (41,23.711) (42,23.554) (43,23.017) (44,22.81) (45,22.859) (46,22.552) (47,22.415) (48,22.447) (49,22.358) (50,22.118) (51,21.984) (52,21.828) (53,21.663) (54,21.55) (55,21.425) (56,21.217) (57,21.173) (58,20.983) (59,21.031) (60,20.773) (61,20.623) (62,20.433) (63,20.24)};
\nextgroupplot[title={Weather: Tlog}]
\addplot[apxhist] coordinates {(-80,10.49) (-79,10.47) (-78,10.44) (-77,10.4) (-76,10.36) (-75,10.32) (-74,10.29) (-73,10.27) (-72,10.26) (-71,10.25) (-70,10.24) (-69,10.24) (-68,10.24) (-67,10.24) (-66,10.21) (-65,10.17) (-64,10.15) (-63,10.13) (-62,10.11) (-61,10.08) (-60,10.06) (-59,10.06) (-58,10.06) (-57,10.06) (-56,10.08) (-55,10.09) (-54,10.09) (-53,10.1) (-52,10.1) (-51,10.13) (-50,10.17) (-49,10.22) (-48,10.28) (-47,10.38) (-46,10.53) (-45,10.73) (-44,10.99) (-43,11.28) (-42,11.58) (-41,11.92) (-40,12.33) (-39,12.76) (-38,13.21) (-37,13.65) (-36,14.06) (-35,14.5) (-34,14.95) (-33,15.4) (-32,15.82) (-31,16.21) (-30,16.56) (-29,16.87) (-28,17.16) (-27,17.47) (-26,17.78) (-25,18.11) (-24,18.47) (-23,18.87) (-22,19.32) (-21,19.82) (-20,20.3) (-19,20.78) (-18,21.28) (-17,21.77) (-16,22.25) (-15,22.71) (-14,23.13) (-13,23.53) (-12,23.89) (-11,24.22) (-10,24.49) (-9,24.69) (-8,24.81) (-7,24.85) (-6,24.85) (-5,24.81) (-4,24.72) (-3,24.57) (-2,24.34) (-1,24.05)};
\addplot[apxgt] coordinates {(-1,24.05) (0,23.74) (1,23.4) (2,23.01) (3,22.59) (4,22.19) (5,21.79) (6,21.39) (7,21.01) (8,20.65) (9,20.35) (10,20.07) (11,19.79) (12,19.51) (13,19.25) (14,19.01) (15,18.8) (16,18.6) (17,18.41) (18,18.25) (19,18.09) (20,17.9) (21,17.67) (22,17.44) (23,17.21) (24,17.02) (25,16.85) (26,16.68) (27,16.54) (28,16.41) (29,16.29) (30,16.16) (31,16.03) (32,15.92) (33,15.82) (34,15.69) (35,15.56) (36,15.43) (37,15.31) (38,15.19) (39,15.08) (40,14.97) (41,14.86) (42,14.76) (43,14.64) (44,14.52) (45,14.4) (46,14.31) (47,14.22) (48,14.11) (49,13.99) (50,13.88) (51,13.78) (52,13.67) (53,13.57) (54,13.47) (55,13.39) (56,13.32) (57,13.28) (58,13.24) (59,13.17) (60,13.1) (61,13.01) (62,12.93) (63,12.85)};
\addplot[apxabl] coordinates {(-1,24.05) (0,24.11) (1,23.761) (2,23.338) (3,23.06) (4,22.563) (5,21.959) (6,21.621) (7,21.367) (8,20.84) (9,20.43) (10,20.167) (11,19.769) (12,19.217) (13,18.981) (14,18.64) (15,18.444) (16,17.934) (17,17.646) (18,17.379) (19,17.322) (20,17.33) (21,17.13) (22,16.646) (23,16.589) (24,16.624) (25,16.369) (26,16.007) (27,15.873) (28,15.775) (29,15.46) (30,15.315) (31,14.997) (32,14.756) (33,14.536) (34,14.326) (35,14.263) (36,13.962) (37,13.851) (38,13.74) (39,13.626) (40,13.302) (41,13.233) (42,13.341) (43,13.033) (44,13.003) (45,12.923) (46,12.762) (47,12.699) (48,12.705) (49,12.826) (50,12.627) (51,12.617) (52,12.472) (53,12.489) (54,12.354) (55,12.157) (56,12.084) (57,12.099) (58,11.974) (59,11.851) (60,11.812) (61,11.669) (62,11.601) (63,11.608)};
\addplot[apxro] coordinates {(-1,24.05) (0,23.958) (1,23.645) (2,23.29) (3,23.013) (4,22.508) (5,21.948) (6,21.673) (7,21.417) (8,20.961) (9,20.586) (10,20.329) (11,19.938) (12,19.474) (13,19.253) (14,18.908) (15,18.706) (16,18.226) (17,17.945) (18,17.743) (19,17.669) (20,17.682) (21,17.477) (22,17.056) (23,16.974) (24,16.992) (25,16.777) (26,16.519) (27,16.377) (28,16.237) (29,15.899) (30,15.81) (31,15.526) (32,15.386) (33,15.149) (34,14.951) (35,14.883) (36,14.589) (37,14.511) (38,14.47) (39,14.364) (40,14.123) (41,14.085) (42,14.174) (43,13.883) (44,13.926) (45,13.835) (46,13.729) (47,13.682) (48,13.668) (49,13.766) (50,13.632) (51,13.595) (52,13.433) (53,13.445) (54,13.362) (55,13.223) (56,13.179) (57,13.198) (58,13.116) (59,13.056) (60,13.017) (61,12.888) (62,12.765) (63,12.715)};
\nextgroupplot[title={Weather: Tpot}]
\addplot[apxhist] coordinates {(-80,273.66) (-79,273.49) (-78,273.27) (-77,273.44) (-76,273.43) (-75,273.47) (-74,273.72) (-73,273.58) (-72,273.15) (-71,273.01) (-70,272.91) (-69,273.22) (-68,273.27) (-67,273.16) (-66,272.86) (-65,272.56) (-64,272.67) (-63,272.7) (-62,272.9) (-61,273.1) (-60,272.68) (-59,272.42) (-58,272.52) (-57,272.37) (-56,272.42) (-55,272.88) (-54,273.06) (-53,273.03) (-52,272.93) (-51,272.82) (-50,273) (-49,273.14) (-48,272.78) (-47,272.61) (-46,272.81) (-45,272.78) (-44,272.71) (-43,273.01) (-42,273) (-41,273) (-40,272.99) (-39,273.21) (-38,273.54) (-37,273.67) (-36,273.99) (-35,273.8) (-34,274.26) (-33,274.67) (-32,275.28) (-31,275.55) (-30,275.52) (-29,275.76) (-28,276.66) (-27,277.08) (-26,277.33) (-25,277.44) (-24,277.78) (-23,278.05) (-22,278.42) (-21,278.94) (-20,279.49) (-19,280.18) (-18,280.6) (-17,281.33) (-16,282.12) (-15,282.8) (-14,283.52) (-13,283.79) (-12,284.03) (-11,284.37) (-10,284.74) (-9,285.21) (-8,285.45) (-7,285.61) (-6,285.91) (-5,286.48) (-4,286.9) (-3,287.05) (-2,286.71) (-1,286.48)};
\addplot[apxgt] coordinates {(-1,286.48) (0,286.37) (1,286.68) (2,287.1) (3,286.57) (4,286.1) (5,285.92) (6,285.6) (7,285.35) (8,285.14) (9,284.82) (10,284.68) (11,284.5) (12,284.22) (13,283.79) (14,283.5) (15,283.24) (16,283.08) (17,282.5) (18,282.17) (19,282.1) (20,282.17) (21,282.05) (22,281.65) (23,281.37) (24,281.14) (25,281) (26,280.97) (27,280.59) (28,280.12) (29,279.45) (30,279.04) (31,279.2) (32,279.24) (33,279.38) (34,279.44) (35,279.25) (36,278.83) (37,278.56) (38,278.49) (39,278.41) (40,278.51) (41,278.07) (42,278.04) (43,278.01) (44,277.99) (45,277.85) (46,277.62) (47,277.32) (48,277.1) (49,276.82) (50,276.82) (51,276.95) (52,277.09) (53,277.23) (54,277.3) (55,276.74) (56,276.34) (57,276.15) (58,275.89) (59,275.95) (60,275.71) (61,275.88) (62,276.02) (63,276.56)};
\addplot[apxabl] coordinates {(-1,286.48) (0,286.87) (1,286.71) (2,286.5) (3,286.5) (4,286.31) (5,285.73) (6,285.65) (7,285.52) (8,285.18) (9,284.97) (10,284.99) (11,284.59) (12,284.18) (13,284.11) (14,283.8) (15,283.56) (16,283.08) (17,282.86) (18,282.63) (19,282.3) (20,282.32) (21,282.02) (22,281.5) (23,281.21) (24,281.42) (25,281.11) (26,280.68) (27,280.37) (28,280.27) (29,279.76) (30,279.69) (31,279.36) (32,279.16) (33,278.94) (34,278.62) (35,278.53) (36,278.2) (37,277.93) (38,277.74) (39,277.55) (40,277.2) (41,276.99) (42,277.14) (43,276.6) (44,276.65) (45,276.35) (46,276.08) (47,275.85) (48,275.85) (49,275.9) (50,275.68) (51,275.63) (52,275.48) (53,275.36) (54,275.27) (55,275.07) (56,274.97) (57,274.99) (58,274.78) (59,274.74) (60,274.7) (61,274.45) (62,274.36) (63,274.44)};
\addplot[apxro] coordinates {(-1,286.48) (0,286.61) (1,286.44) (2,286.3) (3,286.27) (4,286.05) (5,285.54) (6,285.46) (7,285.3) (8,285.01) (9,284.86) (10,284.8) (11,284.43) (12,284.11) (13,284.09) (14,283.81) (15,283.59) (16,283.13) (17,282.92) (18,282.85) (19,282.63) (20,282.65) (21,282.37) (22,281.89) (23,281.62) (24,281.71) (25,281.43) (26,281.16) (27,280.9) (28,280.78) (29,280.32) (30,280.3) (31,280) (32,279.83) (33,279.66) (34,279.31) (35,279.14) (36,278.75) (37,278.57) (38,278.44) (39,278.27) (40,277.99) (41,277.78) (42,277.89) (43,277.39) (44,277.47) (45,277.22) (46,276.89) (47,276.66) (48,276.61) (49,276.57) (50,276.44) (51,276.33) (52,276.16) (53,276.02) (54,275.95) (55,275.75) (56,275.63) (57,275.67) (58,275.46) (59,275.46) (60,275.41) (61,275.2) (62,275.06) (63,275.05)};
\end{groupplot}
\path (group c1r2.outer south west) -- (group c3r2.outer south east) node[midway, below=2pt] {\pgfplotslegendfromname{apxlegendapp-vis-ablation}};
\end{tikzpicture}
\caption{Additional forecasting visualizations comparing ROAR with the ablative variant ROAR w/o Stage I \& II on ETTh2, ETTm1, ETTm2 and Weather.}
\label{fig:app-vis-ablation}
\end{figure}

%% file: plot/experiments/appendix_vis_baselines.tex
\definecolor{apxTs}{HTML}{4E79A7}
\definecolor{apxCr}{HTML}{59A14F}
\definecolor{apxRo}{HTML}{E15759}
\definecolor{apxAb}{HTML}{B07AA1}
\pgfplotsset{
  apxpanel/.style={
    scale only axis, width=3.1cm, height=2.5cm,
    title style={font=\small, yshift=-1mm},
    label style={font=\footnotesize}, tick label style={font=\scriptsize},
    axis line style={black!60}, tick style={black!60}, grid style={black!10},
    xlabel={Time step}, ylabel={Value}, grid=major, no markers, xmin=-80, xmax=64,
    yticklabel style={/pgf/number format/1000 sep={}}, axis background/.style={fill=gray!5},
    legend style={font=\footnotesize, draw=none, fill=none, legend columns=-1,
                  /tikz/every even column/.append style={column sep=8pt}},
    legend cell align=left,
  },
  apxhist/.style={gray!70, line width=0.9pt, forget plot},
  apxgt/.style={black, dashed, line width=1.2pt, line join=round},
  apxabl/.style={apxAb, line width=1.1pt},
  apxts/.style={apxTs, line width=1.0pt},
  apxcr/.style={apxCr, line width=1.0pt},
  apxro/.style={apxRo, line width=1.4pt},
}
\begin{figure}[t]
\centering
\begin{tikzpicture}
\begin{groupplot}[group style={group size=3 by 2, horizontal sep=1.6cm, vertical sep=1.9cm}, apxpanel]
\nextgroupplot[title={ETTh2: OT}, legend to name=apxlegendapp-vis-baselines]
\addplot[apxhist] coordinates {(-80,3.0665) (-79,3.0665) (-78,3.0665) (-77,3.0665) (-76,3.0665) (-75,3.0665) (-74,3.286) (-73,2.627) (-72,1.7485) (-71,1.3085) (-70,0.4295) (-69,-0.0095) (-68,-0.4495) (-67,-0.8885) (-66,-0.8885) (-65,1.089) (-64,5.044) (-63,9.878) (-62,13.833) (-61,15.591) (-60,15.811) (-59,15.151) (-58,14.273) (-57,12.954) (-56,11.196) (-55,9.2185) (-54,7.9005) (-53,7.0215) (-52,6.582) (-51,6.582) (-50,6.582) (-49,6.3625) (-48,6.1425) (-47,5.923) (-46,5.7035) (-45,5.2635) (-44,5.044) (-43,4.3845) (-42,3.9455) (-41,5.2635) (-40,8.3395) (-39,12.514) (-38,15.811) (-37,15.371) (-36,14.492) (-35,15.591) (-34,14.932) (-33,13.613) (-32,12.075) (-31,10.757) (-30,9.878) (-29,9.4385) (-28,8.7795) (-27,8.12) (-26,7.6805) (-25,7.0215) (-24,6.582) (-23,5.7035) (-22,4.8245) (-21,3.9455) (-20,3.0665) (-19,2.1875) (-18,2.1875) (-17,3.9455) (-16,7.241) (-15,10.537) (-14,13.613) (-13,13.613) (-12,14.932) (-11,12.954) (-10,11.416) (-9,9.658) (-8,8.12) (-7,7.241) (-6,6.582) (-5,5.7035) (-4,5.2635) (-3,4.3845) (-2,3.9455) (-1,2.847)};
\addplot[apxgt] coordinates {(-1,2.847) (0,1.968) (1,1.3085) (2,0.4295) (3,-0.0095008) (4,-0.4495) (5,-0.8885) (6,-1.1085) (7,1.089) (8,5.044) (9,9.4385) (10,12.075) (11,13.833) (12,14.711) (13,14.712) (14,14.053) (15,12.295) (16,10.098) (17,8.7795) (18,8.12) (19,7.6805) (20,7.241) (21,6.3625) (22,5.7035) (23,4.6045) (24,3.7255) (25,3.0665) (26,2.1875) (27,1.7485) (28,1.3085) (29,0.6495) (30,0.4295) (31,2.4075) (32,5.4835) (33,8.999) (34,12.295) (35,13.614) (36,16.25) (37,16.031) (38,14.932) (39,12.954) (40,11.196) (41,9.658) (42,8.34) (43,7.241) (44,6.3625) (45,5.7035) (46,5.044) (47,4.6045) (48,4.165) (49,3.5055) (50,3.0665) (51,2.627) (52,2.1875) (53,1.7485) (54,1.968) (55,3.9455) (56,6.802) (57,9.4385) (58,12.514) (59,13.833) (60,16.031) (61,16.25) (62,15.371) (63,13.394)};
\addplot[apxts] coordinates {(-1,2.847) (0,1.5858) (1,0.92585) (2,0.18131) (3,-0.25567) (4,-0.62435) (5,-0.93433) (6,-0.9241) (7,0.82617) (8,4.0147) (9,8.0465) (10,10.994) (11,12.717) (12,13.019) (13,12.664) (14,11.594) (15,9.9773) (16,8.1154) (17,6.7092) (18,5.8104) (19,5.3597) (20,4.9092) (21,4.5697) (22,4.1498) (23,3.4349) (24,3.1041) (25,2.5844) (26,1.9803) (27,1.558) (28,1.0497) (29,0.60116) (30,0.51185) (31,1.8941) (32,4.6413) (33,8.0497) (34,10.643) (35,12.09) (36,12.32) (37,12.164) (38,11.411) (39,10.056) (40,8.4397) (41,7.2117) (42,6.3427) (43,5.9361) (44,5.5777) (45,5.368) (46,5.1551) (47,4.6483) (48,4.3595) (49,3.8649) (50,3.2161) (51,2.8188) (52,2.2404) (53,1.7146) (54,1.7298) (55,3.0751) (56,5.5084) (57,8.5899) (58,10.961) (59,12.304) (60,12.555) (61,12.586) (62,11.762) (63,10.578)};
\addplot[apxcr] coordinates {(-1,2.847) (0,1.344) (1,0.65296) (2,0.02595) (3,-0.33034) (4,-0.67093) (5,-0.9081) (6,-0.81688) (7,0.82125) (8,3.9651) (9,8.0153) (10,10.759) (11,12.49) (12,12.958) (13,12.84) (14,11.796) (15,9.9187) (16,7.9998) (17,6.4845) (18,5.5548) (19,5.2016) (20,4.7832) (21,4.3105) (22,3.9291) (23,3.3212) (24,2.919) (25,2.3566) (26,1.8401) (27,1.4681) (28,0.94043) (29,0.58526) (30,0.53572) (31,1.8846) (32,4.3195) (33,7.6571) (34,10.071) (35,11.78) (36,12.005) (37,12.005) (38,11.305) (39,9.6755) (40,7.9176) (41,6.5613) (42,5.5898) (43,5.2692) (44,4.9868) (45,4.6928) (46,4.4397) (47,3.9969) (48,3.7538) (49,3.2916) (50,2.7349) (51,2.3562) (52,1.8553) (53,1.4681) (54,1.5123) (55,2.7529) (56,4.8334) (57,7.7153) (58,9.9382) (59,11.425) (60,11.641) (61,11.912) (62,11.141) (63,9.8938)};
\addplot[apxro] coordinates {(-1,2.847) (0,1.7916) (1,0.83779) (2,0.070015) (3,-0.25832) (4,-0.83925) (5,-1.2406) (6,-0.6567) (7,1.4926) (8,5.0662) (9,9.2878) (10,12.337) (11,14.14) (12,14.567) (13,14.294) (14,12.992) (15,11.207) (16,9.2751) (17,7.5957) (18,6.6325) (19,5.9384) (20,5.378) (21,5.1038) (22,4.9188) (23,4.069) (24,3.7855) (25,3.3949) (26,2.7947) (27,2.1511) (28,1.5204) (29,1.0342) (30,1.4875) (31,3.4307) (32,5.7903) (33,9.5282) (34,12.021) (35,13.844) (36,14.267) (37,14.281) (38,13.204) (39,11.826) (40,10.07) (41,8.6514) (42,7.5834) (43,7.016) (44,6.4735) (45,6.2246) (46,6.0965) (47,5.6116) (48,5.3966) (49,5.1071) (50,4.2602) (51,3.6667) (52,3.2044) (53,2.9498) (54,2.9765) (55,4.6282) (56,7.2066) (57,10.445) (58,12.58) (59,13.775) (60,14.281) (61,14.465) (62,13.614) (63,12.396)};
\legend{Ground Truth,TS-RAG,Cross-RAG,ROAR (ours)}
\nextgroupplot[title={ETTm1: HUFL}]
\addplot[apxhist] coordinates {(-80,7.1) (-79,7.435) (-78,9.176) (-77,9.31) (-76,9.846) (-75,8.975) (-74,8.573) (-73,9.042) (-72,7.77) (-71,8.774) (-70,9.645) (-69,10.114) (-68,8.239) (-67,8.439) (-66,8.841) (-65,9.176) (-64,8.908) (-63,9.377) (-62,8.707) (-61,7.502) (-60,6.564) (-59,8.506) (-58,8.774) (-57,7.904) (-56,7.904) (-55,6.698) (-54,8.172) (-53,7.837) (-52,8.172) (-51,8.506) (-50,8.439) (-49,8.172) (-48,9.98) (-47,8.707) (-46,8.841) (-45,7.703) (-44,5.827) (-43,5.559) (-42,4.823) (-41,3.751) (-40,2.277) (-39,2.813) (-38,3.751) (-37,-4.823) (-36,-5.023) (-35,-2.009) (-34,-3.483) (-33,-1.34) (-32,-7.301) (-31,-1.942) (-30,-1.34) (-29,-3.885) (-28,-12.458) (-27,-15.204) (-26,-14.602) (-25,-12.056) (-24,-13.664) (-23,-7.234) (-22,-8.908) (-21,-14.602) (-20,-11.855) (-19,-10.583) (-18,-9.31) (-17,-7.502) (-16,-7.703) (-15,-6.631) (-14,-5.224) (-13,-4.22) (-12,-2.478) (-11,-0.268) (-10,0.938) (-9,2.947) (-8,4.555) (-7,7.167) (-6,7.703) (-5,8.908) (-4,8.506) (-3,8.506) (-2,8.372) (-1,8.908)};
\addplot[apxgt] coordinates {(-1,8.908) (0,8.439) (1,7.971) (2,8.038) (3,8.506) (4,8.038) (5,7.301) (6,7.167) (7,7.033) (8,6.43) (9,7.1) (10,7.301) (11,8.707) (12,9.109) (13,9.176) (14,8.506) (15,10.717) (16,10.449) (17,10.315) (18,11.788) (19,11.721) (20,11.119) (21,12.592) (22,11.855) (23,11.387) (24,11.32) (25,11.721) (26,11.922) (27,11.186) (28,11.855) (29,10.449) (30,10.449) (31,11.588) (32,10.717) (33,8.975) (34,9.779) (35,10.248) (36,9.98) (37,9.779) (38,9.913) (39,9.444) (40,9.578) (41,10.449) (42,10.315) (43,10.382) (44,9.444) (45,10.516) (46,10.583) (47,11.454) (48,11.721) (49,10.583) (50,10.449) (51,8.573) (52,6.899) (53,5.023) (54,3.55) (55,3.081) (56,0.804) (57,-0.871) (58,-3.952) (59,-5.76) (60,-5.626) (61,-6.966) (62,-7.703) (63,-7.77)};
\addplot[apxts] coordinates {(-1,8.908) (0,8.2365) (1,8.4897) (2,8.3188) (3,7.9427) (4,7.8066) (5,7.6073) (6,7.6481) (7,7.6175) (8,7.2288) (9,7.225) (10,7.4489) (11,7.622) (12,7.6612) (13,7.871) (14,8.2263) (15,8.7233) (16,9.3952) (17,9.7709) (18,10.029) (19,10.416) (20,10.515) (21,10.393) (22,10.414) (23,10.251) (24,10.297) (25,10.466) (26,10.477) (27,10.117) (28,9.8407) (29,9.6103) (30,9.5923) (31,9.5491) (32,9.2601) (33,9.1231) (34,9.1657) (35,9.0783) (36,9.0554) (37,9.2976) (38,9.3811) (39,9.4156) (40,9.6155) (41,9.7085) (42,9.6314) (43,9.7375) (44,9.9991) (45,9.9375) (46,10.118) (47,10.323) (48,10.258) (49,9.6057) (50,8.2547) (51,6.6262) (52,5.0963) (53,3.3282) (54,1.7923) (55,0.10315) (56,-2.0145) (57,-3.8929) (58,-5.6034) (59,-7.8755) (60,-9.5901) (61,-10.177) (62,-10.62) (63,-10.943)};
\addplot[apxcr] coordinates {(-1,8.908) (0,7.5872) (1,7.7447) (2,7.6981) (3,7.4722) (4,7.4349) (5,7.2629) (6,7.2771) (7,7.3245) (8,7.2201) (9,7.1903) (10,7.2231) (11,7.5283) (12,7.653) (13,7.5977) (14,7.8342) (15,8.2387) (16,8.508) (17,9.0162) (18,9.377) (19,10.076) (20,10.309) (21,9.8209) (22,9.9996) (23,10.115) (24,10.056) (25,10.412) (26,10.448) (27,9.9387) (28,9.6188) (29,9.0731) (30,9.0824) (31,9.3678) (32,9.0555) (33,8.9242) (34,8.8189) (35,8.7219) (36,8.7094) (37,8.8166) (38,8.6451) (39,8.7754) (40,8.7403) (41,9.0628) (42,8.9307) (43,9.3291) (44,9.7112) (45,9.5881) (46,9.7564) (47,10.22) (48,10.097) (49,9.5551) (50,8.3289) (51,6.6533) (52,5.341) (53,3.7644) (54,2.3958) (55,1.2469) (56,-0.41393) (57,-1.8489) (58,-3.1715) (59,-5.3939) (60,-7.2222) (61,-7.6988) (62,-8.3973) (63,-8.7582)};
\addplot[apxro] coordinates {(-1,8.908) (0,8.4448) (1,8.4232) (2,8.3984) (3,7.9106) (4,7.732) (5,7.3134) (6,7.4979) (7,7.4693) (8,7.3857) (9,7.5729) (10,7.7556) (11,7.957) (12,8.3305) (13,8.4712) (14,8.6867) (15,9.1625) (16,9.6476) (17,10.056) (18,10.576) (19,11.01) (20,11.015) (21,10.843) (22,10.949) (23,10.863) (24,11.158) (25,11.312) (26,11.3) (27,10.699) (28,10.407) (29,10.064) (30,10.121) (31,10.219) (32,9.9415) (33,9.7776) (34,10.091) (35,9.9429) (36,10.048) (37,10.052) (38,10.165) (39,10.179) (40,10.366) (41,10.448) (42,10.518) (43,10.41) (44,10.9) (45,10.939) (46,11.474) (47,11.402) (48,11.602) (49,10.836) (50,9.6282) (51,8.0746) (52,6.2974) (53,4.5848) (54,3.4332) (55,1.9218) (56,-0.1773) (57,-1.8397) (58,-2.9018) (59,-4.766) (60,-6.4876) (61,-7.0247) (62,-7.2837) (63,-7.353)};
\nextgroupplot[title={ETTm1: MUFL}]
\addplot[apxhist] coordinates {(-80,9.985) (-79,10.163) (-78,11.655) (-77,9.594) (-76,11.051) (-75,11.122) (-74,9.666) (-73,11.798) (-72,11.371) (-71,9.346) (-70,10.696) (-69,11.442) (-68,11.158) (-67,10.661) (-66,10.803) (-65,11.478) (-64,11.691) (-63,11.727) (-62,9.31) (-61,8.955) (-60,8.422) (-59,7) (-58,6.076) (-57,3.98) (-56,-4.797) (-55,-7.711) (-54,-8.99) (-53,-10.732) (-52,-11.549) (-51,-12.26) (-50,-12.686) (-49,-13.468) (-48,-14.214) (-47,-13.894) (-46,-15.564) (-45,-13.29) (-44,-17.483) (-43,-16.808) (-42,-13.823) (-41,-14.285) (-40,-14.25) (-39,-13.468) (-38,-11.975) (-37,-13.326) (-36,-10.021) (-35,-11.371) (-34,-10.696) (-33,-8.173) (-32,-6.787) (-31,-5.117) (-30,1.102) (-29,2.843) (-28,3.482) (-27,4.62) (-26,5.437) (-25,6.041) (-24,6.148) (-23,5.934) (-22,5.97) (-21,6.041) (-20,6.361) (-19,6.325) (-18,6.467) (-17,6.538) (-16,6.503) (-15,6.432) (-14,6.61) (-13,6.325) (-12,6.076) (-11,5.899) (-10,5.757) (-9,5.757) (-8,7.96) (-7,7.604) (-6,8.102) (-5,7.711) (-4,8.386) (-3,10.092) (-2,9.523) (-1,9.594)};
\addplot[apxgt] coordinates {(-1,9.594) (0,9.63) (1,8.564) (2,9.808) (3,10.412) (4,9.594) (5,10.056) (6,10.376) (7,10.661) (8,9.985) (9,9.95) (10,10.305) (11,9.985) (12,9.523) (13,9.097) (14,9.843) (15,10.163) (16,10.021) (17,10.341) (18,9.061) (19,10.412) (20,10.27) (21,11.087) (22,9.63) (23,9.843) (24,11.016) (25,10.447) (26,10.483) (27,9.523) (28,11.229) (29,10.589) (30,9.523) (31,9.666) (32,8.848) (33,10.767) (34,10.234) (35,9.559) (36,9.843) (37,7.071) (38,7.143) (39,4.797) (40,0.959) (41,-0.924) (42,-2.736) (43,-4.477) (44,-5.437) (45,-6.041) (46,-6.361) (47,-6.716) (48,-7) (49,-8.671) (50,-10.021) (51,-11.513) (52,-12.366) (53,-13.29) (54,-13.361) (55,-12.935) (56,-12.224) (57,-11.513) (58,-10.661) (59,-11.087) (60,-9.808) (61,-9.63) (62,-8.031) (63,-6.076)};
\addplot[apxts] coordinates {(-1,9.594) (0,9.0041) (1,9.0368) (2,9.1902) (3,9.4465) (4,9.0198) (5,8.8246) (6,9.2335) (7,9.5386) (8,9.4695) (9,9.2963) (10,9.5791) (11,9.7329) (12,9.2613) (13,9.0965) (14,9.3317) (15,9.6647) (16,9.4981) (17,9.5744) (18,10.364) (19,9.8357) (20,9.818) (21,9.7879) (22,9.8869) (23,9.6959) (24,9.165) (25,8.707) (26,9.1063) (27,9.6946) (28,9.5108) (29,9.4392) (30,9.5552) (31,9.5538) (32,9.386) (33,8.7143) (34,8.1915) (35,7.408) (36,6.7009) (37,5.6511) (38,4.3812) (39,1.8998) (40,-4.6016) (41,-8.1511) (42,-9.3615) (43,-11.221) (44,-12.174) (45,-12.835) (46,-13.246) (47,-13.78) (48,-14.02) (49,-14.176) (50,-14.105) (51,-14.182) (52,-14.859) (53,-14.889) (54,-14.462) (55,-14.177) (56,-14.391) (57,-14.166) (58,-13.696) (59,-13.125) (60,-12.234) (61,-11.135) (62,-10.122) (63,-8.786)};
\addplot[apxcr] coordinates {(-1,9.594) (0,9.1237) (1,8.9278) (2,8.9699) (3,9.0835) (4,8.9839) (5,8.9647) (6,8.9344) (7,9.6363) (8,9.3728) (9,9.0204) (10,9.3042) (11,9.3431) (12,9.2463) (13,8.7849) (14,8.4721) (15,9.2903) (16,8.6244) (17,8.3171) (18,9.1354) (19,8.7677) (20,9.0401) (21,8.5857) (22,8.1986) (23,8.9269) (24,8.5079) (25,7.5616) (26,8.5322) (27,9.4068) (28,9.0649) (29,8.9034) (30,8.8822) (31,9.1321) (32,8.8639) (33,7.9888) (34,7.2476) (35,5.6882) (36,5.0532) (37,3.8949) (38,2.6858) (39,0.6131) (40,-5.0667) (41,-7.8046) (42,-9.371) (43,-10.643) (44,-11.56) (45,-12.115) (46,-12.526) (47,-12.815) (48,-12.9) (49,-13.055) (50,-13.078) (51,-13.239) (52,-13.479) (53,-13.466) (54,-13.21) (55,-13.071) (56,-13.1) (57,-12.724) (58,-12.566) (59,-12.477) (60,-11.839) (61,-11.357) (62,-10.808) (63,-9.8544)};
\addplot[apxro] coordinates {(-1,9.594) (0,9.2998) (1,9.3408) (2,9.578) (3,9.5241) (4,9.3085) (5,9.2732) (6,9.4984) (7,9.8772) (8,9.7981) (9,9.673) (10,9.9831) (11,10.107) (12,9.9821) (13,9.9571) (14,9.6342) (15,9.6425) (16,9.2613) (17,9.3396) (18,9.9103) (19,9.6677) (20,9.582) (21,9.4746) (22,9.6983) (23,10.153) (24,9.7963) (25,9.4356) (26,9.656) (27,10.042) (28,9.851) (29,10.119) (30,10.32) (31,10.603) (32,10.294) (33,9.7544) (34,9.313) (35,8.469) (36,7.4045) (37,6.5606) (38,5.4132) (39,2.6315) (40,-2.5872) (41,-6.3655) (42,-7.0156) (43,-8.4183) (44,-9.2489) (45,-10.119) (46,-10.364) (47,-10.707) (48,-11.44) (49,-11.915) (50,-12.79) (51,-12.766) (52,-13.309) (53,-12.724) (54,-12.109) (55,-12.054) (56,-11.921) (57,-11.613) (58,-10.84) (59,-10.138) (60,-9.3646) (61,-8.6775) (62,-7.6089) (63,-6.7245)};
\nextgroupplot[title={ETTm2: OT}]
\addplot[apxhist] coordinates {(-80,11.635) (-79,11.635) (-78,11.196) (-77,11.196) (-76,10.757) (-75,10.757) (-74,10.538) (-73,10.318) (-72,10.318) (-71,10.097) (-70,10.097) (-69,9.878) (-68,9.878) (-67,9.6585) (-66,9.6585) (-65,9.4385) (-64,9.4385) (-63,9.2185) (-62,8.999) (-61,8.999) (-60,8.7795) (-59,8.5595) (-58,8.3395) (-57,8.12) (-56,7.9005) (-55,7.9005) (-54,7.6805) (-53,7.4605) (-52,7.4605) (-51,7.241) (-50,7.0215) (-49,7.0215) (-48,6.582) (-47,6.582) (-46,6.3625) (-45,6.3625) (-44,6.1425) (-43,5.923) (-42,5.923) (-41,5.7035) (-40,5.7035) (-39,5.4835) (-38,5.2635) (-37,5.2635) (-36,5.044) (-35,4.8245) (-34,4.8245) (-33,4.8245) (-32,4.8245) (-31,5.2635) (-30,5.7035) (-29,6.1425) (-28,7.0215) (-27,7.9005) (-26,8.7795) (-25,9.6585) (-24,10.757) (-23,11.636) (-22,12.734) (-21,13.613) (-20,14.712) (-19,15.591) (-18,16.47) (-17,17.348) (-16,17.569) (-15,17.788) (-14,18.008) (-13,18.447) (-12,19.326) (-11,19.765) (-10,20.205) (-9,20.645) (-8,20.864) (-7,20.864) (-6,20.865) (-5,20.645) (-4,20.424) (-3,20.424) (-2,20.205) (-1,19.766)};
\addplot[apxgt] coordinates {(-1,19.766) (0,19.546) (1,18.887) (2,18.667) (3,18.228) (4,17.569) (5,17.128) (6,16.25) (7,15.811) (8,15.371) (9,14.712) (10,14.272) (11,13.833) (12,13.393) (13,12.954) (14,12.514) (15,12.295) (16,11.856) (17,11.636) (18,11.196) (19,10.976) (20,10.757) (21,10.538) (22,10.318) (23,10.098) (24,9.878) (25,9.6585) (26,9.4385) (27,9.2185) (28,8.999) (29,8.7795) (30,8.7795) (31,8.5595) (32,8.3395) (33,8.3395) (34,8.12) (35,7.9005) (36,7.9005) (37,7.6805) (38,7.4605) (39,7.4605) (40,7.241) (41,7.0215) (42,7.0215) (43,6.8015) (44,6.582) (45,6.3625) (46,6.3625) (47,6.3625) (48,6.1425) (49,5.923) (50,5.923) (51,5.7035) (52,5.4835) (53,5.4835) (54,5.2635) (55,5.2635) (56,5.044) (57,5.044) (58,4.8245) (59,4.6045) (60,4.6045) (61,4.6045) (62,4.3845) (63,4.3845)};
\addplot[apxts] coordinates {(-1,19.766) (0,19.863) (1,19.411) (2,18.84) (3,18.433) (4,17.871) (5,17.195) (6,16.368) (7,15.758) (8,15.161) (9,14.671) (10,14.12) (11,13.582) (12,12.927) (13,12.381) (14,12.17) (15,11.824) (16,11.315) (17,10.929) (18,10.62) (19,10.09) (20,9.7025) (21,9.4437) (22,9.1704) (23,8.6812) (24,8.3563) (25,8.0991) (26,8.0027) (27,7.8747) (28,7.6801) (29,7.4037) (30,7.0922) (31,6.9615) (32,6.7363) (33,6.6719) (34,6.5074) (35,6.3349) (36,6.0731) (37,5.6699) (38,5.4211) (39,5.1197) (40,4.9333) (41,4.8638) (42,4.6049) (43,4.0903) (44,3.8454) (45,3.492) (46,3.422) (47,3.341) (48,3.2081) (49,3.0602) (50,2.9822) (51,2.888) (52,2.7587) (53,2.5767) (54,2.465) (55,2.3353) (56,2.2573) (57,2.2821) (58,2.1582) (59,2.1166) (60,1.9493) (61,1.9217) (62,1.8143) (63,1.8177)};
\addplot[apxcr] coordinates {(-1,19.766) (0,19.765) (1,19.272) (2,18.673) (3,18.353) (4,17.903) (5,17.323) (6,16.599) (7,16.158) (8,15.44) (9,14.863) (10,14.427) (11,13.937) (12,13.39) (13,13.024) (14,12.807) (15,12.56) (16,11.931) (17,11.594) (18,11.428) (19,10.93) (20,10.728) (21,10.296) (22,9.9872) (23,9.6953) (24,9.5487) (25,9.2155) (26,9.0794) (27,8.8936) (28,8.8204) (29,8.5706) (30,8.335) (31,8.3185) (32,7.8974) (33,7.6778) (34,7.4552) (35,7.1617) (36,6.9289) (37,6.6623) (38,6.4449) (39,6.3033) (40,6.0556) (41,5.9516) (42,5.7192) (43,5.4702) (44,5.2944) (45,4.9646) (46,4.7568) (47,4.5736) (48,4.4524) (49,4.1162) (50,4.0806) (51,4.0034) (52,3.9072) (53,3.7166) (54,3.6089) (55,3.4361) (56,3.3414) (57,3.2748) (58,3.194) (59,3.1072) (60,2.9301) (61,2.828) (62,2.8641) (63,2.9442)};
\addplot[apxro] coordinates {(-1,19.766) (0,19.914) (1,19.426) (2,18.896) (3,18.388) (4,17.782) (5,17.205) (6,16.669) (7,16.063) (8,15.533) (9,14.776) (10,14.384) (11,13.816) (12,13.32) (13,12.947) (14,12.597) (15,12.36) (16,11.87) (17,11.387) (18,11.25) (19,11.023) (20,10.856) (21,10.346) (22,10.015) (23,10.008) (24,9.9373) (25,9.6318) (26,9.3505) (27,9.4655) (28,9.1165) (29,8.9372) (30,8.8376) (31,8.7187) (32,8.5301) (33,8.2003) (34,8.2148) (35,7.9982) (36,7.7385) (37,7.4849) (38,7.5975) (39,7.4881) (40,7.1869) (41,6.7949) (42,6.8203) (43,6.5754) (44,6.5185) (45,6.1243) (46,6.054) (47,5.7825) (48,5.6939) (49,5.7369) (50,5.5171) (51,5.616) (52,5.2552) (53,5.0008) (54,4.7631) (55,4.5335) (56,4.4437) (57,4.3163) (58,4.3609) (59,4.0589) (60,3.9106) (61,3.6915) (62,3.7578) (63,3.7077)};
\nextgroupplot[title={ETTm2: OT}]
\addplot[apxhist] coordinates {(-80,32.29) (-79,31.851) (-78,31.411) (-77,31.191) (-76,30.532) (-75,30.093) (-74,29.653) (-73,29.214) (-72,28.774) (-71,28.335) (-70,27.896) (-69,27.456) (-68,27.017) (-67,26.577) (-66,26.138) (-65,25.698) (-64,25.259) (-63,24.819) (-62,24.6) (-61,24.16) (-60,23.941) (-59,23.721) (-58,23.501) (-57,23.062) (-56,23.062) (-55,22.842) (-54,22.842) (-53,22.842) (-52,22.842) (-51,22.842) (-50,22.842) (-49,22.842) (-48,22.842) (-47,22.842) (-46,23.062) (-45,23.062) (-44,23.062) (-43,23.281) (-42,23.281) (-41,23.281) (-40,23.281) (-39,23.281) (-38,23.062) (-37,23.062) (-36,23.062) (-35,23.062) (-34,22.842) (-33,22.842) (-32,22.842) (-31,22.842) (-30,22.842) (-29,22.842) (-28,22.622) (-27,22.622) (-26,22.622) (-25,22.622) (-24,22.622) (-23,22.622) (-22,22.622) (-21,22.622) (-20,22.622) (-19,22.622) (-18,22.622) (-17,22.622) (-16,22.842) (-15,23.062) (-14,23.501) (-13,23.941) (-12,24.38) (-11,24.819) (-10,25.259) (-9,26.138) (-8,26.797) (-7,27.236) (-6,27.676) (-5,27.896) (-4,28.335) (-3,28.994) (-2,29.653) (-1,30.752)};
\addplot[apxgt] coordinates {(-1,30.752) (0,31.631) (1,32.07) (2,32.509) (3,32.509) (4,33.169) (5,33.389) (6,33.609) (7,33.389) (8,33.609) (9,33.609) (10,33.828) (11,34.268) (12,34.268) (13,34.268) (14,34.048) (15,34.048) (16,33.609) (17,33.609) (18,33.169) (19,32.95) (20,32.729) (21,32.29) (22,31.631) (23,31.191) (24,30.532) (25,30.093) (26,29.434) (27,28.994) (28,28.555) (29,28.115) (30,27.676) (31,27.236) (32,26.797) (33,26.357) (34,26.138) (35,25.918) (36,25.479) (37,25.479) (38,25.039) (39,24.819) (40,24.819) (41,24.6) (42,24.38) (43,24.38) (44,24.16) (45,24.16) (46,23.941) (47,23.721) (48,23.721) (49,23.501) (50,23.501) (51,23.281) (52,23.062) (53,23.062) (54,22.842) (55,22.842) (56,22.622) (57,22.402) (58,22.402) (59,22.183) (60,21.963) (61,21.743) (62,21.743) (63,21.523)};
\addplot[apxts] coordinates {(-1,30.752) (0,32.033) (1,32.306) (2,32.58) (3,32.901) (4,33.139) (5,33.425) (6,33.648) (7,33.861) (8,34.017) (9,34.176) (10,34.327) (11,34.342) (12,34.213) (13,34.08) (14,33.855) (15,33.629) (16,33.598) (17,33.328) (18,33.001) (19,32.657) (20,32.18) (21,31.754) (22,31.252) (23,30.707) (24,30.123) (25,29.593) (26,29.105) (27,28.611) (28,28.111) (29,27.661) (30,27.138) (31,26.708) (32,26.293) (33,25.847) (34,25.459) (35,25.127) (36,24.72) (37,24.399) (38,24.051) (39,23.675) (40,23.605) (41,23.442) (42,23.221) (43,22.975) (44,22.812) (45,22.699) (46,22.564) (47,22.445) (48,22.215) (49,22.045) (50,22.019) (51,22.09) (52,22.112) (53,22.028) (54,21.944) (55,21.82) (56,21.667) (57,21.486) (58,21.473) (59,21.546) (60,21.47) (61,21.473) (62,21.241) (63,21.009)};
\addplot[apxcr] coordinates {(-1,30.752) (0,31.974) (1,32.143) (2,32.572) (3,32.958) (4,33.081) (5,33.346) (6,33.753) (7,34.086) (8,34.278) (9,34.352) (10,34.491) (11,34.535) (12,34.458) (13,34.381) (14,34.167) (15,33.94) (16,33.762) (17,33.303) (18,33.068) (19,32.667) (20,32.199) (21,31.728) (22,31.224) (23,30.759) (24,30.101) (25,29.554) (26,29.028) (27,28.482) (28,27.964) (29,27.567) (30,27.069) (31,26.855) (32,26.467) (33,25.878) (34,25.415) (35,25.151) (36,24.804) (37,24.482) (38,24.185) (39,23.918) (40,23.884) (41,23.658) (42,23.405) (43,23.261) (44,23.094) (45,22.935) (46,22.74) (47,22.511) (48,22.215) (49,22.028) (50,22.033) (51,21.958) (52,21.933) (53,21.757) (54,21.73) (55,21.658) (56,21.507) (57,21.412) (58,21.43) (59,21.497) (60,21.381) (61,21.264) (62,21.19) (63,21.06)};
\addplot[apxro] coordinates {(-1,30.752) (0,31.93) (1,32.266) (2,32.653) (3,32.874) (4,33.005) (5,33.294) (6,33.644) (7,34.016) (8,34.06) (9,34.219) (10,34.413) (11,34.555) (12,34.486) (13,34.49) (14,34.229) (15,33.946) (16,33.655) (17,33.379) (18,33.059) (19,32.826) (20,32.46) (21,31.958) (22,31.298) (23,30.874) (24,30.508) (25,29.984) (26,29.393) (27,28.746) (28,28.211) (29,27.796) (30,27.447) (31,26.992) (32,26.917) (33,26.569) (34,26.058) (35,25.704) (36,25.331) (37,25.022) (38,24.765) (39,24.571) (40,24.511) (41,24.22) (42,24.224) (43,23.795) (44,23.825) (45,23.756) (46,23.541) (47,23.447) (48,23.347) (49,23.195) (50,22.991) (51,22.884) (52,22.807) (53,22.853) (54,22.691) (55,22.668) (56,22.604) (57,22.517) (58,22.338) (59,22.365) (60,22.214) (61,22.136) (62,21.909) (63,21.6)};
\nextgroupplot[title={Weather: p}]
\addplot[apxhist] coordinates {(-80,955.94) (-79,956.02) (-78,956.01) (-77,956.09) (-76,956.19) (-75,956.19) (-74,956.21) (-73,956.29) (-72,956.32) (-71,956.35) (-70,956.37) (-69,956.36) (-68,956.2) (-67,956.18) (-66,956.14) (-65,956.16) (-64,956.07) (-63,956.05) (-62,956.03) (-61,955.89) (-60,955.83) (-59,955.83) (-58,955.73) (-57,955.64) (-56,955.58) (-55,955.59) (-54,955.74) (-53,955.94) (-52,956.06) (-51,956.08) (-50,956.11) (-49,956.18) (-48,956.27) (-47,956.32) (-46,956.32) (-45,956.19) (-44,956.13) (-43,956.11) (-42,956.23) (-41,956.46) (-40,956.55) (-39,956.62) (-38,956.71) (-37,956.66) (-36,956.77) (-35,956.8) (-34,956.58) (-33,956.57) (-32,956.63) (-31,956.67) (-30,956.74) (-29,956.78) (-28,956.85) (-27,956.88) (-26,956.76) (-25,956.73) (-24,956.84) (-23,956.89) (-22,956.89) (-21,956.98) (-20,956.97) (-19,956.95) (-18,957.02) (-17,957) (-16,957.07) (-15,957.2) (-14,957.27) (-13,957.43) (-12,957.42) (-11,957.41) (-10,957.55) (-9,957.61) (-8,957.63) (-7,957.58) (-6,957.59) (-5,957.75) (-4,957.86) (-3,957.95) (-2,957.96) (-1,958.16)};
\addplot[apxgt] coordinates {(-1,958.16) (0,958.31) (1,958.33) (2,958.37) (3,958.54) (4,958.6) (5,958.74) (6,958.81) (7,958.91) (8,959.02) (9,959.08) (10,959.2) (11,959.27) (12,959.36) (13,959.44) (14,959.54) (15,959.69) (16,959.77) (17,959.92) (18,960.05) (19,960.22) (20,960.41) (21,960.56) (22,960.66) (23,960.75) (24,960.93) (25,961.05) (26,961.14) (27,961.36) (28,961.45) (29,961.58) (30,961.84) (31,961.98) (32,962.09) (33,962.35) (34,962.53) (35,962.7) (36,963) (37,963.25) (38,963.53) (39,963.78) (40,963.98) (41,964.21) (42,964.4) (43,964.61) (44,964.86) (45,965.1) (46,965.3) (47,965.56) (48,965.81) (49,965.97) (50,966.21) (51,966.46) (52,966.62) (53,966.82) (54,966.89) (55,966.98) (56,967.19) (57,967.34) (58,967.44) (59,967.7) (60,967.92) (61,967.99) (62,968.11) (63,968.12)};
\addplot[apxts] coordinates {(-1,958.16) (0,958.26) (1,958.33) (2,958.4) (3,958.46) (4,958.43) (5,958.44) (6,958.62) (7,958.62) (8,958.64) (9,958.71) (10,958.74) (11,958.79) (12,958.88) (13,958.87) (14,958.88) (15,958.86) (16,959.05) (17,958.96) (18,959.04) (19,959.05) (20,959.08) (21,959.11) (22,959.22) (23,959.34) (24,959.47) (25,959.31) (26,959.47) (27,959.64) (28,959.5) (29,959.49) (30,959.76) (31,959.86) (32,959.96) (33,959.92) (34,959.94) (35,959.97) (36,960.04) (37,960.14) (38,960.16) (39,960.29) (40,960.48) (41,960.34) (42,960.65) (43,960.56) (44,960.63) (45,960.64) (46,960.96) (47,961.24) (48,961.61) (49,961.63) (50,961.75) (51,961.93) (52,961.91) (53,961.95) (54,962.2) (55,962.52) (56,962.7) (57,962.65) (58,962.92) (59,963.17) (60,963.36) (61,963.49) (62,963.5) (63,963.64)};
\addplot[apxcr] coordinates {(-1,958.16) (0,958.45) (1,958.52) (2,958.57) (3,958.45) (4,958.56) (5,958.65) (6,958.85) (7,958.91) (8,958.96) (9,958.88) (10,958.96) (11,959.17) (12,959.37) (13,959.39) (14,959.5) (15,959.53) (16,959.63) (17,959.81) (18,960.21) (19,960.5) (20,960.73) (21,960.95) (22,961.06) (23,961.2) (24,961.08) (25,961.19) (26,961.32) (27,961.42) (28,961.25) (29,961.38) (30,961.54) (31,961.9) (32,961.94) (33,961.97) (34,962.06) (35,962.18) (36,962.29) (37,962.24) (38,962.07) (39,962.22) (40,962.3) (41,962.49) (42,962.66) (43,963.05) (44,962.82) (45,963.1) (46,963.13) (47,963.34) (48,963.73) (49,963.53) (50,963.47) (51,963.58) (52,963.62) (53,963.88) (54,964.38) (55,964.59) (56,964.93) (57,964.84) (58,965.34) (59,965.52) (60,965.66) (61,966.13) (62,965.86) (63,966.21)};
\addplot[apxro] coordinates {(-1,958.16) (0,957.46) (1,957.31) (2,957.62) (3,957.71) (4,958.04) (5,958.15) (6,958.51) (7,958.39) (8,958.6) (9,958.6) (10,959.13) (11,959.16) (12,959.57) (13,959.52) (14,959.45) (15,959.7) (16,959.86) (17,959.78) (18,960.21) (19,960.5) (20,960.63) (21,960.93) (22,960.96) (23,961.04) (24,961.52) (25,961.44) (26,961.96) (27,961.86) (28,961.93) (29,961.84) (30,962.13) (31,962.34) (32,962.48) (33,962.63) (34,962.74) (35,962.7) (36,962.96) (37,963.3) (38,963.08) (39,963.51) (40,963.8) (41,963.79) (42,963.91) (43,964.21) (44,964.21) (45,964.61) (46,964.69) (47,964.91) (48,965.16) (49,965.2) (50,965.47) (51,965.76) (52,965.54) (53,965.87) (54,966.04) (55,966.33) (56,966.4) (57,966.36) (58,966.73) (59,966.55) (60,967.02) (61,967.39) (62,967.65) (63,968.04)};
\end{groupplot}
\path (group c1r2.outer south west) -- (group c3r2.outer south east) node[midway, below=2pt] {\pgfplotslegendfromname{apxlegendapp-vis-baselines}};
\end{tikzpicture}
\caption{Additional forecasting visualizations comparing ROAR with TS-RAG and Cross-RAG on ETTh2, ETTm1, ETTm2 and Weather.}
\label{fig:app-vis-baselines}
\end{figure}